\documentclass[review,english]{elsarticle}

\usepackage{tikz}
\usepackage{tabularx}
\usepackage{mathrsfs}
\usepackage{multirow}
\usepackage{amsmath}
\usepackage{booktabs}
\usepackage{tablefootnote}
\usepackage{amssymb}
\usepackage{graphicx}
\usepackage[outdir=./]{epstopdf}

\usetikzlibrary{shapes,arrows,fit,calc,positioning}
\tikzset{box/.style={draw, diamond, thick, text centered, minimum height=0.5cm, minimum width=1cm}}
\tikzset{leaf/.style={draw, rectangle, thick, text centered, minimum height=0.5cm, minimum width=1cm}}
\tikzset{line/.style={draw, thick, -latex'}}

\date{}
\begin{document}
\title{Evolutionary learning of interpretable decision trees}

\author{Leonardo Lucio Custode}
\ead{leonardo.custode@unitn.it}
\author{Giovanni Iacca}
\ead{giovanni.iacca@unitn.it}
\address{Department of Information Engineering and Computer Science\\
University of Trento\\
Via Sommarive 9, 38123 Povo (Trento), Italy}
\date{{\copyright\ 2021. License: CC-BY-NC-ND 4.0
http://creativecommons.org/licenses/by-nc-nd/4.0/}}

\begin{abstract}
Reinforcement learning techniques achieved human-level performance in several tasks in the last decade.
However, in recent years, the need for interpretability emerged: we want to be able to understand how a system works and the reasons behind its decisions.
Not only we need interpretability to assess the safety of the produced systems, we also need it to extract knowledge about unknown problems.
While some techniques that optimize decision trees for reinforcement learning do exist, they usually employ greedy algorithms or they do not exploit the rewards given by the environment.
This means that these techniques may easily get stuck in local optima.
In this work, we propose a novel approach to interpretable reinforcement learning that uses decision trees. 
We present a two-level optimization scheme that combines the advantages of evolutionary algorithms with the advantages of Q-learning.
This way we decompose the problem into two sub-problems: the problem of finding a meaningful and useful decomposition of the state space, and the problem of associating an action to each state.
We test the proposed method on three well-known reinforcement learning benchmarks, on which it results competitive with respect to the state-of-the-art in both performance and interpretability.
Finally, we perform an ablation study that confirms that using the two-level optimization scheme gives a boost in performance in non-trivial environments with respect to a one-layer optimization technique.
\end{abstract}


\maketitle

\section{Introduction}
\label{sec:intro}
While machine learning is achieving very promising results in a variety of fields, there is an emergent need to understand what happens in the learned model, for testing, security and safety purposes.

There are mainly two approaches that try to address this problem: explainable AI (XAI) and interpretable AI (IAI) (which is actually a subset of XAI).

The field of explainable AI, in recent years, has seen a significant increase in the number of scientific contributions related to the topic. 
It is important to note, however, that these techniques are not applicable to all the tasks.
In fact, as stated by \cite{rudin_stop_2019}, it is not safe to apply XAI techniques to safety-critical or high-stakes systems.
This is due to the fact that explanations are usually \textit{approximations} of the actual models and, as a consequence, do not represent \textit{exactly} the models.

Interpretable AI techniques, instead, are based on the use of interpretable models, i.e. models that can be easily understood and inspected by an human operator \cite{barredo_arrieta_explainable_2020}.
These techniques, besides the ability to assess security and safety of the produced models, can also serve to better \textit{understand} a problem.
In fact, by looking at an interpretable model (with good performance), an human operator can \textit{extract} knowledge about the problem faced.

However, interpretable AI techniques are not widely used in practice, due to their (usually) lower performance.
In fact, it is widely accepted (although not proved) that a trade-off between interpretability and performance exists.

Recent work has addressed the problem of building interpretable reinforcement learning models.
In \cite{silva_optimization_2020}, the authors implement a differentiable version of decision trees and optimize them by using backpropagation.
Dhebar et al. \cite{dhebar_interpretable-ai_2020} propose nonlinear decision trees to approximate and refine an oracle policy.

While the results of these approaches seem very promising, the structure of the tree must be defined a-priori.
This requires us to either perform a trial-and-error search or to include prior knowledge.

In this work, we present a novel approach to the training of interpretable reinforcement learning agents that combines artificial evolution and lifelong reinforcement-learning.
This two-level optimization algorithm allows us to decrease the amount of prior knowledge given to the algorithm.
Our approach is able to generate agents in the form of decision trees that are able to learn both a decomposition of the space and the state-action mapping.

The contributions of this paper are the following:
\begin{itemize}
    \item We propose a two-level optimization approach that optimizes both the topology of the tree and the decisions taken for each state
    \item We perform experimental tests on classic reinforcement learning problems: CartPole, MountainCar and LunarLander
    \item We perform a comparison of the produced agents w.r.t. the interpretable and the non-interpretable state-of-the-art
    \item We quantitatively measure the interpretability of our solutions and compare it to the state-of-the-art
    \item We interpret the solutions produced to understand how the agents work
\end{itemize}

This paper is structured as follows.
In Section \ref{ref:theory} we give some background on the field and related work, while Section \ref{sec:method} describes the method used in our approach.
Then, in Section \ref{sec:results} we present the results of our experiments. In Section \ref{sec:discussion} we will discuss our results by comparing them to the interpretable state-of-the-art, performing an ablation study and interpreting the produced solutions. Finally, in Section \ref{sec:conclusions} we draw the conclusions of this work.

\section{Related work}
\label{ref:theory}
In this section we will give some background on the research problem being faced.

The use of decision trees to learn in reinforcement learning tasks has been explored in several previous work . 

McCallum, in \cite{mccallum_1996_reinforcement}, proposes U-Trees: a kind of trees able to perform reinforcement learning that handle the following sub-problems: choice of the memories, selective perception and value function approximation.
In \cite{uther1998tree}, the authors extend U-Trees in order to make them able to cope with continuous environment. They propose two novel tests that are used to create new conditions that split the state-space. They test the proposed approach in two environments, a continuous one and an ordered-discrete one, and their results show that their approach is competitive with respect to other approaches.

Pyeatt and Howe, in \cite{pyeatt2001decision}, propose a novel splitting criterion to build trees that are able to perform value function approximation. In their experiments, they compare the performance obtained by using their approach to the ones obtained using other splitting criteria, a table-lookup approach and a neural network. The results show that the proposed approach usually achieves better performance than all the other approaches.

In \cite{roth_conservative_2019} the authors propose a method that predicts the gain obtained by adding a split to the tree and select the best split to grow the tree.
The experimental results show that this method is more effective than the method proposed in \cite{pyeatt2001decision} on the tested environment.

Silva et al. \cite{silva_optimization_2020} propose an approach to interpretable reinforcement learning that uses Proximal Policy Optimization on differentiable decision trees. Moreover, they provide an analysis of the learning process while using either Q-learning or PPO. The experimental results show that this approach is able to produce competitive solutions in some of the tasks. However, it is also shown that when discretizing the differentiable decision trees into typical decision trees, the performance may heavily decrease.

In \cite{dhebar_interpretable-ai_2020}, the authors used evolutionary algorithms to evolve non-linear decision trees. By non-linear, the authors mean that each split does not define a linear hyperplane in the feature-space. The experimental results show that this approach is able to obtain competitive performance with respect to a neural network based approach.

In \cite{di_chio_evolving_2011}, the authors use the grammatical evolution algorithm \cite{goos_grammatical_1998} to evolve behavior trees (tree-based structures that allow more complex operations than a decision tree) for the Mario AI competitions.
The proposed agent can perform basic actions or pre-determined combinations of basic actions.
Their solution achieved the fourth place in the Mario AI competition.
However, in this work the authors only evolve a controller, not exploiting the rewards given by the environment to increase the performance of the agent.

Hallawa et al., in \cite{hallawa_evo-rl_2020}, use behavior trees as evolved instinctive behaviors in agents that are then combined with a learned behavior. 
While behavior trees are usually interpretable, the authors did not take explicitly into account the interpretability of the whole model, which comprises both a behavior tree and either a neural network or a Q-learning table.

Several work applied the evolutionary computation paradigm to evolve tree-based structures outside the reinforcement learning domain.

Kr\c{e}towski, in \cite{kretowski_memetic_2008}, proposes a memetic algorithm based on genetic programming \cite{koza_genetic_1992} and local search to optimize decision trees.
The results presented show that this approach is able to obtain performance that is comparable to the state-of-the-art, while keeping the size of the tree significantly lower.

In \cite{czajkowski_multi-objective_2019}, the authors propose a multi-objective EA to evolve regression trees and model trees.
They use a Pareto front to optimize RMSE, number of nodes, number of attributes.
The experimental results show that this approach is able to obtain performance that are comparable or better than the state-of-the-art while using less nodes and less attributes.

In \cite{llora_evolution_nodate, llora_knowledge-independent_nodate}, the authors use the Genetic and Artificial Life Environment (GALE) to evolve decision trees. Their results show that GALE is able to produce decision trees that are competitive with the state-of-the-art.

\section{Method}
\label{sec:method}
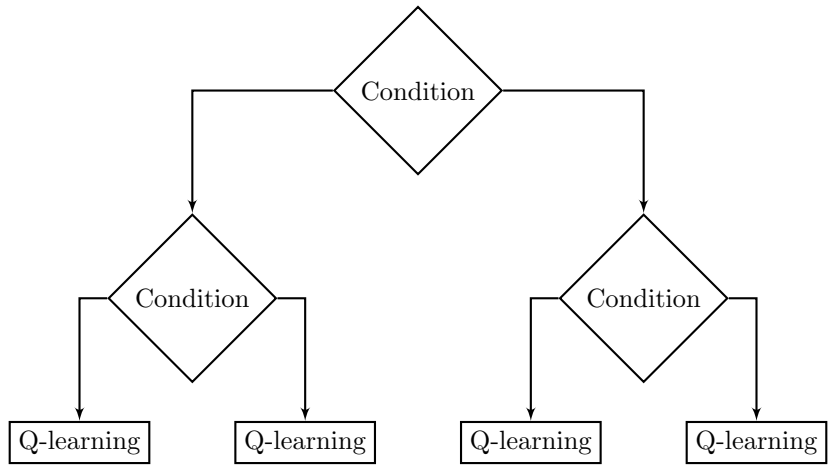
\begin{figure}
    \centering
    \begin{tikzpicture}[auto]
            \node [box]                                         (root)  {Condition};
            \node [box, below=0.5cm of root, xshift=-3cm]       (t)     {Condition};
            \node [leaf, below=0.5cm of t, xshift=-1.5cm]       (tt)    {Q-learning};
            \node [leaf, below=0.5cm of t, xshift=1.5cm]        (tf)    {Q-learning};
            \node [box, below=0.5cm of root, xshift=3cm]        (f)     {Condition};
            \node [leaf, below=0.5cm of f, xshift=-1.5cm]       (ft)    {Q-learning};
            \node [leaf, below=0.5cm of f, xshift=1.5cm]        (ff)    {Q-learning};
            
            \path [line] (root) -|          (t);
            \path [line] (root) -|          (f);
            \path [line] (t) -|             (tt);
            \path [line] (t) -|             (tf);
            \path [line] (f) -|             (ft);
            \path [line] (f) -|             (ff);
        \end{tikzpicture}
    \caption{A high-level representation of the proposed agent in the form of a decision tree.}
    \label{fig:tree_agent}
\end{figure}

In this work, we aim to produce interpretable agents in the form of decision trees.
Decision trees are trees (usually binary trees) where each inner node represents a ``split'' (i.e. a test on a condition) and each leaf node contains a decision.
A representation of the proposed decision trees is shown in Figure \ref{fig:tree_agent}.

When using decision trees for reinforcement learning tasks, there are two problems that need to be assessed:
\begin{enumerate}
    \item How do we choose the splits?
    \item Given a leaf, what action do we need to assign to this leaf?
\end{enumerate}
Of course, there is an important relationship between splits and decisions taken in the leaves, so changing one of these without changing the other may lead to significant changes in performance.

Several works \cite{roth_conservative_2019, mccallum_1996_reinforcement, uther1998tree, pyeatt2001decision} use greedy heuristics to induce the trees. However, this approaches have the following drawbacks:
\begin{itemize}
    \item They use greedy rules to expand the trees: since inducing decision trees is an NP-complete problem \cite{hyafil_constructing_1976}, this may cause the induction of sub-optimal trees (i.e. stuck in local optima) of poor quality \cite{mccallum_1996_reinforcement, fan_regression_2005}.
    \item They use tests to expand the trees: this causes these algorithms to suffer from the curse of dimensionality because, for each expansion of the tree, all the input variables need to be tested \cite{mccallum_1996_reinforcement, uther1998tree}.
\end{itemize}

Other works \cite{di_chio_evolving_2011} (and \cite{kretowski_memetic_2008, czajkowski_multi-objective_2019, llora_evolution_nodate, llora_knowledge-independent_nodate}, even if they are not applied to reinforcement learning tasks) induce trees by means of evolutionary approaches. 
However, these approaches only rely on the evolutionary algorithm. 
In RL tasks, not exploiting the reinforcement signals given by the environment may slow down the evolution and so result in a less-efficient process.

Our approach, instead, aims to combine artificial evolution and reinforcement-learning methods to take the best of both worlds.
We propose a Baldwinian-evolutionary approach to optimize simultaneously the structure of the tree and the state-action function.
Baldwinian evolution is an evolutionary theory that, opposed to Darwinian evolution and Lamarckian evolution, states that what an individual learns during his life is not passed to their offspring. However, the knowledge acquired by the individual may be an evolutionary advantage that modifies the fitness landscape.

We do so by using an evolutionary algorithm to evolve the structure of the decision tree, while using $\mathcal{Q}$-learning to learn the state-action function.
This way, we search for trees that decompose the state-space in such a manner that, when taking optimal actions, maximize the reward of the agent. 

The evolutionary algorithm we use is the Grammatical Evolution (GE) \cite{goos_grammatical_1998}. This evolutionary algorithm evolves (context-free) grammars in the Backus-Naur Form.

\begin{figure}
    \centering
    \includegraphics[scale=0.8]{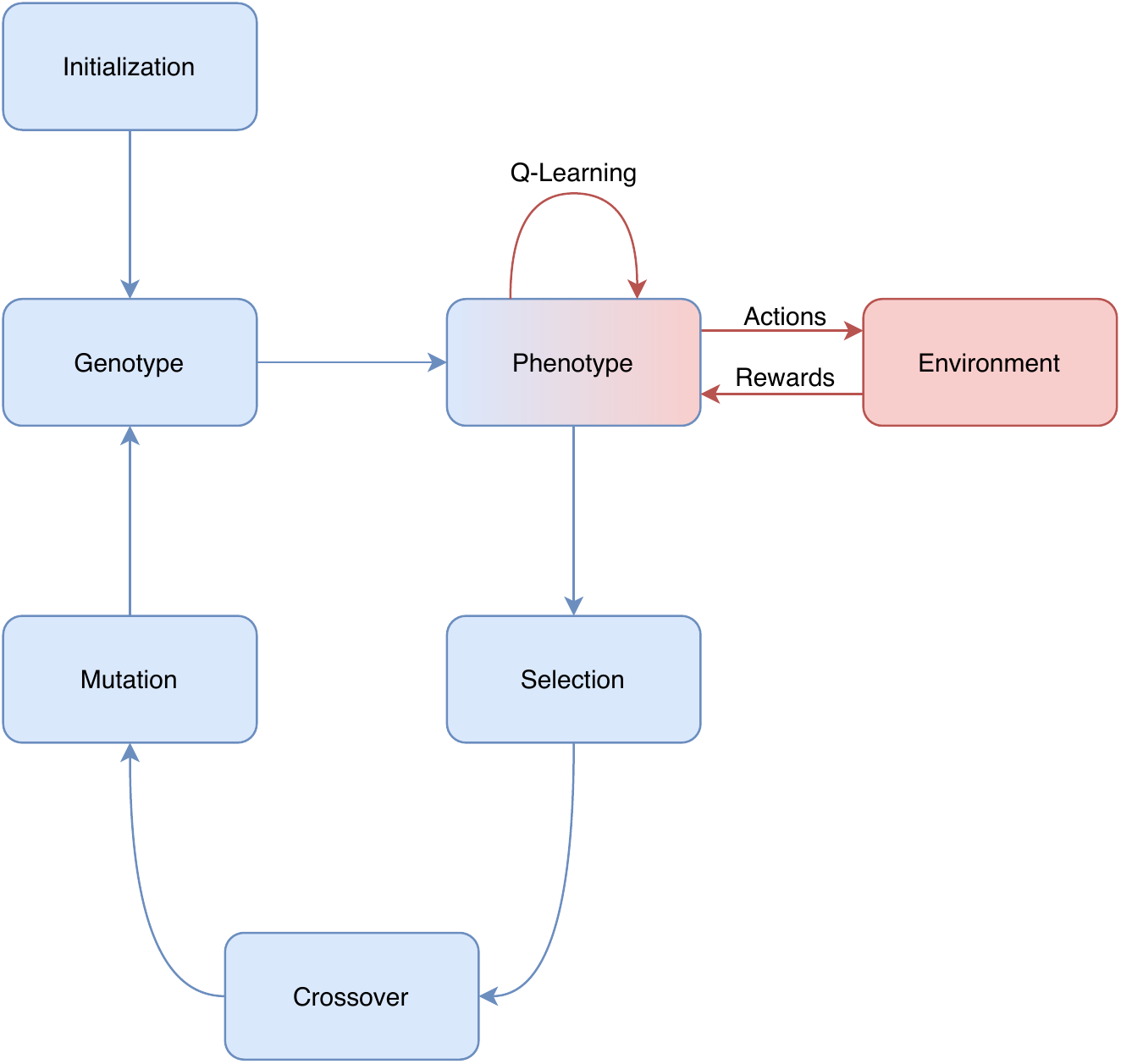}
    \caption{A scheme of the inner working of the proposed algorithm. The blue blocks are the ones that derive from the evolutionary part of our algorithm, while the red blocks are the ones that derive from the Q-learning part.}
    \label{fig:scheme}
\end{figure}

Figure \ref{fig:scheme} shows a block diagram that clarifies the inner working of the proposed algorithm.
The blue-colored parts are the processes inherent the evolutionary part of our algorithm, while the red-colored ones are the processes inherent the reinforcement-learning part.

\subsection{Evolutionary algorithm}
To evolve decision trees, we evolve an associated grammar, similarly to the approach described in \cite{goos_grammatical_1998}.
In this subsection we will describe our algorithm design, highlighting the differences with the original Grammatical Evolution.

\subsubsection{Individual encoding}
The genotype of an individual is encoded as a list of codons (represented as integers).
However, differently from \cite{goos_grammatical_1998}, the genotype has fixed length.

\subsubsection{Mutation operator}
Instead of the mutation operator described in \cite{goos_grammatical_1998}, we use a classical uniform mutation.
This operator mutates each gene according to a probability.
The new value of the gene is drawn uniformly from the range of variation of the variable.
However, since the grammar may have a different number of productions for each rule, we uniformly draw a random number between 0 and a number bigger than the maximum number of productions in the grammar.
Then, by using the modulo operator, we choose the production from the production rule.

\subsubsection{Crossover operator}
As a crossover operator, we use the standard one-point crossover operator.
This operator simply sets a random cutting point and creates two individuals by mixing the two sub-strings of the genotype.
This means that we do not prune the individuals that have genes that are not expressed in the phenotype.

\subsubsection{Replacement of the individuals}
Instead of replacing all the individuals with their offspring (intended as the copies that undergo mutation/crossover), we replace a parent only when the fitness of its offspring is better than the fitness of the parent.
Moreover, when an offspring has two parents (i.e. is a product of crossover), it replaces the parent with lowest fitness.
In case two offspring have better fitness than only one of the parents, the best one replaces the worst parent.

This mechanism allows us to preserve diversity between the individuals and at the same time makes the fitness trend monotonically increasing.

\subsubsection{Fitness evaluation}
The fitness evaluation process consists in the following steps.
First of all, the genotype is translated to the corresponding phenotype.
Then, for each timestep, the policy encoded by the tree is executed and the reward signals obtained from the environment are used to update the Q-values of the leaves.

\section{Results}
\label{sec:results}
To test our approach, we test it in the following OpenAI Gym \cite{brockman2016openai} environments: 
\begin{itemize}
    \item CartPole-v1
    \item MountainCar-v0
    \item LunarLander-v2
\end{itemize}

In this section, we will show the results obtained and compare them to the state-of-the-art using two metrics: the score given by the simulator and a metric of complexity (from the interpretability point of view).

We adopted the interpretability metric proposed in \cite{virgolin_learning_2020}:
\[ \mathcal{M}_{orig} = 79.1 - 0.2 l - 0.5 n_o - 3.4 n_{nao} - 4.5 n_{naoc} \]
where:
\begin{itemize}
    \item $l$ is the size of the formula (i.e. sum of constants, variables and operations)
    \item $n_o$ is the number of operations
    \item $n_{nao}$ is the number of non-arithmetical operations
    \item $n_{naoc}$ is the number of consecutive compositions of non-arithmetical operations.
\end{itemize}

However, this metric was meant to be bounded between 0 and 100, so we modified the metric in order to make it work as a \textit{complexity}.
For this purpose, the metric used is the following:
\[ \mathcal{M} = -0.2 + 0.2 l + 0.5 n_o + 3.4 n_{nao} + 4.5 n_{naoc} \]
The changes we made yield the following properties:
\begin{itemize}
    \item By changing the sign of all the terms, we obtain that a model with an higher complexity is more hard to interpret
    \item We replaced the constant with -0.2, so that when we have a constant (best case from the point of view of the interpretability) its complexity becomes 0
\end{itemize}
Furthermore, it is important to note that this metric is in line with what Lipton states in \cite{lipton_mythos_2017}. 
In fact, we can easily note that huge decision trees will be as interpretable as black-box methods, because the terms $l, n_o, n_{nao}$ and $n_{naoc}$ will have a high magnitude.
Also, $\mathcal{M}$ seems to be (loosely) in line with what the authors say in \cite{barcelo_model_2020}.
In fact, by using the number of operations (although there are other variables in the metric we use) we loosely resemble the computational complexity of the model that we are executing.

To assess the statistical repeatability of our experiments, we perform 10 independent runs for each setting.
For each run, as required by \cite{brockman2016openai}, we test the best model for each run over 100 independent episodes to assess its performance.
By ``testing'', we mean that the policy is executed in 100 unseen episodes.

\subsection{Simplification mechanism}
To make our solutions even more interpretable, we introduce a simplification mechanism that is executed on the final solutions.
The simplification mechanism is the following.
First of all, we execute the given policy for 100 episodes in a validation environment.
Here, we keep a counter for each node of the tree that is increased each time the node is visited.
Then, once this phase is finished, we remove all the nodes that have not been visited.
Finally, we iteratively search for nodes in the tree whose leaves correspond to the same action.
Each time such a node is found, it is replaced with a leaf that contains the action common to the two leaves.
The iteration stops when the tree does not contain nodes of this type.

\subsection{Description of the environments}
In this subsection we will describe the environments used and their properties.

\subsubsection{CartPole-v1}
In this task the agent has to balance a pole that stands on top of a cart by moving the cart either to the left or to the right.

\paragraph{Observation space}
The state of the environment is composed of the following features:
\begin{itemize}
    \item Cart position: $x \in \left[-4.8, 4.8\right] m$
    \item Cart velocity: $v \in \left]-\infty, \infty\right[ m/s$
    \item Pole angle: $\theta \in \left[-0.418, 0.418\right] rad$
    \item Pole angular velocity: $\omega \in \left]-\infty, \infty\right[ rad/s$ 
\end{itemize}

\paragraph{Action space}
The actions that the agent can perform are:
\begin{itemize}
    \item Push the cart to the left by applying a force of 10N ($move\_left$)
    \item Push the cart to the right by applying a force of 10N ($move\_right$)
\end{itemize}

\paragraph{Rewards}
The agent receives a reward of +1 for each timestep.

\paragraph{Termination criterion}
The simulation terminates if:
\begin{itemize}
    \item The cart position lies outside the bounds for the $x$ variable
    \item The angle of the pole lies outside the bounds for the $\theta$ variable
\end{itemize}

\paragraph{Resolution criterion}
This task is considered as solved if the agent receives a mean total reward $R \geq 475$ on 100 runs.

\subsubsection{MountainCar-v0}
In this environment the agent has to drive a car, which is initially in a valley, up on a hill.
However, the engine of the car is not powerful enough so the agent has to learn how to build momentum by exploiting the two hills.

\paragraph{Observation space}
The state of the environment consists in the following variables:
\begin{itemize}
    \item Horizontal position of the car: $x \in \left[-1.2, 0.6\right] m$
    \item Horizontal velocity of the car: $v \in \left[-0.07, 0.07\right] m/s$
\end{itemize}

\paragraph{Action space}
The agent can perform 3 actions:
\begin{enumerate}
    \item Accelerate to the left by applying a force of 0.001N
    \item Do not accelerate
    \item Accelerate to the right by applying a force of 0.001N
\end{enumerate}

\paragraph{Rewards}
The agent receives a reward of -1 point for each timestep.

\paragraph{Termination criterion}
The simulation terminates after 200 timesteps.

\paragraph{Resolution criterion}
This task is considered as solved if the agent receives a mean total reward $R \geq -110$ on 100 runs.

\subsubsection{LunarLander-v2}
In this task the agent has to land a lander on a landing pad.

\paragraph{Observation space}
The state of the environment consists of 8 variables:
\begin{itemize}
    \item Horizontal position: $p_x$
    \item Vertical position: $p_y$
    \item Horizontal velocity: $v_x$
    \item Vertical velocity: $v_y$
    \item Angle w.r.t. the vertical axis: $\theta$
    \item Angular velocity: $\omega$
    \item Left leg contact: $c_{l}$
    \item Right leg contact: $c_{r}$
\end{itemize}

\paragraph{Action space}
The agent can perform 4 actions:
\begin{enumerate}
    \item All engines disabled: $nop$
    \item Enable left engine: $left$
    \item Enable main engine: $main$
    \item Enable right engine: $right$
\end{enumerate}

\paragraph{Rewards}
The reward for moving from the initial point to the landing pad with final velocity of zero varies between 100 and 140 points.
If the lander crashes it receives a reward of -100 points.
If the lander lands correctly it receives a reward of +100 points.
For each leg contact the agent receives a reward of +10 points.
Firing the main engine in a timestep gives a reward of -0.3 points, while firing a side-engine gives a reward of -0.03.

\paragraph{Termination criterion}
The simulator ends if either 1000 timesteps are passed, the lander crashes or it passes the bounds of the environment.

\paragraph{Resolution criterion}
This task is considered as solved if the mean total reward $R \geq 200$ on 100 runs.

\subsection{CartPole-v1}
\subsubsection{Experimental setup}
In this setting, we tested two different grammars: one to evolve \textit{orthogonal} decision trees and one to evolve \textit{oblique} decision trees.
Orthogonal decision trees are decision trees in which each condition tests a single variable. This results in hyperplanes that are orthogonal to the axis of the variable tested.
On the other hand, oblique decision trees handle multiple variables for each split, resulting in oblique hyperplanes.

The orthogonal and oblique grammars are shown in Tables \ref{tab:cp_ort_grammar} and \ref{tab:cp_obl_grammar} respectively.
The settings used for the grammatical evolution in the orthogonal and oblique cases are shown in Tables \ref{tab:cp_ort_params_ge} and \ref{tab:cp_obl_params_ge}.
Finally, the settings used for the Q-learning algorithm are shown in Table \ref{tab:cp_params_q}.
All the parameters have been chosen by performing a manual tuning.

\begin{table}
    \centering
    \begin{tabular}{|c|c|} \hline
        \textbf{Rule} & \textbf{Production} \\ \hline
        dt & $<if>$ \\ 
        if & $if\ <condition>\ then\ <action>\ else\ <action>$ \\ 
        condition & $input\_var\ <comp\_op>\ <const_{input\_var}>$ \\ 
        action & $leaf\ |\ <if>$ \\ 
        comp\_op & $lt\ |\ gt$ \\ 
        $const_x$ & [-4.8, 4.8) with step 0.5 \\ 
        $const_v$ & [-5, 5) with step 0.5 \\ 
        $const_\theta$ & [-0.418, 0.418) with step 0.01 \\ 
        $const_\omega$ & [-0.836, 0.836) with step 0.01 \\ \hline
    \end{tabular}
    \caption{Grammar used to evolve orthogonal decision trees in the CartPole-v1 environment. The symbol ``$\mid$" denotes the possibility to choose between different symbols. ``comp\_op'' is a short version of ``comparison operator" and ``lt" and ``gt" are respectively the ``less than" and ``greater than" operators. $input\_var$ represents one of the possible inputs in the given environment. Note that each input variable has a separate set of constants.}
    \label{tab:cp_ort_grammar}
\end{table}

\begin{table}
    \centering
    \begin{tabular}{|c|c|} \hline
        \textbf{Rule} & \textbf{Production} \\ \hline
        dt & $<if>$ \\ 
        if & $if\ <condition>\ then\ <action>\ else\ <action>$ \\ 
        condition & $lt((\sum\limits_{i=1}^{n\_variables} <const> input_i), <const>)$ \\ 
        action & $leaf\ |\ <if>$ \\ 
        $const$ & $[-1, 1]$ with step $10^{-3}$ \\ \hline
    \end{tabular}
    \caption{Grammar used to evolve oblique decision trees in the CartPole-v1 environment. The symbol ``$\mid$" denotes the possibility to choose between different symbols. ``lt" refers to the ``less than" operator.}
    \label{tab:cp_obl_grammar}
\end{table}

\begin{table}
    \centering
    \begin{tabular}{|c|c|} \hline
        \textbf{Parameter} & \textbf{Value} \\ \hline
        Population size & 200 \\ 
        Generations & 100 \\ 
        Genotype length & 1024 \\ 
        Crossover probability & 0 \\ 
        Mutation probability & 1 \\ 
        Mutation type & Uniform, with gene probability=0.1 \\ \hline
    \end{tabular}
    \caption{Parameters used for the Grammatical Evolution with orthogonal grammar in the CartPole-v1 environment.}
    \label{tab:cp_ort_params_ge}
\end{table}

\begin{table}
    \centering
    \begin{tabular}{|c|c|} \hline
        \textbf{Parameter} & \textbf{Value} \\ \hline
        Population size & 200 \\ 
        Generations & 50 \\ 
        Genotype length & 100 \\ 
        Crossover probability & 0 \\ 
        Mutation probability & 1 \\ 
        Mutation type & Uniform, with gene probability=0.1 \\ \hline
    \end{tabular}
    \caption{Parameters used for the Grammatical Evolution with oblique grammar in the CartPole-v1 environment.}
    \label{tab:cp_obl_params_ge}
\end{table}

The number of episodes used for Q-learning is quite low.
This is because, since this is a ``simple'' environment, we want to lower the computational cost of the search by \textit{exploiting} the randomness used to initialize the state-action function.
This means that, in this case, $\mathcal{Q}$-learning is used to ``fine-tune'' the state-action function instead of learning it from scratch.

\subsubsection{Results}
The results are shown in Tables \ref{tab:cp_ort_results} and \ref{tab:cp_obl_results}.
These tables show an interesting result.
In fact, while the orthogonal grammar is able to solve the task in the 100\% of the cases, the test score was the optimal one ($500 \pm 0$) only in the 40\% of the cases.
On the other hand, the oblique grammar solves the task in the 90\% of the cases, but achieves the optimal score in the 80\% of the runs.
This suggests us that, while the oblique grammar makes the search space more complex, it usually leads to more stable (as in Lyapunov's concept of stability) solutions.

\begin{table}[p]
\begin{center}
\begin{tabular}{|c|c|c|c|c|} \hline  
 \textbf{Run} & \textbf{Training mean} & \textbf{Testing mean} & \textbf{Testing std} & $\mathcal{M}$\\ \hline
 R1 & 500.00 & \textbf{498.24} & 9.49 & 53.40 \\ 
 R2 & 500.00 & \textbf{499.13} & 8.66 & 89.00 \\ 
 R3 & 500.00 & \textbf{500.00} & 0.00 & 35.60 \\ 
 R4 & 500.00 & \textbf{500.00} & 0.00 & 53.40 \\ 
 R5 & 499.10 & \textbf{497.51} & 12.89 & 53.40 \\ 
 R6 & 500.00 & \textbf{500.00} & 0.00 & 53.40 \\ 
 R7 & 498.40 & \textbf{483.05} & 62.16 & 35.60 \\ 
 R8 & 500.00 & \textbf{500.00} & 0.00 & 53.40 \\ 
 R9 & 500.00 & \textbf{499.44} & 5.57 & 35.60 \\ 
 R10 & 500.00 & \textbf{496.05} & 13.84 & 35.60 \\ \hline
\end{tabular}
\end{center}
\caption{Scores obtained by training interpretable agents on the CartPole-v1 environment by using the orthogonal grammar.}
\label{tab:cp_ort_results}
\end{table}

\begin{table}[p]
\begin{center}
\begin{tabular}{|c|c|c|c|c|} \hline  
    \textbf{Run} & \textbf{Training mean} & \textbf{Testing mean} & \textbf{Testing std} & \textbf{$\mathcal{M}$}\\ \hline
R1 & 500.00 & \textbf{500.00} & 0.00 & 24.10 \\ 
R2 & 500.00 & \textbf{495.68} & 42.98 & 24.10 \\ 
R3 & 500.00 & \textbf{500.00} & 0.00 & 48.20 \\ 
R4 & 500.00 & \textbf{500.00} & 0.00 & 24.10 \\ 
R5 & 500.00 & \textbf{500.00} & 0.00 & 24.10 \\ 
R6 & 500.00 & \textbf{500.00} & 0.00 & 24.10 \\ 
R7 & 500.00 & \textbf{500.00} & 0.00 & 24.10 \\ 
R8 & 500.00 & \textbf{500.00} & 0.00 & 24.10 \\ 
R9 & 500.00 & \textbf{500.00} & 0.00 & 24.10 \\ 
R10 & 500.00 & 460.95 & 132.43 & 24.10 \\ \hline
\end{tabular}
\end{center}
\caption{Scores obtained by training interpretable agents on the CartPole-v1 environment by using the oblique grammar.}
\label{tab:cp_obl_results}
\end{table}

\begin{figure}
    \centering
    \includegraphics[scale=0.7]{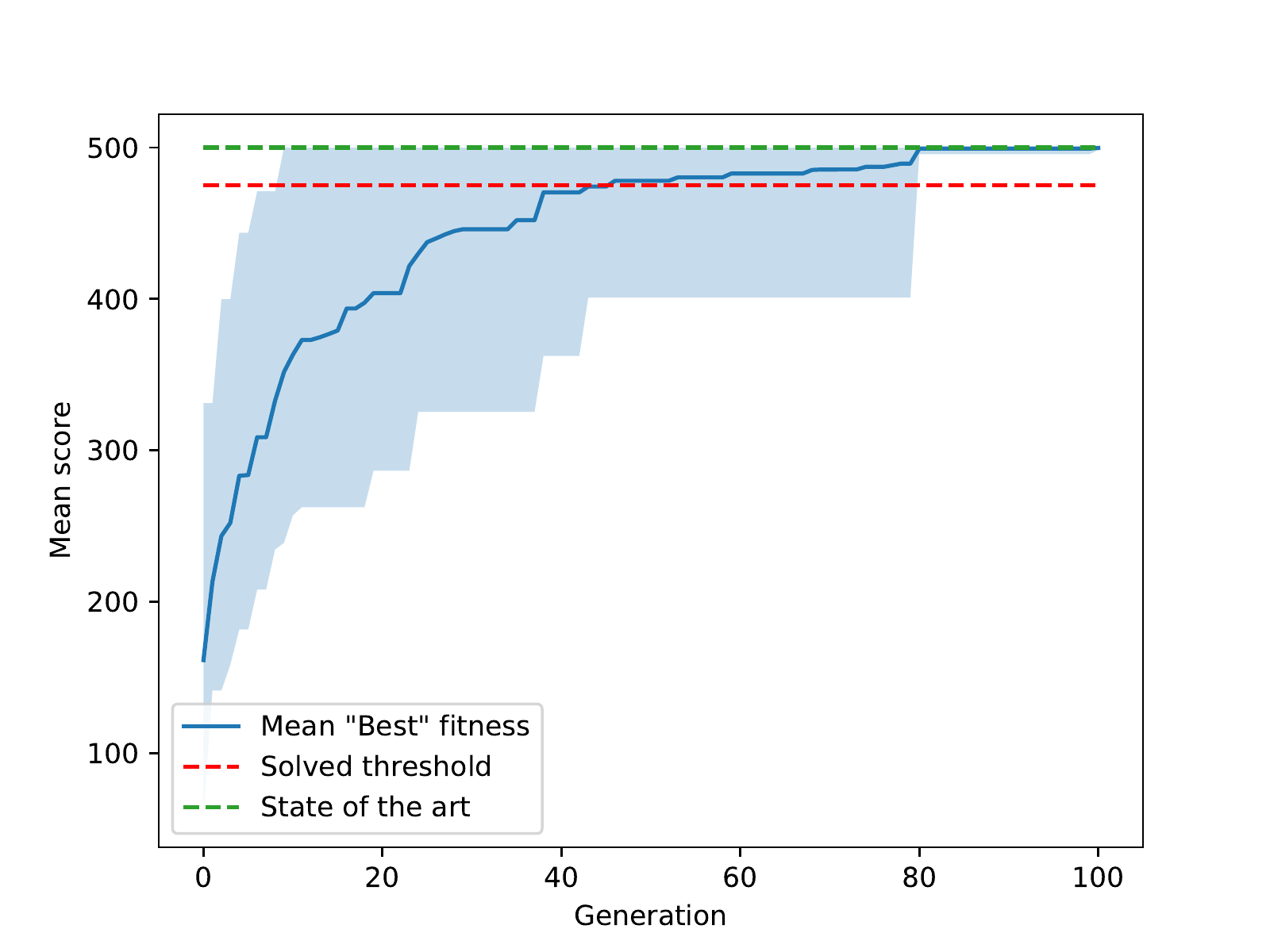}
    \caption{Fitness of the best solution on CartPole-v1, obtained by using the orthogonal grammar, at each generation averaged across 10 runs.}
    \label{fig:cp_ort_fitnesstrend}
\end{figure}

\begin{figure}
    \centering
    \includegraphics[scale=0.7]{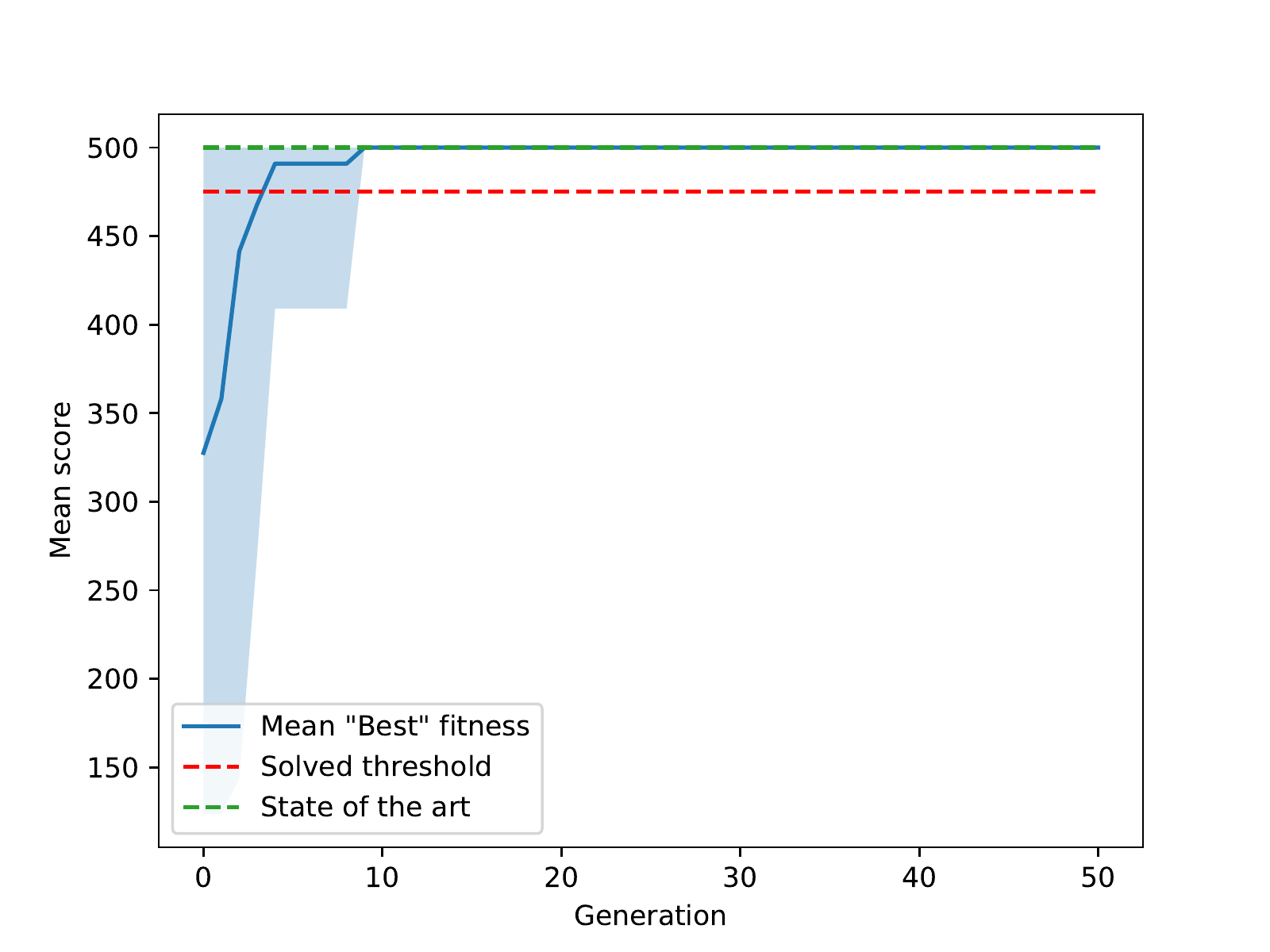}
    \caption{Fitness of the best solution on CartPole-v1, obtained by using the oblique grammar, at each generation averaged across 10 runs.}
    \label{fig:cp_obl_fitnesstrend}
\end{figure}

\begin{figure}
    \centering
    \includegraphics[scale=0.7]{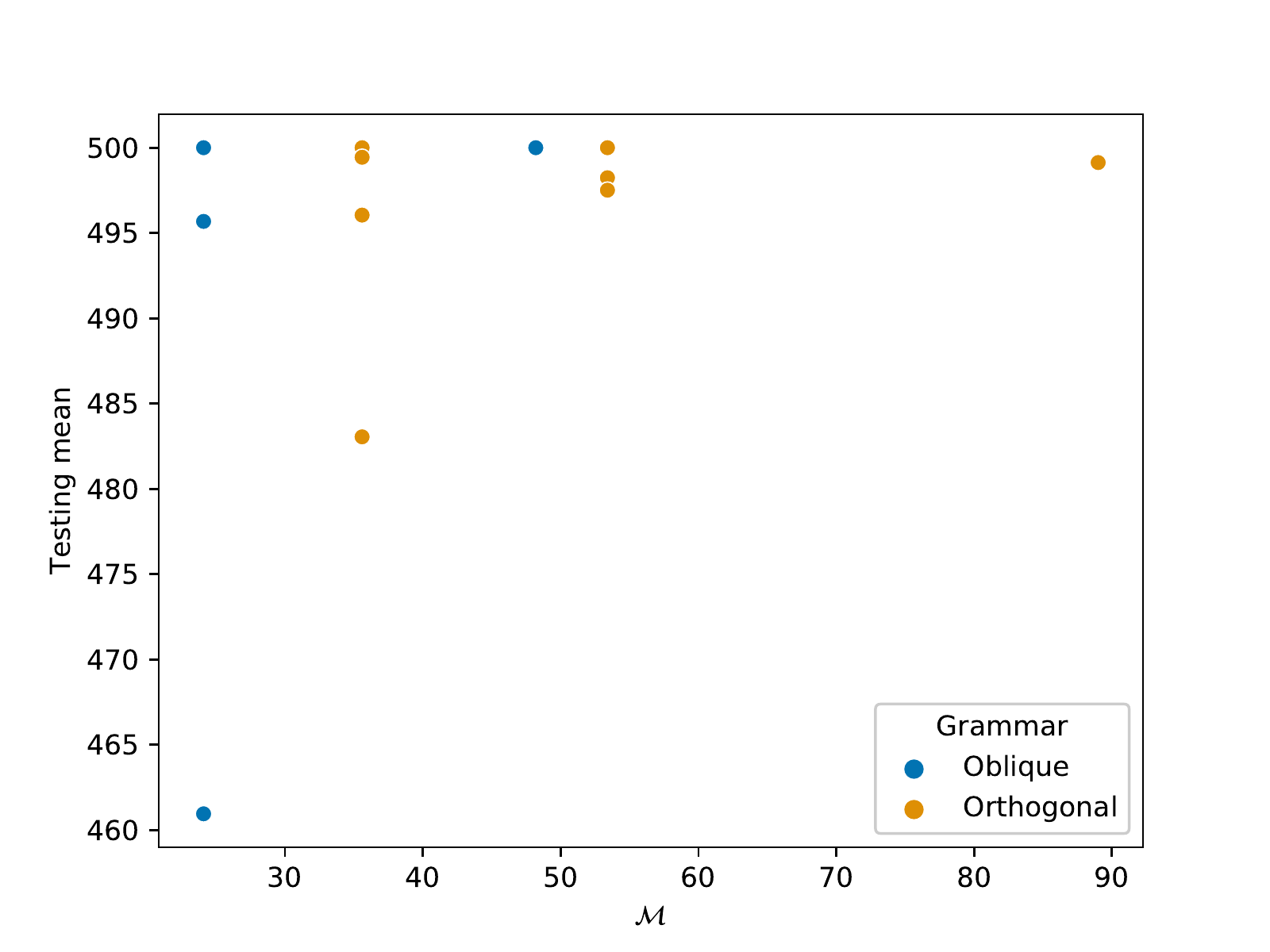}
    \caption{A score-$\mathcal{M}$ plot of the solutions obtained by using the two types of grammars.}
    \label{fig:cp_grammar_comparison}
\end{figure}

In Figures \ref{fig:cp_ort_fitnesstrend} and \ref{fig:cp_obl_fitnesstrend} we show the mean best fitness for each generation averaged across all the runs.
We observe that, while both settings are able to converge towards the optimal fitness, the oblique grammar converges more quickly than the orthogonal one to the maximum fitness.

In Figure \ref{fig:cp_grammar_comparison}, a comparison of the solutions obtained by using the orthogonal and oblique grammars is shown.
We can easily observe that, usually, the solutions obtained with the oblique grammar have a lower $\mathcal{M}$ than the ones obtained by using the orthogonal grammar.
This is due to the fact that most of the produced oblique trees use only one split, resulting in a lower number of non-arithmetical operations.

To better assess the hypothesis made earlier, i.e. that oblique trees are more stable than orthogonal ones, in Tables \ref{tab:cp_ort_10000steps} and \ref{tab:cp_obl_10000steps} we compare all the trees produced by using the two grammars on a modified environment that has a maximum duration of $10^4$ timesteps instead of 500.
These results confirm our hypothesis, showing that all the oblique trees are able to obtain significantly better scores, often obtaining a perfect score (i.e. $10^4 \pm 0$) also in this setting.
Figure \ref{fig:cp_timesteps_comparison} shows how the testing mean score varies by varying the number of maximum timesteps for the best agents.

\begin{figure}[p]
    \centering
    \includegraphics[scale=0.69]{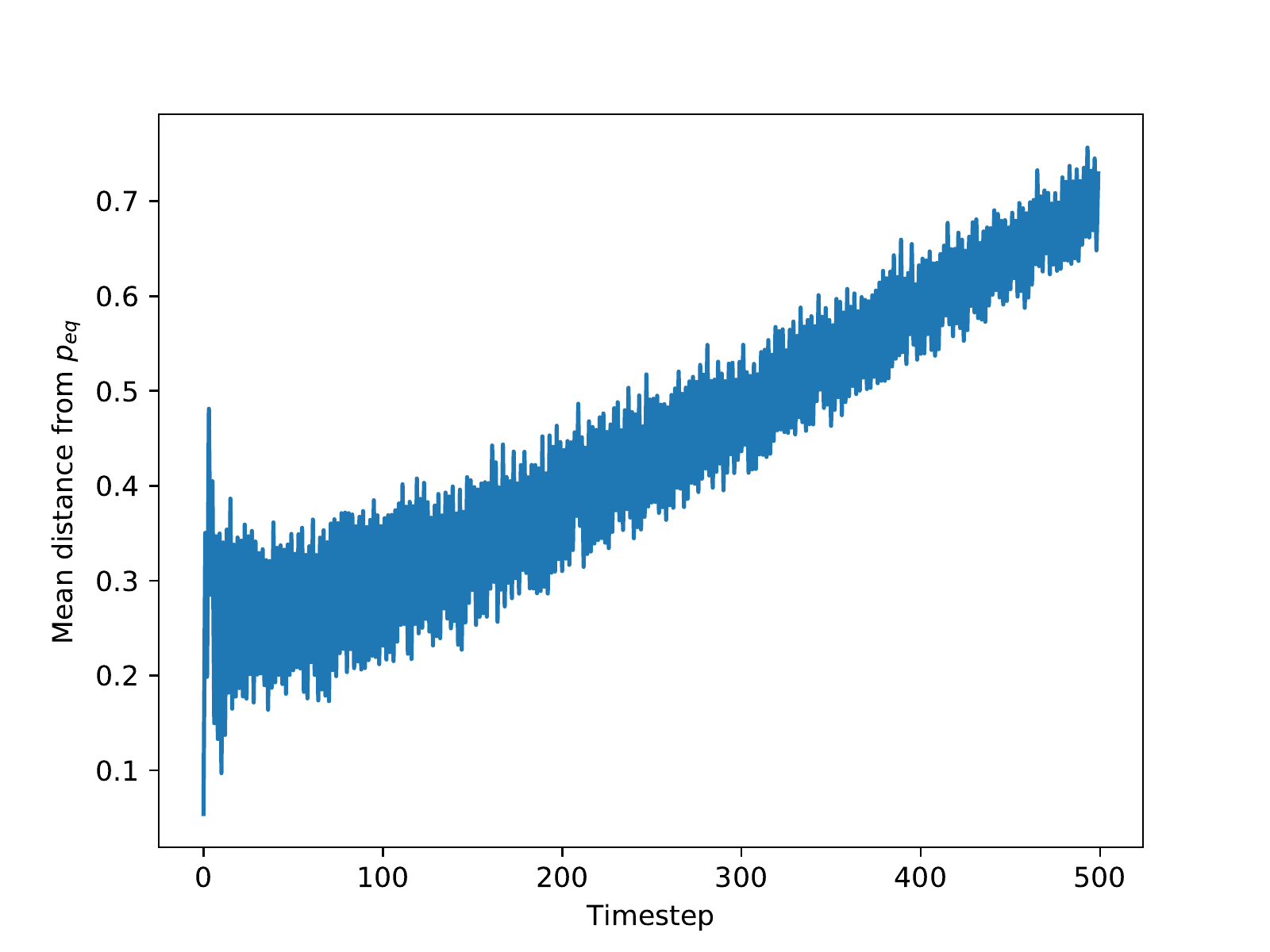}
    \caption{Mean distance of the cart-pole system from the point of equilibrium when using the best orthogonal tree as policy.}
    \label{fig:cp_ort_stability}
\end{figure}

\begin{figure}
    \centering
    \includegraphics[scale=0.69]{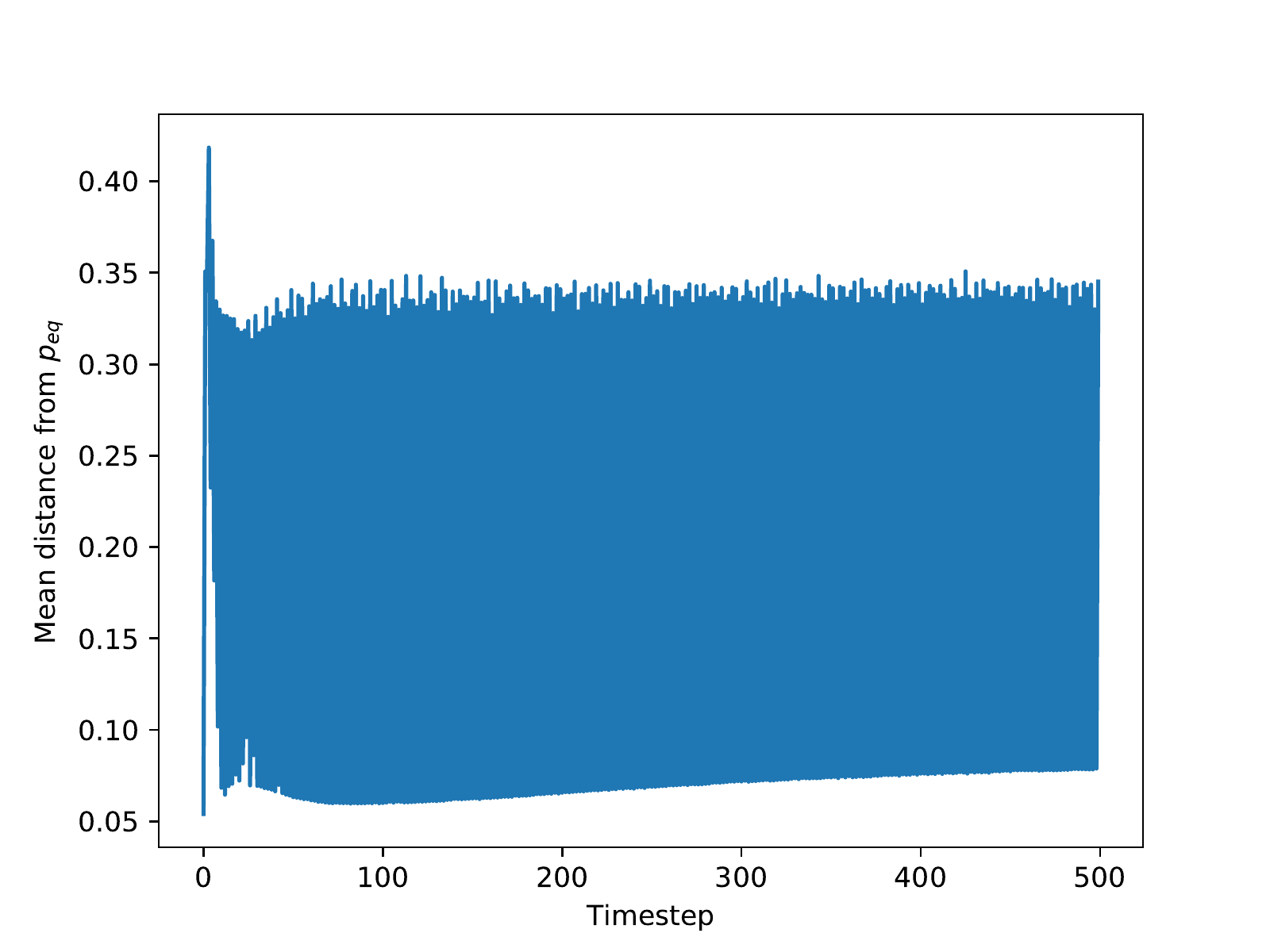}
    \caption{Mean distance of the cart-pole system from the point of equilibrium when using the best oblique tree as policy.}
    \label{fig:cp_obl_stability}
\end{figure}

\begin{figure}
    \centering
    \includegraphics[scale=0.69]{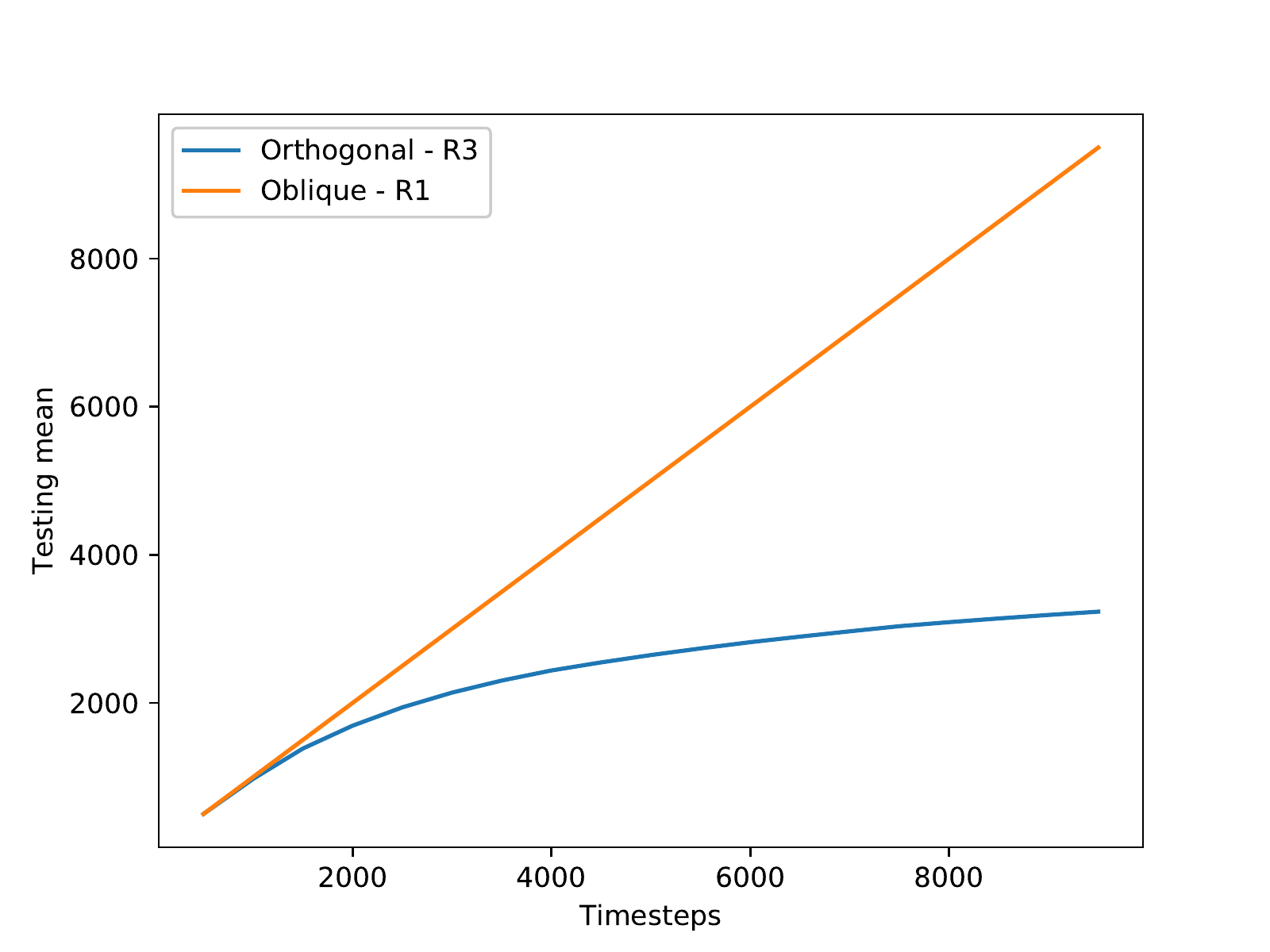}
    \caption{Comparison between the best orthogonal tree with the best oblique at different maximum timesteps for the CartPole-v1 environment}
    \label{fig:cp_timesteps_comparison}
\end{figure}

In Figures \ref{fig:cp_ort_stability} and \ref{fig:cp_obl_stability} we show the mean distance from the point of equilibrium ($p_{eq} = [0, 0, 0, 0]^T$) averaged over 100 episodes (of length 500 timesteps).
In these figures we can easily observe that the oblique policy seems to be stable (according to the Lyapunov's concept of stability) while the orthogonal policy does not.

\begin{table}[ht!]
    \centering
    \begin{tabular}{|c|c|c|} \hline
        \textbf{Run} & \textbf{Testing mean} & \textbf{Testing std} \\ \hline
        R1 & 878.08 & 346.85 \\ 
        R2 & 767.94 & 202.24 \\ 
        R3 & 3271.99 & 2718.79 \\ 
        R4 & 5845.54 & 2898.37 \\ 
        R5 & 1237.18 & 775.238 \\ 
        R6 & 2589.85 & 2715.03 \\ 
        R7 & 4561.71 & 3670.16 \\ 
        R8 & 5738.87 & 3227.96 \\ 
        R9 & 1179.21 & 543.78 \\ 
        R10 & 688.75 & 183.64 \\ \hline
    \end{tabular}
    \caption{Scores obtained by testing the solutions obtained by using an orthogonal grammar on a $10^4$-steps-long version of CartPole-v1}
    \label{tab:cp_ort_10000steps}
\end{table}

\begin{table}[ht!]
    \centering
    \begin{tabular}{|c|c|c|} \hline
        \textbf{Run} & \textbf{Testing mean} & \textbf{Testing std} \\ \hline
        R1 & 10000.00 & 0.00 \\ 
        R2 & 9900.68 & 988.22 \\ 
        R3 & 10000.00 & 0.00 \\ 
        R4 & 10000.00 & 0.00 \\ 
        R5 & 10000.00 & 0.00 \\ 
        R6 & 10000.00 & 0.00 \\ 
        R7 & 10000.00 & 0.00 \\ 
        R8 & 10000.00 & 0.00 \\ 
        R9 & 10000.00 & 0.00 \\ 
        R10 & 9200.95 & 2709.71 \\ \hline
    \end{tabular}
    \caption{Scores obtained by testing the solutions obtained by using an oblique grammar on a $10^4$-steps-long version of CartPole-v1}
    \label{tab:cp_obl_10000steps}
\end{table}

\begin{figure}
    \centering
    \includegraphics[scale=0.69]{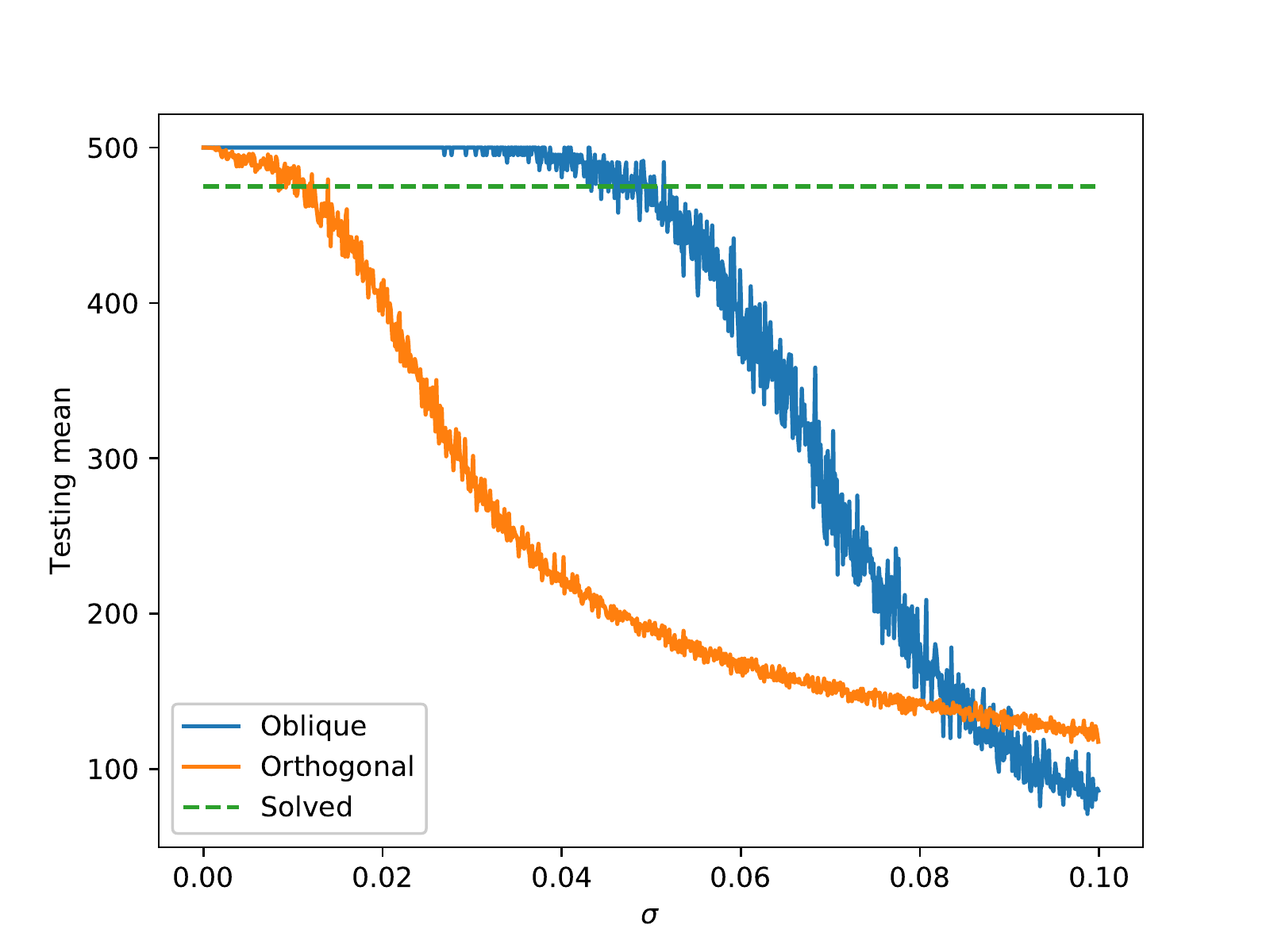}
    \caption{Performance of the two best agents on CartPole-v1 at the variation of the input noise.}
    \label{fig:cp_noise_comparison}
\end{figure}

Moreover, we also tested the robustness of the produced agents with respect to noise on the inputs received by the sensors.
In Figure \ref{fig:cp_noise_comparison} we show how the performance of the two best agents vary with respect to additive input noise ( distributed as $\mathcal{N}(0,\sigma^2)$).
The orthogonal tree was robust to noises with $\sigma$ in the order of twice the sampling step used for the constants.
On the other hand, the oblique tree proved to be significantly more robust, being able to cope with noises that have a $\sigma$ about 50 times bigger than the sampling step used for the constants.

\begin{figure}[ht!]
    \centering
    \includegraphics[scale=0.51]{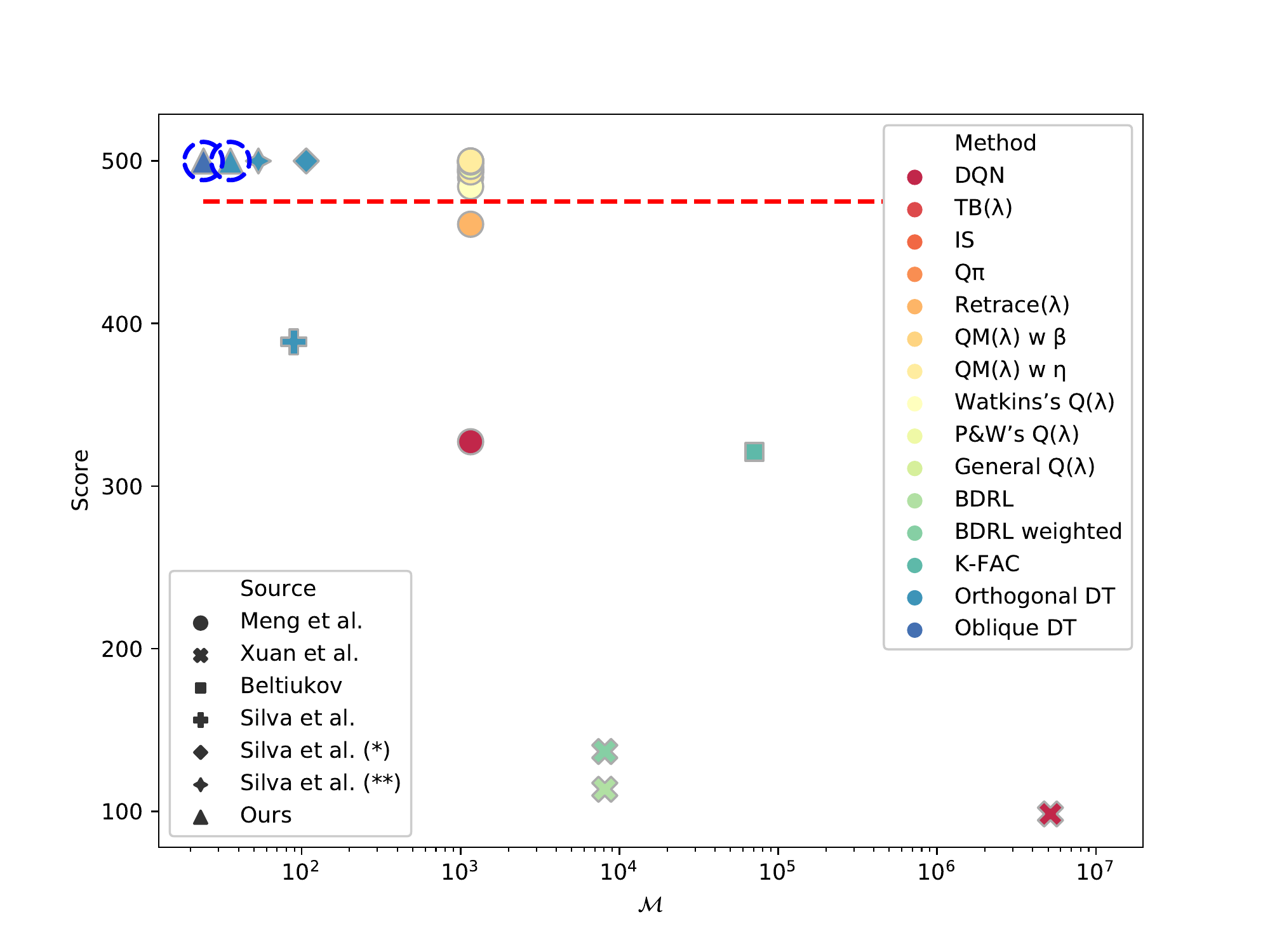}
    \caption{Score-$\mathcal{M}$ comparison of our solutions and the state-of-the-art in the CartPole-v1 environment. It is compared to the ones described in \cite{meng_qualitative_2019}, \cite{xuan_bayesian_2018} \cite{silva_optimization_2020}, and \cite{van_der_aalst_optimizing_2020}.
    (*): Result confirmed by personal communication with the first author of the study.
    (**): The tree has been simplified by using the technique used in our work.
    }
    \label{fig:cp_sota_comparison}
\end{figure}

Finally, in Table \ref{tab:cp_sota_comparison} and Figure \ref{fig:cp_sota_comparison} we compare our best solutions with other solutions found in literature.
The complexities computed for the neural-network based approaches are approximations, i.e. we did not take into account all the details of the methods but only the network architectures, resulting in a slightly lower complexity. In our opinion, for the purpose of comparing our solution with the non-interpretable state-of-the-art, these small differences are negligible.
Our solutions that have been used for the comparison are shown in Figures \ref{fig:cp_ort_best} and \ref{fig:cp_obl_best}.
The other produced solutions can be found in the repository of the project\footnote{https://gitlab.com/leocus/ge\_q\_dts, accessed: 11 dec 2020.}.

\begin{table}[p]
    \begin{tabularx}{1\linewidth}{|X|l|l|} \hline
    \textbf{Method}                  & \textbf{Score} & $\mathcal{M}$  \\ \hline
    Deep Q Network {\cite{meng_qualitative_2019}}            & 327.30 & 1157.20 \\
    Tree-Backup($\lambda$)\ {\cite{meng_qualitative_2019}}          & 494.70 & 1157.20   \\
    Importance-Sampling {\cite{meng_qualitative_2019}}             & 498.70 & 1157.20     \\
    Q$\pi$ {\cite{meng_qualitative_2019}}             & 489.90 & 1157.20       \\
    Retrace($\lambda$)\ {\cite{meng_qualitative_2019}}     & 461.10 & 1157.20         \\
    Qualitatively measured policy discrepancy\  w/ \ $\beta$\ {\cite{meng_qualitative_2019}}      & 499.90 & 1157.20           \\
    Qualitatively measured policy discrepancy\  w/ \ $\eta$\ {\cite{meng_qualitative_2019}}      & 493.20 & 1157.20             \\
    Watkins’s Q($\lambda$)\ {\cite{meng_qualitative_2019}} & 484.30 & 1157.20               \\
    Qualitatively measured policy discrepancy\  w/ \ $\beta$\ {\cite{meng_qualitative_2019}}      & 494.90 & 1157.20\\
    Qualitatively measured policy discrepancy\  w/ \ $\eta$\ {\cite{meng_qualitative_2019}}      & 493.30 & 1157.20\\
    Peng \& Williams’s Q($\lambda$)\ {\cite{meng_qualitative_2019}}    & 496.70 & 1157.20\\
    Qualitatively measured policy discrepancy\  w/ \ $\beta$\ {\cite{meng_qualitative_2019}}      & \textbf{500.00}  & 1157.20\\
    Qualitatively measured policy discrepancy\  w/ \ $\eta$\ {\cite{meng_qualitative_2019}}      & 499.40 & 1157.20\\
    General Q($\lambda$)\ {\cite{meng_qualitative_2019}}   & 499.90 & 1157.20\\
    Qualitatively measured policy discrepancy\  w/ \ $\beta$\ {\cite{meng_qualitative_2019}}      & \textbf{500.00}  & 1157.20\\
    Qualitatively measured policy discrepancy\  w/ \ $\eta$\ {\cite{meng_qualitative_2019}}      & \textbf{500.00}  & 1157.20\\
    Deep Q Network {\cite{xuan_bayesian_2018}}            & 98.33 & 5170174.80 \\
    Bayesian Deep Reinforcement Learning {\cite{xuan_bayesian_2018}}           & 113.52 & 8090.40 \\
    Bayesian Deep Reinforcement Learning weighted {\cite{xuan_bayesian_2018}}  & 136.75 & 8090.40 \\
    Kronecker-Factored Approximate Curvature {\cite{van_der_aalst_optimizing_2020}} & 321.00 & 70786.20 \\
    Differentiable Decision Trees {\cite{silva_optimization_2020}} & 388.76 & {89.20} \\ 
    Differentiable Decision Trees {\cite{silva_optimization_2020}} (*) & \textbf{500.00} & 106.80 \\
    Differentiable Decision Trees {\cite{silva_optimization_2020}} (**) & \textbf{500.00} & \textbf{53.40} \\ \hline
    Ours – Orthogonal & \textbf{500.00} & \textbf{35.60}  \\
    Ours – Oblique & \textbf{500.00} & \textbf{24.10}  \\ \hline 
    \end{tabularx}

    \caption{Comparison of the solutions obtained by using the proposed approach with respect to the state-of-the-art.
    The results from \cite{meng_qualitative_2019} are averaged over ten independent runs.
    The results from \cite{silva_optimization_2020} regard the discretized tree shown in Figure \ref{fig:cp_tree_silva} (From Figure 3 - right in \cite{silva_optimization_2020}) tested on the same episodes used for the evaluation of our solutions.
    \\
    (*): Result confirmed by personal communication with the first author of the study.
    (**): The tree has been simplified by using the technique used in our work.
    }
    \label{tab:cp_sota_comparison}
\end{table}

\begin{center}
    \begin{figure}[ht!]
        \centering
        \begin{tikzpicture}[auto]
            \node [box]                                         (root)  {$\omega < 0.074$};
            \node [box, below=0.cm of root, xshift=-3cm]        (t)     {$\theta < 0.022$};
            \node [leaf, below=0.5cm of t, xshift=-2cm]         (tt)    {move\_left};
            \node [leaf, below=0.5cm of t, xshift=2cm]          (tf)    {move\_right};
            \node [leaf, below=0.5cm of root, xshift=3cm]       (f)     {move\_right};
            
            \draw (root) -| (t) node [midway, above] (TextNode) {True};
            \draw (root) -| (f) node [midway, above] (TextNode) {False};
            \draw (t) -| (tt) node [midway, above] (TextNode) {True};
            \draw (t) -| (tf) node [midway, above] (TextNode) {False};
        \end{tikzpicture}
    \caption{Tree representation of one of the best individuals evolved in the CartPole-v1 environment by using the orthogonal grammar.}
    \label{fig:cp_ort_best}
    \end{figure}
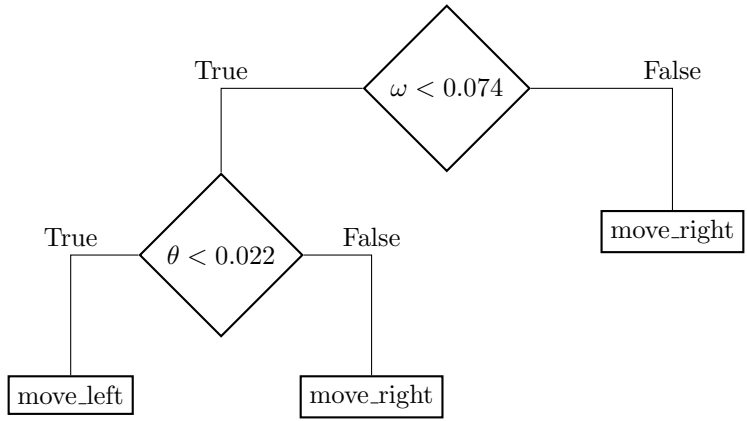
\end{center}

\begin{center}
    \begin{figure}[ht!]
        \centering
        \begin{tikzpicture}[auto]
            \node [box]                                         (root)  {\begin{tabular}{c}$-0.274x-0.543v+$\\$-0.904\theta-0.559\omega$\\$ < -0.169$\end{tabular}};
            \node [leaf, below=0.5cm of root, xshift=-3cm]       (t)     {move\_right};
            \node [leaf, below=0.5cm of root, xshift=3cm]       (f)     {move\_left};
            
            \draw (root) -| (t) node [midway, above] (textnode) {true};
            \draw (root) -| (f) node [midway, above] (textnode) {false};
        \end{tikzpicture}
    \caption{Tree representation of one of the best individuals evolved in the CartPole-v1 environment by using the orthogonal grammar.}
    \label{fig:cp_obl_best}
    \end{figure}
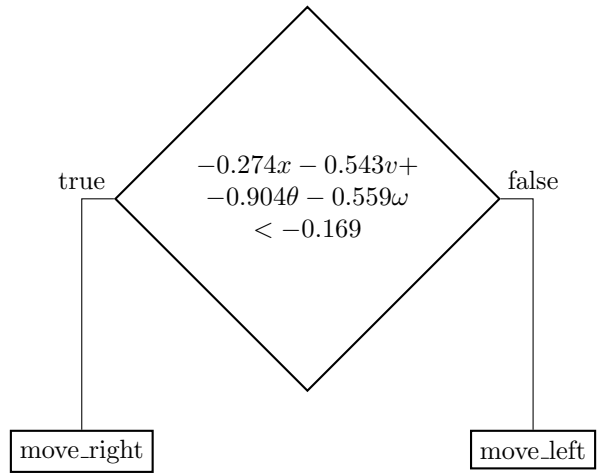
\end{center}

\subsection{MountainCar}
\subsubsection{Experimental setup}
Also in this task, we tested both an orthogonal and an oblique grammar.
The two grammars are shown in Tables \ref{tab:mc_ort_grammar} and \ref{tab:mc_obl_grammar}. 
Note that, in this environment we normalize the variables in the oblique case whereas in the others we do not.
This is because the ranges of variation of the two variables are quite different. 
Moreover, a preliminary experimental phase confirmed that it was hard to obtain good results by not normalizing the inputs.

The parameters used for the Grammatical Evolution are shown in Tables \ref{tab:mc_ort_params_ge} and \ref{tab:mc_obl_params_ge}.
The settings for the Q-learning algorithm are shown in Tables \ref{tab:mc_ort_params_q} and \ref{tab:mc_obl_params_q}.
Also in this case, we set the number of episodes to 10 to exploit the randomness of the initialization, since this is considered as a ``simple'' task.

\begin{table}
    \centering
    \begin{tabular}{|c|c|} \hline
        \textbf{Rule} & \textbf{Production} \\ \hline
        dt & $<if>$ \\ 
        if & $if\ <condition>\ then\ <action>\ else\ <action>$ \\ 
        condition & $input\_var\ <comp_op>\ <const_{input\_var}>$ \\ 
        action & $leaf\ |\ <if>$ \\ 
        comp\_op & $lt\ |\ gt$ \\ 
        $const_x$ & [-1.2, 0.6) with step 0.05 \\ 
        $const_v$ & [-0.07, 0.07) with step 0.005 \\ \hline
    \end{tabular}
    \caption{Grammar used to evolve orthogonal decision trees in the MountainCar-v0 environment. The symbol ``$\mid$" denotes the possibility to choose between different symbols. ``comp\_op'' is a short version of ``comparison operator" and ``lt" and ``gt" are respectively the ``less than" and ``greater than" operators. $input\_var$ represents one of the possible inputs in the given environment. Note that each input variable has a separate set of constants.}
    \label{tab:mc_ort_grammar}
\end{table}

\begin{table}
    \centering
    \begin{tabular}{|c|c|} \hline
        \textbf{Rule} & \textbf{Production} \\ \hline
        dt & $<if>$ \\ 
        if & $if\ <condition>\ then\ <action>\ else\ <action>$ \\ 
        condition & $lt((\sum\limits_{i=1}^{n\_variables} <const> \widehat{input_i}, <const>)$ \\ 
        action & $leaf\ |\ <if>$ \\ 
        $const$ & $[-1, 1]$ with step $10^{-3}$ \\ \hline
    \end{tabular}
    \caption{Grammar used to evolve oblique decision trees in the MountainCar-v0 environment. The symbol ``$\mid$" denotes the possibility to choose between different symbols. ``lt" refers to the ``less than" operator. 
    $\widehat{input_i}$ refers to the normalized $input_i$ variable. For the normalization, the bounds [-1.2, 0.7] and [-0.07, 0.07] were used.}
    \label{tab:mc_obl_grammar}
\end{table}

\begin{table}
    \centering
    \begin{tabular}{|c|c|} \hline
        \textbf{Parameter} & \textbf{Value} \\ \hline
        Population size & 200 \\ 
        Generations & 1000 \\ 
        Genotype length & 1024 \\ 
        Crossover probability & 0 \\ 
        Mutation probability & 1 \\ 
        Mutation type & Uniform, with gene probability=0.05 \\ \hline
    \end{tabular}
    \caption{Parameters used for the Grammatical Evolution with orthogonal grammar in the MountainCar-v0 environment.}
    \label{tab:mc_ort_params_ge}
\end{table}

\begin{table}
    \centering
    \begin{tabular}{|c|c|} \hline
        \textbf{Parameter} & \textbf{Value} \\ \hline
        Population size & 200 \\ 
        Generations & 2000 \\ 
        Genotype length & 100 \\ 
        Crossover probability & 0.1 \\ 
        Crossover operator & One-point crossover \\ 
        Selection operator & Tournament selection with size 2 \\ 
        Mutation probability & 1 \\ 
        Mutation type & Uniform, with gene probability=0.1 \\ \hline
    \end{tabular}
    \caption{Parameters used for the Grammatical Evolution with oblique grammar in the MountainCar-v0 environment.}
    \label{tab:mc_obl_params_ge}
\end{table}

\begin{table}
    \centering
    \begin{tabular}{|c|c|} \hline
        \textbf{Parameter} & \textbf{Value} \\ \hline
        Algorithm & $\varepsilon$-greedy Q-learning \\ 
        $\varepsilon$ & 0.05 \\ 
        Initialization strategy& Uniform $\in [-1, 1]$ \\ 
        Learning rate & 0.001 \\ 
        Number of episodes & 10 \\ \hline
    \end{tabular}
    \caption{Parameters used for the Q-learning algorithm in the CartPole-v1 and MountainCar-v0 (only with orthogonal trees) environments.}
    \label{tab:cp_params_q}
    \label{tab:mc_ort_params_q}
\end{table}

\begin{table}
    \centering
    \begin{tabular}{|c|c|} \hline
        \textbf{Parameter} & \textbf{Value} \\ \hline
        Algorithm & $\varepsilon$-greedy Q-learning \\ 
        $\varepsilon$ & 0.01 \\ 
        Initialization strategy& Uniform $\in [-1, 1]$ \\ 
        Learning rate & 0.001 \\ 
        Number of episodes & 10 \\ \hline
    \end{tabular}
    \caption{Parameters used for the Q-learning algorithm in the MountainCar-v0 environment when evolving oblique trees.}
    \label{tab:mc_obl_params_q}
\end{table}

\subsubsection{Results}
The results obtained by the best solution for each run are shown in Tables \ref{tab:mc_ort_results} and \ref{tab:mc_obl_results}.
In Table \ref{tab:mc_obl_results} there are some values in parenthesis. 
This is because, given the difference in performance between training and testing scores, we proceeded with further investigation of the results.
We deduced that in the latest steps of the training of such agents a change happened in the Q-values of a leaf.
This change made the Q-values of the action taken with the current greedy policy have a value approximately equal to the another action.
This caused a destructive change in the policy, so, in order to give more information, we included the test score of the solution by reverting the destructive change.
Moreover, this change has only been used in this table.
For the remainder of this work, we will assume that their test score is $-200$.
\begin{table}
    \centering
    \begin{tabular}{|c|c|c|c|c|} \hline  
        \textbf{Run} & \textbf{Training score} & \textbf{Testing mean} & \textbf{Testing std} & \textbf{$\mathcal{M}$} \\ \hline
        R1  & -109.3           & \textbf{-106.17}      & 4.69        & 89            \\
        R2  & -110.5           & \textbf{-108.62}      & 16.72       & 124.6         \\
        R3  & -105.6           & \textbf{-102.26}      & 9.51        & 71.2          \\
        R4  & -108.1           & \textbf{-101.72}      & 3.14        & 106.8         \\
        R5  & -112.9           & -116.15      & 1.03        & 71.2          \\
        R6  & -107.2           & \textbf{-101.72}      & 3.14        & 106.8         \\
        R7  & -120.5           & -117.84      & 0.95        & 35.6          \\
        R8  & -115.7           & -115.51      & 1.18        & 35.6          \\
        R9  & -109.3           & \textbf{-106.63}      & 4.68        & 89            \\
        R10  & -107.1           & \textbf{-104.94}      & 3.56        & 53.4          \\ \hline
    \end{tabular}
    \caption{Scores obtained by training interpretable agents on the MountainCar-v0 environment when using the orthogonal grammar.}
    \label{tab:mc_ort_results}
\end{table}

\begin{table}
    \centering
    \begin{tabular}{|c|c|c|c|c|} \hline
        \textbf{Run} & \textbf{Training score} & \textbf{Testing mean} & \textbf{Testing std} & \textbf{$\mathcal{M}$} \\ \hline
        R1 & -108.90 & \textbf{-106.66} & 9.30 & 70.00 \\ 
        R2 & -106.50 & -110.18 & 24.90 & 46.60 \\ 
        R3 & -105.60 & \textbf{-106.50} & 15.27 & 23.40 \\ 
        R4 & -109.10 & -200.00 (-112.62) & 0.00 (23.08) & 0.00 (46.6) \\ 
        R5 & -106.10 & \textbf{-106.06} & 12 & 46.80 \\ 
        R6 & -110.40 & -116.66 & 16.01 & 46.60 \\ 
        R7 & -112.80 & -114.44 & 10.74 & 46.40 \\ 
        R8 & -105.00 & -200.00 (\textbf{-107.5}) & 0.00 (13.46) & 0.00 (23.4) \\ 
        R9 & -103.20 & \textbf{-106.02} & 15.41 & 46.80 \\ 
        R10 & -111.40 & -116.49 & 16.75 & 46.80 \\ \hline
    \end{tabular}
    \caption{Scores obtained by training interpretable agents on the MountainCar-v0 environment when using the oblique grammar.}
    \label{tab:mc_obl_results}
\end{table}

As we can see from \ref{tab:mc_ort_results}, the solutions obtained by using the orthogonal grammar solve the task in the 70\% of the cases.
On the other hand, as we can see from Table \ref{tab:mc_obl_results}, oblique trees perform poorly on this problem.
This suggests us that this problem is harder to solve by using oblique trees than orthogonal ones.
While this may seem counter-intuitive, since oblique trees are a generalization of orthogonal trees, it may be because our grammar (the one used to produce oblique trees) makes it difficult to obtain an orthogonal decision tree.

\begin{figure}[p]
    \centering
    \includegraphics[scale=0.69]{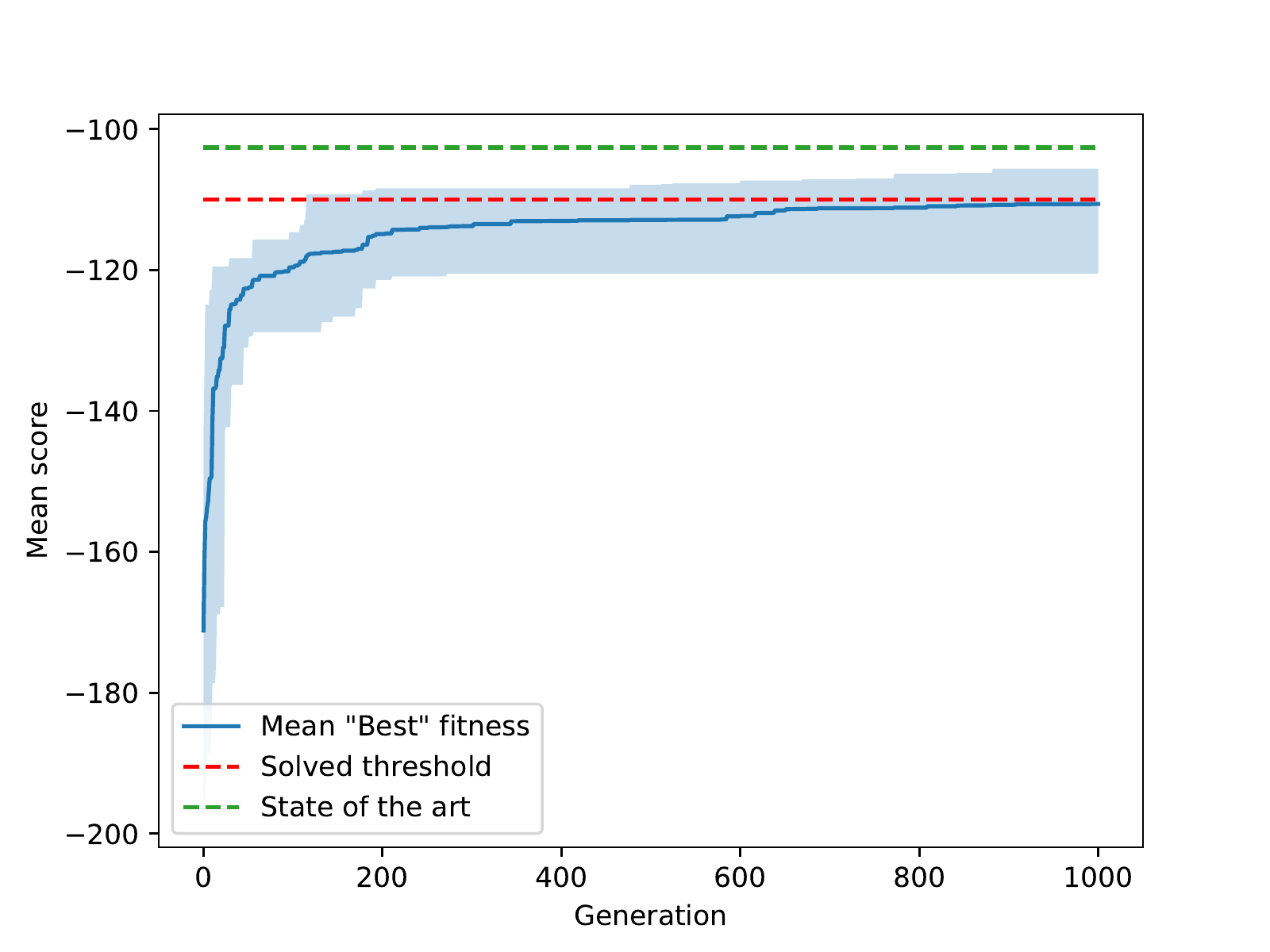}
    \caption{Fitness trend for the best individual in the MountainCar-v0 environment averaged on all the runs, when using orthogonal trees.}
    \label{fig:mc_ort_fitnesstrend}
\end{figure}

\begin{figure}[p]
    \centering
    \includegraphics[scale=0.69]{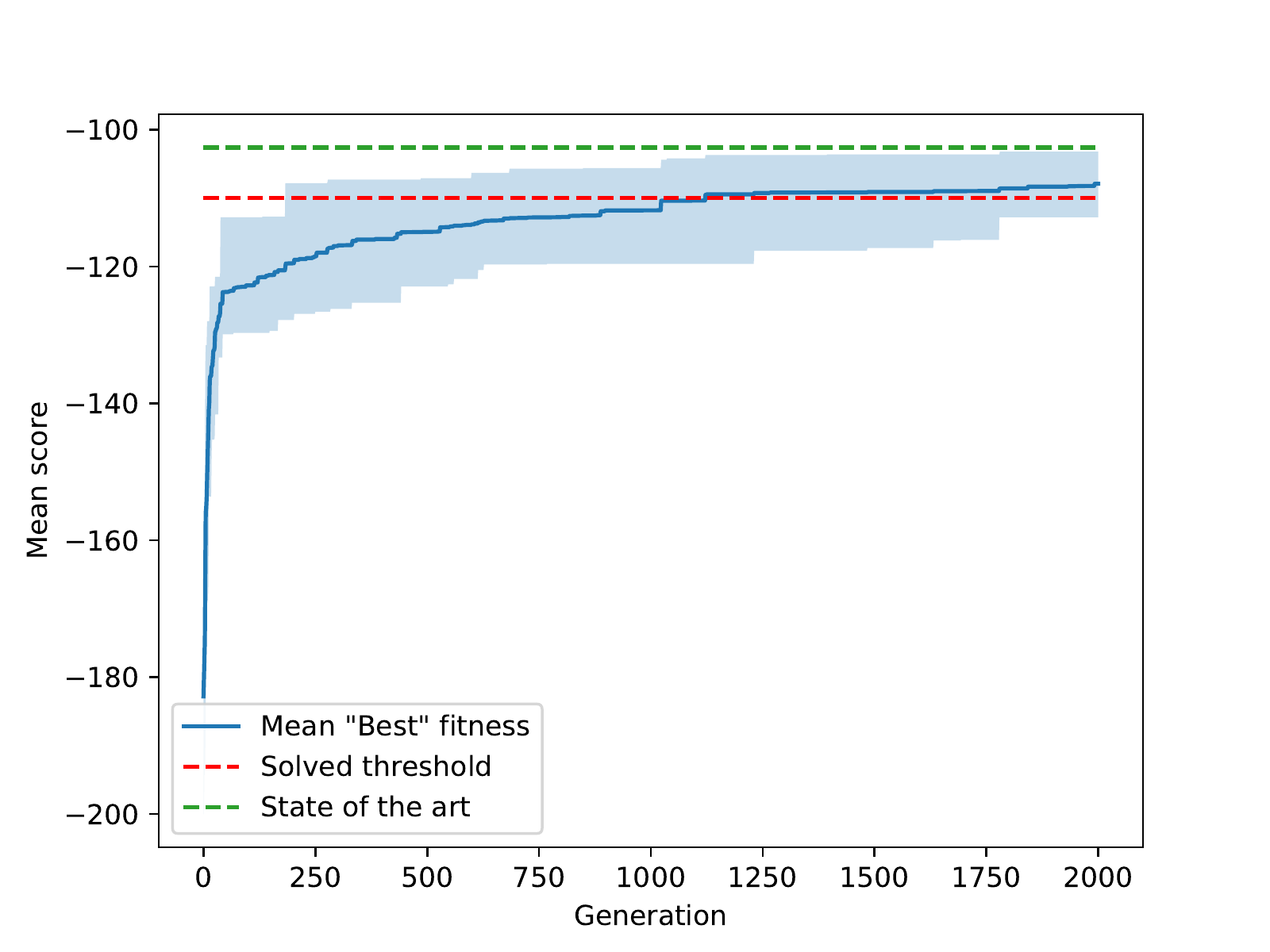}
    \caption{Fitness trend for the best individual in the MountainCar-v0 environment averaged on all the runs, when using oblique trees.}
    \label{fig:mc_obl_fitnesstrend}
\end{figure}

The fitness trend for the best individual averaged on each run are shown in Figure \ref{fig:mc_ort_fitnesstrend} and \ref{fig:mc_obl_fitnesstrend} for the orthogonal and oblique cases, respectively.

\begin{figure}
    \centering
    \includegraphics[scale=0.69]{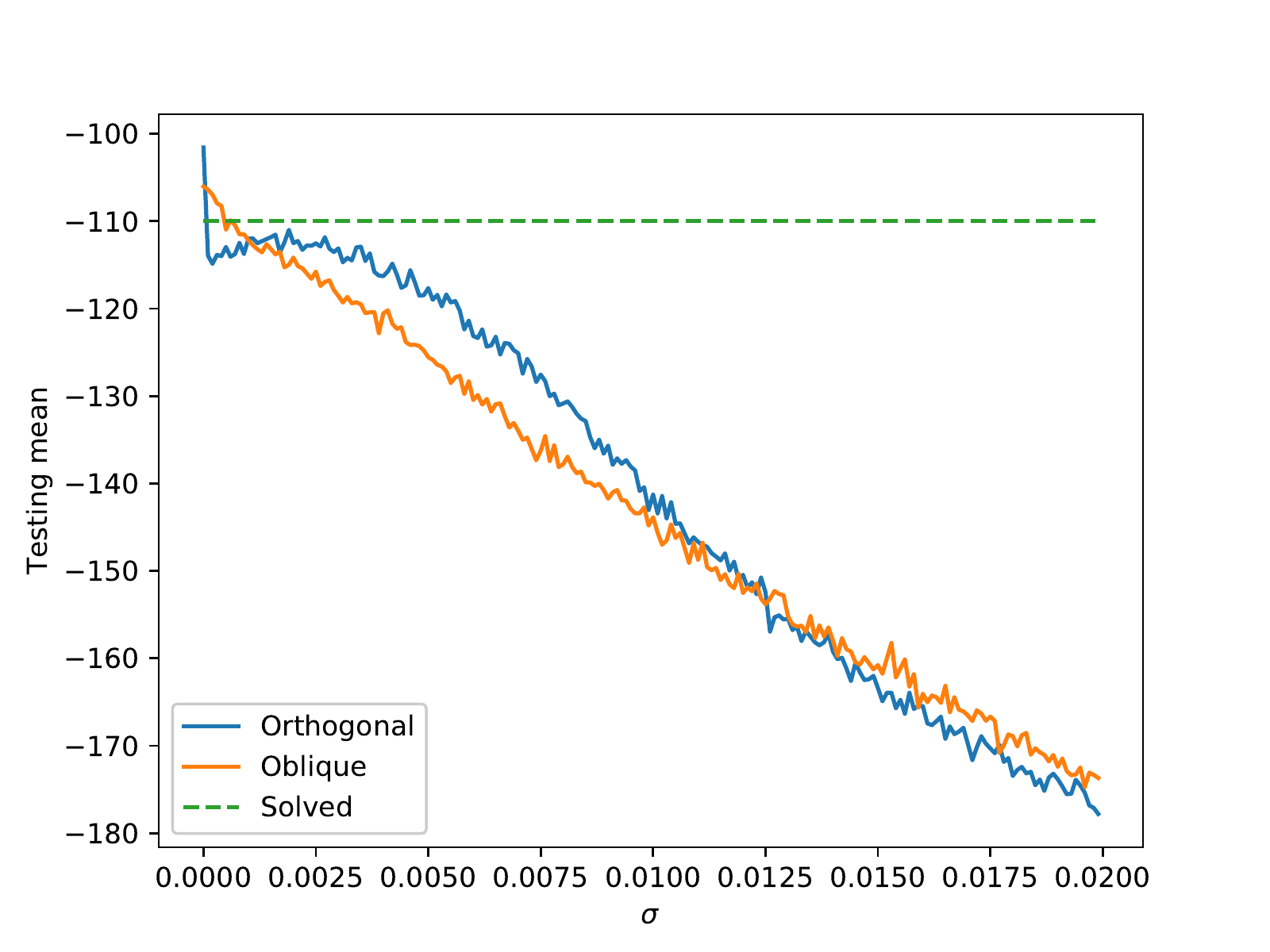}
    \caption{Plot of the mean testing score with different input noises for the best orthogonal and oblique models on MountainCar-v0.}
    \label{fig:mc_noise_comparison}
\end{figure}

To compare the two approaches, we compare the robustness to input noise for both versions.
The result is shown in Figure \ref{fig:mc_noise_comparison}.
In this case both approaches proved to be not so robust to noise.
Surprisingly, we can observe that the orthogonal tree was not even robust to input noise that had 
$\sigma < \underset{i}{min}(step_i)$
where $step_i$ is the sampling step for the constants of the $i$-th variable.

\begin{figure}
    \centering
    \includegraphics[scale=0.55]{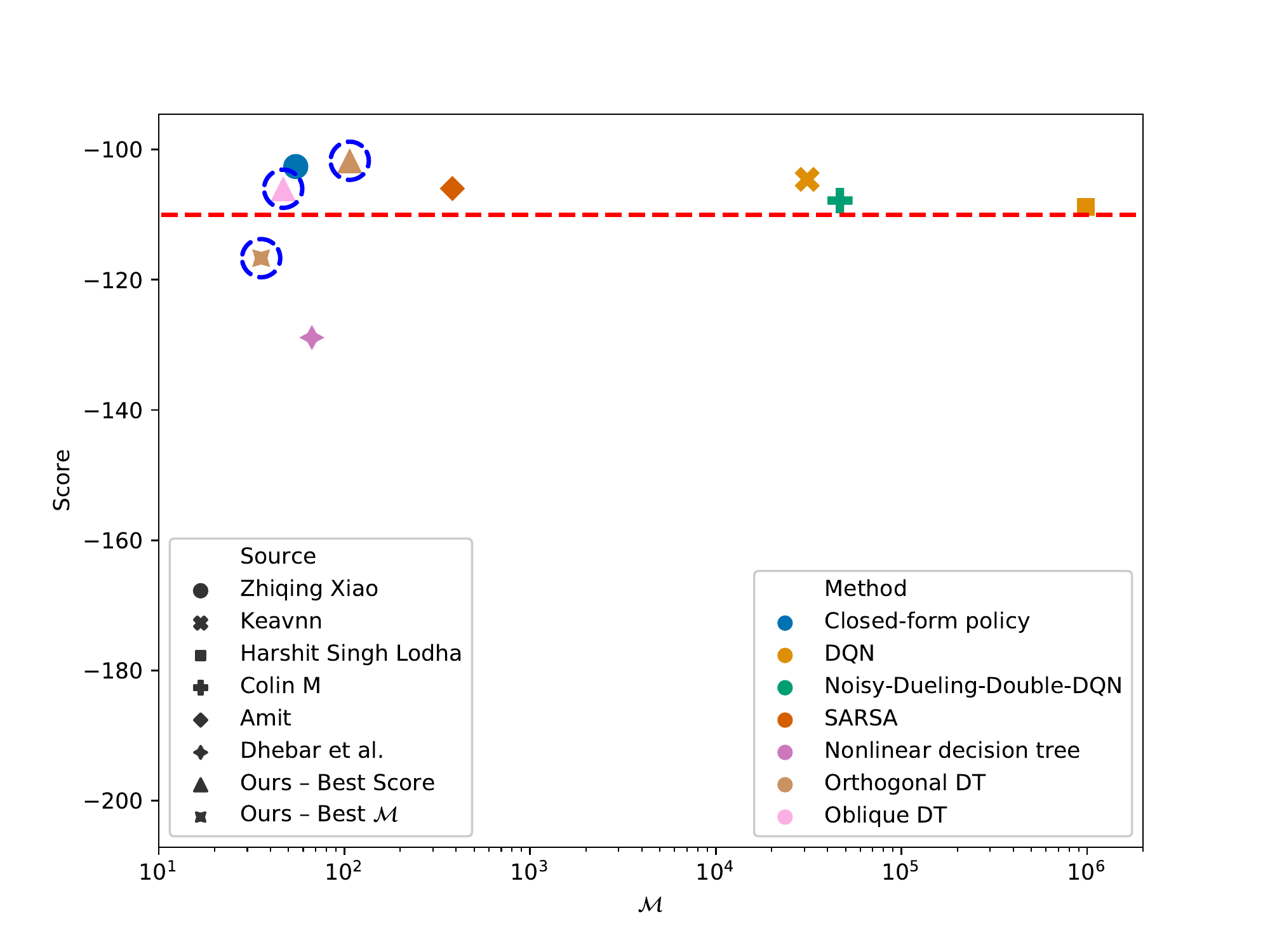}
    \caption{Score-$\mathcal{M}$ comparison of our solutions and the state-of-the-art in the MountainCar-v0 environment.
    }
    \label{fig:mc_sota_comparison}
\end{figure}

Finally, we perform a comparison of our solutions w.r.t. the state-of-the-art.
In Table \ref{tab:mc_sota_comparison} and Figure \ref{fig:mc_sota_comparison} we show the results of our comparison.
The best trees (on testing mean score) that have been used for the comparison are shown in Figures \ref{fig:mc_ort_best} and \ref{fig:mc_obl_best}.

\newpage

\begin{table}[!ht]
    \resizebox{\textwidth}{!}{
    \begin{tabular}{|l|l|l|l|} \hline
        \textbf{Source} & \textbf{Method} & \textbf{Score} & \textbf{$\mathcal{M}$} \\ \hline
        Zhiqing Xiao\tablefootnote{github.com/ZhiqingXiao/OpenAIGymSolution, accessed: 11 dec 2020.} & Closed-form policy & \textbf{-102.61} & \textbf{54.7} \\
        Keavnn\tablefootnote{github.com/StepNeverStop/RLs, accessed: 11 dec 2020.} & Soft Q Networks \cite{liu2019soft}& -104.58 & 31079.2\\
        Harshit Singh\tablefootnote{github.com/harshitandro/Deep-Q-Network, accessed: 11 dec 2020.} & Deep Q Network & -108.85 & 984160.3\\
        Colin M\tablefootnote{github.com/CM-Data/Noisy-Dueling-Double-DQN-MountainCar, accessed: 11 dec 2020.} & Double Deep Q Network & -107.83 & 46681.6\\
        Amit\tablefootnote{github.com/amitkvikram/rl-agent, accessed: 11 dec 2020.} & Tabular SARSA & -105.99 & 381.5 \\
        Dhebar et al. \cite{dhebar_interpretable-ai_2020} & Nonlinear DT (Open loop) & -128.87 & 66.8 \\ \hline
        Ours – Best Score         & Orthogonal DT & \textbf{-101.72} & 106.80 \\
        Ours – Best Score         & Oblique DT    & -106.02 & \textbf{46.80}  \\
        Ours – Best $\mathcal{M}$ & Orthogonal DT & -116.68 & \textbf{35.60}  \\
        Ours – Best $\mathcal{M}$ & Oblique DT    & -200.00  & \textbf{0.00} \\ \hline
    \end{tabular}
}
    \caption{Comparison of the mean (testing) score of the solutions obtained by using the proposed approach versus the state-of-the-art.}
    \label{tab:mc_sota_comparison}
\end{table}

\begin{center}
    \begin{figure}
        \centering
            \begin{tikzpicture}[scale=1, transform shape]
            \node [box] (olaqxlzo) {$v < 0.0$};
            \node [box, below=0.5cm of olaqxlzo, xshift=-2.5cm] (vkguinfb) {$x > -0.9$};
            \node [leaf, below=0.5cm of vkguinfb, xshift=-1.5cm] (gpqkjnvf) {acc\_left};
            \node [leaf, below=0.5cm of vkguinfb, xshift=+1.5cm] (diwgdmlc) {acc\_right};
            \node [box, below=0.5cm of olaqxlzo, xshift=+2.5cm] (umbwrvip) {$x > -0.3$};
            \node [leaf, below=0.5cm of umbwrvip, xshift=-1.5cm] (ysdjdaqa) {acc\_right};
            \node [box, below=0.5cm of umbwrvip, xshift=+1.5cm] (lqmythsd) {$v < 0.035$};
            \node [box, below=0.5cm of lqmythsd, xshift=-1.5cm] (nqufyuzd) {$x > -0.45$};
            \node [box, below=0.5cm of nqufyuzd, xshift=-1.5cm] (gwndcwmn) {$x < -0.4$};
            \node [leaf, below=0.5cm of gwndcwmn, xshift=-1.5cm] (tcutdokg) {acc\_right};
            \node [leaf, below=0.5cm of gwndcwmn, xshift=+1.5cm] (vbsccybu) {acc\_left};
            \node [leaf, below=0.5cm of nqufyuzd, xshift=+1.5cm] (bzjxbywp) {acc\_right};
            \node [leaf, below=0.5cm of lqmythsd, xshift=+1.5cm] (vusfbjam) {acc\_right};
            
            \draw (olaqxlzo) -| (vkguinfb) node [midway, above] (TextNode) {True};
            \draw (vkguinfb) -| (gpqkjnvf) node [midway, above] (TextNode) {True};
            \draw (vkguinfb) -| (diwgdmlc) node [midway, above] (TextNode) {False};
            \draw (olaqxlzo) -| (umbwrvip) node [midway, above] (TextNode) {False};
            \draw (umbwrvip) -| (ysdjdaqa) node [midway, above] (TextNode) {True};
            \draw (umbwrvip) -| (lqmythsd) node [midway, above] (TextNode) {False};
            \draw (lqmythsd) -| (nqufyuzd) node [midway, above] (TextNode) {True};
            \draw (nqufyuzd) -| (gwndcwmn) node [midway, above] (TextNode) {True};
            \draw (gwndcwmn) -| (tcutdokg) node [midway, above] (TextNode) {True};
            \draw (gwndcwmn) -| (vbsccybu) node [midway, above] (TextNode) {False};
            \draw (nqufyuzd) -| (bzjxbywp) node [midway, above] (TextNode) {False};
            \draw (lqmythsd) -| (vusfbjam) node [midway, above] (TextNode) {False};
            \end{tikzpicture}
        \caption{Best orthogonal decision tree (w.r.t. score) evolved in the MountainCar-v0 environment.}
        \label{fig:mc_ort_best}
    \end{figure}
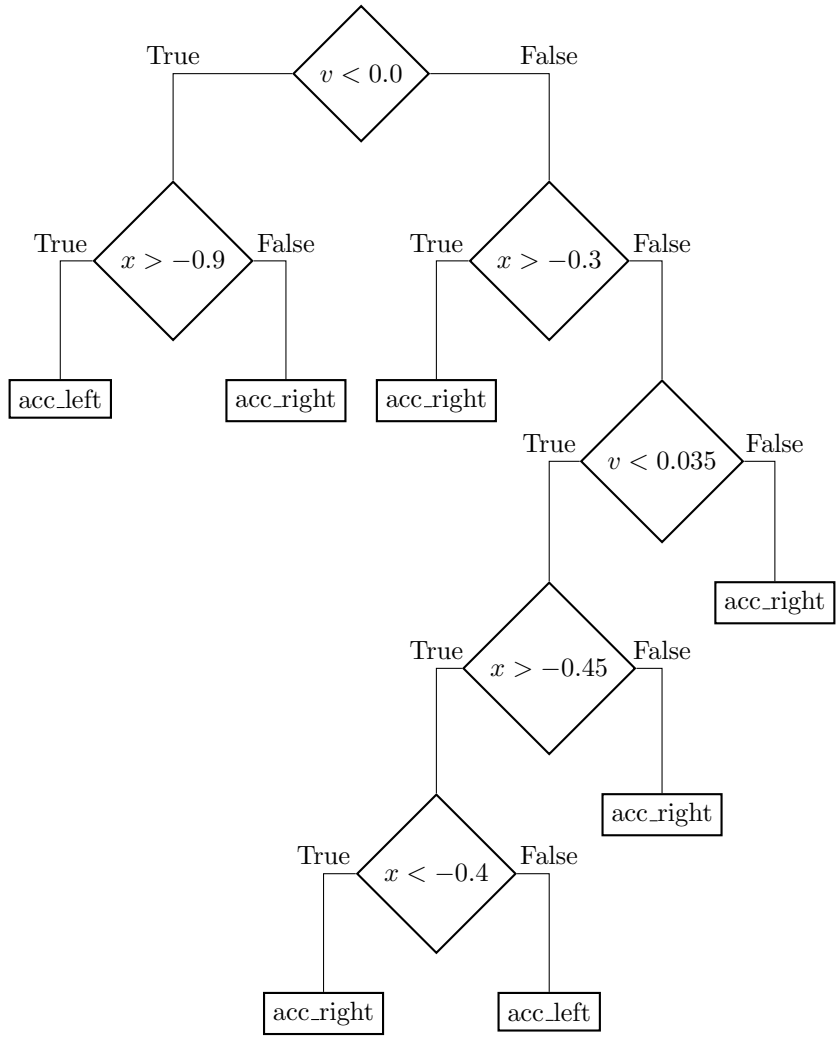
\end{center}

\begin{center}
    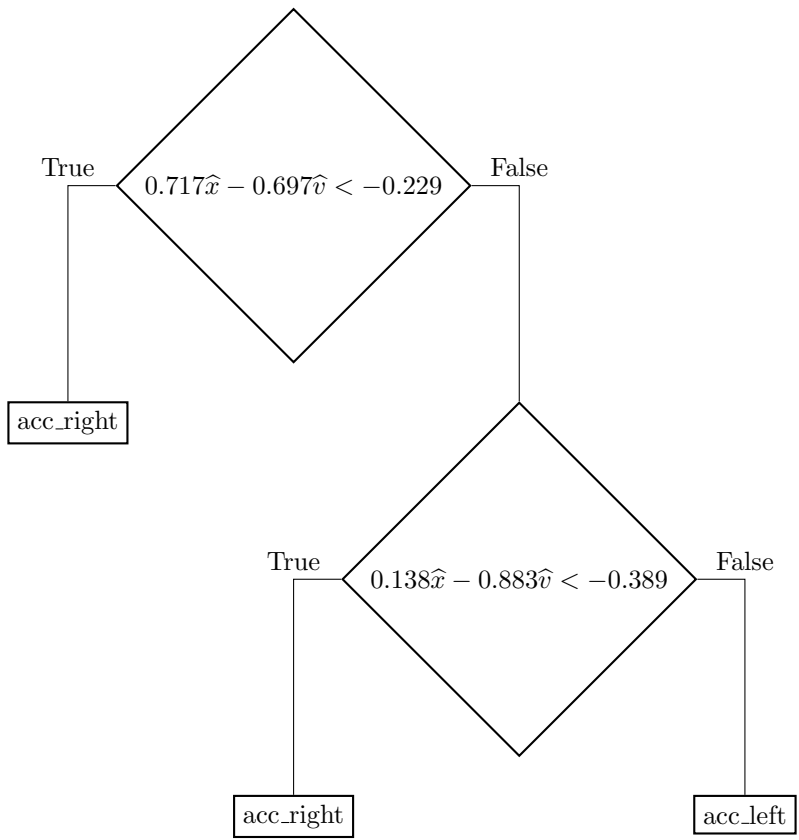
\begin{figure}
        \centering
            \begin{tikzpicture}[scale=1, transform shape]
            \node [box] (fkssodwn) {$0.717\widehat{x}-0.697\widehat{v} < -0.229$};
            \node [leaf, below=0.5cm of fkssodwn, xshift=-3cm] (qdntrztv) {acc\_right};
            \draw (fkssodwn) -| (qdntrztv) node [midway, above] (TextNode) {True};
            \node [box, below=0.5cm of fkssodwn, xshift=+3cm] (kzgujzvb) {$0.138\widehat{x}-0.883\widehat{v} < -0.389$};
            \draw (fkssodwn) -| (kzgujzvb) node [midway, above] (TextNode) {False};
            \node [leaf, below=0.5cm of kzgujzvb, xshift=-3cm] (owlsoovr) {acc\_right};
            \draw (kzgujzvb) -| (owlsoovr) node [midway, above] (TextNode) {True};
            \node [leaf, below=0.5cm of kzgujzvb, xshift=+3cm] (wingxycf) {acc\_left};
            \draw (kzgujzvb) -| (wingxycf) node [midway, above] (TextNode) {False};
            
            \end{tikzpicture}
        \caption{Best oblique decision tree (w.r.t. score) evolved in the MountainCar-v0 environment.}
        \label{fig:mc_obl_best}
    \end{figure}
\end{center}

\newpage

\subsection{LunarLander}
\subsubsection{Experimental setup}
In this case, we were not able to find a configuration that gave satisfying results with orthogonal trees.
For this reason, in this case we will show only the results obtained by using an oblique grammar.

The grammar used for this task is shown in Table \ref{tab:ll_obl_grammar}, while the parameters used for the grammatical evolution and the Q-learning are shown in Tables \ref{tab:ll_obl_params_ge} and \ref{tab:ll_params_q}, respectively.

\begin{table}[p]
    \centering
    \begin{tabular}{|c|c|} \hline
        \textbf{Rule} & \textbf{Production} \\ \hline
        dt & $<if>$ \\ 
        if & $if\ <condition>\ then\ <action>\ else\ <action>$ \\ 
        condition & $lt((\sum\limits_{i=1}^{n\_variables} <const> input_i), 0)$ \\ 
        action & $leaf\ |\ <if>$ \\ 
        $const$ & $[-1, 1]$ with step $10^{-3}$ \\ \hline
    \end{tabular}
    \caption{Grammar used to evolve oblique decision trees in the LunarLander-v2. The symbol ``$\mid$" denotes the possibility to choose between different symbols. ``lt" refers to the ``less than" operator.}
    \label{tab:ll_obl_grammar}
\end{table}

\begin{table}
    \centering
    \begin{tabular}{|c|c|} \hline
        \textbf{Parameter} & \textbf{Value} \\ \hline
        Population size & 100 \\ 
        Generations & 100 \\ 
        Genotype length & 100 \\ 
        Crossover probability & 0.1 \\ 
        Crossover operator & One-point crossover \\ 
        Selection operator & Tournament selection with size 2 \\ 
        Mutation probability & 1 \\ 
        Mutation type & Uniform, with gene probability=0.05 \\ \hline
    \end{tabular}
    \caption{Parameters used for the Grammatical Evolution with oblique grammar in the LunarLander-v2 environment.}
    \label{tab:ll_obl_params_ge}
\end{table}

\begin{table}
    \centering
    \begin{tabular}{|c|c|} \hline
        \textbf{Parameter} & \textbf{Value} \\ \hline
        Algorithm & $\varepsilon$-greedy Q-learning with $\varepsilon$-decay \\ 
        $\varepsilon_0$ & 1 \\ 
        Decay multiplier & 0.99 \\ 
        Initialization strategy& Constant, with value 0 \\ 
        Learning rate & $1/k$, k is the number of visits of the action \\ 
        Number of episodes & 1000 with early stopping \\ 
        Early stopping period & 30 episodes \\ \hline
    \end{tabular}
    \caption{Parameters used for the Q-learning algorithm in the LunarLander-v2 environment.}
    \label{tab:ll_params_q}
\end{table}

In this case, as shown in Table \ref{tab:ll_params_q}, we significantly increased the number of episodes used for the training. 
This is due to the following reasons:
\begin{itemize}
    \item The LunarLander-v2 environment is not as easy to solve as the previous environments.
    \item In this case, we did not use a random initialization of the leaves, to leverage only $\mathcal{Q}$-learning to learn the state-action function.
\end{itemize}
Moreover, as shown in Table \ref{tab:ll_params_q}, we used a slightly different $\mathcal{Q}$-learning approach for this environment.
In fact, in this case, we are using a decay for the $\varepsilon$ parameter, in order to explore better the search space. 
The decay works as follows: in the $k$-th visit to the leaf, an $\varepsilon = \varepsilon_0 \cdot decay^k$ is used.
The learning rate has been set to $\frac{1}{k}$, where k is the number of visits to the state-action pair.
This guarantees that the state-action function converges to the optimum with $k\rightarrow\infty$.
Finally, to save computation time, we implemented an early stopping criterion such that if the mean score over the current period is smaller than the one obtained in the previous period, then the training is stopped.
This is based on the following assumption: if the current mean score is worse than the previous one, then we can assume that the state-action function is converging to its optimum, so the small oscillations due to the randomness made it worse than the previous mean.

\subsubsection{Results}
\begin{figure}[ht!]
    \centering
    \includegraphics[scale=0.7]{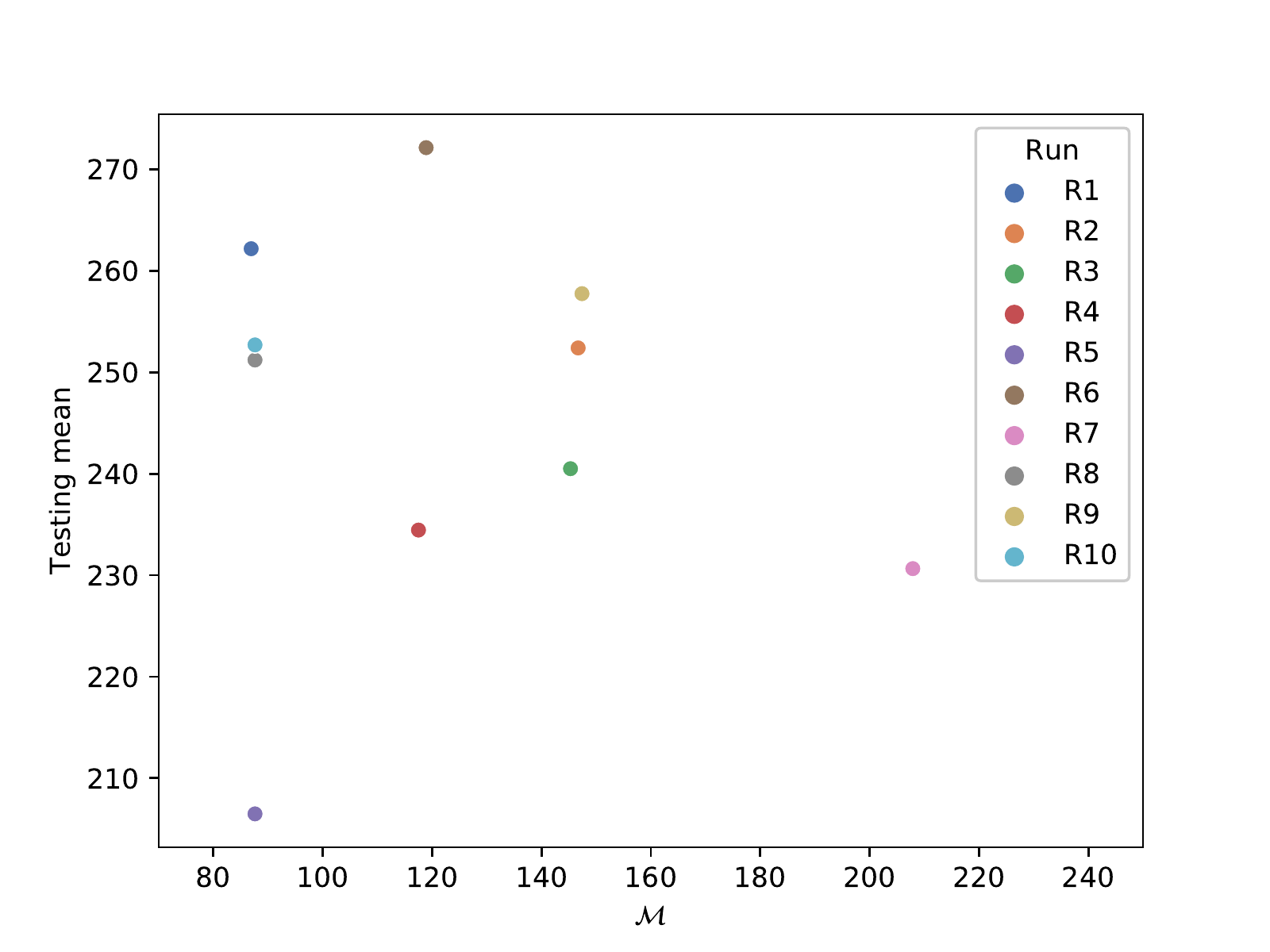}
    \caption{Score-$\mathcal{M}$ plot of the solutions obtained for the LunarLander-v2 environment.}
    \label{fig:ll_solutions}
\end{figure}

The results obtained in this environment are summarized in Table \ref{tab:ll_obl_results} and plotted in Figure \ref{fig:ll_solutions}.
We can easily observe that there is a local Pareto front between interpretability and performance composed of the solutions obtained in Runs 1 and 6.

\begin{table}[ht!]
\centering
    \begin{tabular}{|c|c|c|c|c|} \hline
    \textbf{Run} & \textbf{Training score} & \textbf{Testing mean} & \textbf{Testing std} & \textbf{$\mathcal{M}$} \\ \hline
         R1 & 265.77 & \textbf{262.18} & 29.32 & 86.90 \\ 
         R2 & 256.07 & \textbf{252.40} & 21.80 & 146.70 \\ 
         R3 & 251.95 & \textbf{240.50} & 37.21 & 145.30 \\ 
         R4 & 248.19 & \textbf{234.45} & 79.07 & 117.50 \\ 
         R5 & 220.59 & \textbf{206.48} & 64.74 & 87.60 \\ 
         R6 & 274.77 & \textbf{272.14} & 28.31 & 118.90 \\ 
         R7 & 254.33 & \textbf{230.65} & 78.29 & 207.90 \\ 
         R8 & 256.09 & \textbf{251.21} & 32.81 & 87.60 \\ 
         R9 & 265.21 & \textbf{257.75} & 31.45 & 147.40 \\ 
         R10 & 262.86 & \textbf{252.70} & 43.11 & 87.60 \\ \hline
    \end{tabular}
    \caption{Summary of the best interpretable agents obtained for each run on the LunarLander-v2 environment.
    The task is considered solved when the mean score on 100 independent runs is greater or equal to 200.}
    \label{tab:ll_obl_results}
\end{table}

In this case, our approach is able to solve the task in the 100\% of the cases. 

\begin{figure}
    \centering
    \includegraphics[scale=0.68]{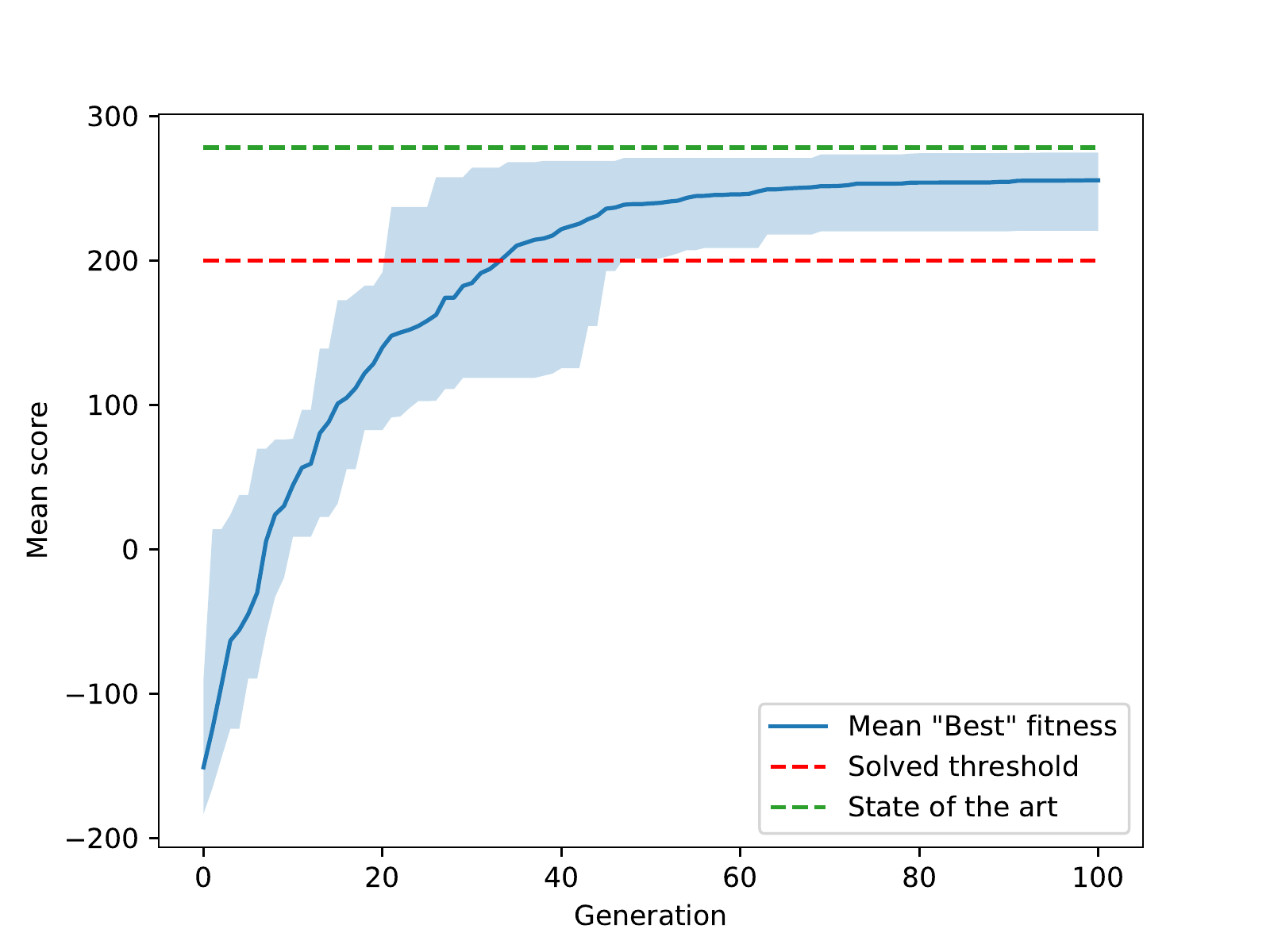}
    \caption{Mean fitness trend for the best individual in the population on the LunarLander-v2 environment.}
    \label{fig:ll_obl_fitnesstrend}
\end{figure}

Figure \ref{fig:ll_obl_fitnesstrend} shows the average fitness trend for the best solution in each run.

\begin{figure}
    \centering
    \includegraphics[scale=0.405]{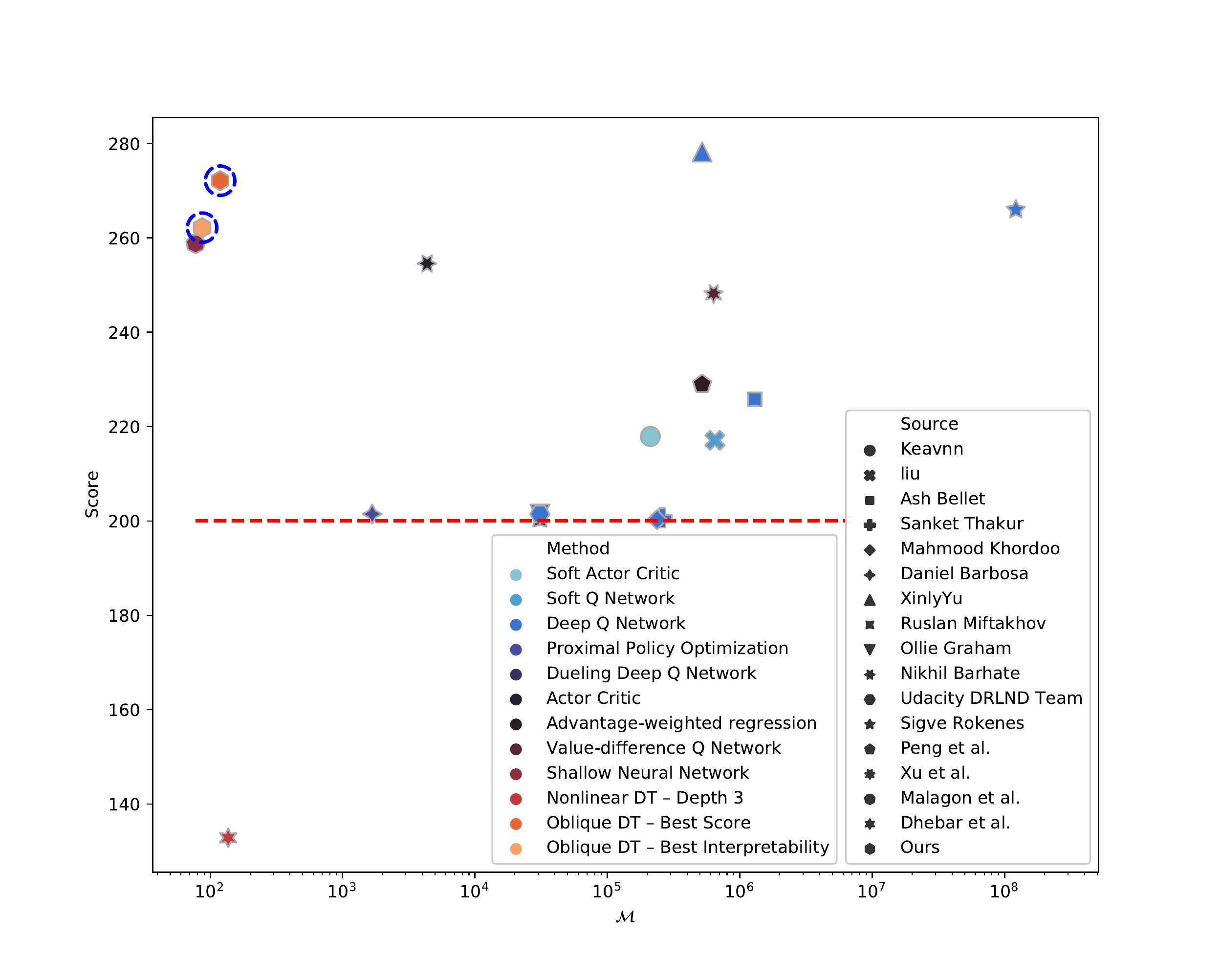}
    \caption{Plot of the score and $\mathcal{M}$ of our solutions compared to the state-of-the-art.
    Our solutions are the ones with a dashed circle around the symbol.
    The red line represents the ``solved'' threshold, fixed at 200 for the LunarLander-v2 task.
    }
    \label{fig:ll_sota_comparison}
\end{figure}

A comparison of our two best solutions (w.r.t score and interpretability) and the state-of-the-art is shown in Table \ref{tab:ll_sota_comparison} and Figure \ref{fig:ll_sota_comparison}.
As we can observe, even though we do not achieve (in absolute) the best performance and the best $\mathcal{M}$, our solutions represent the best compromise between the two metrics.
Moreover, in Figure \ref{fig:ll_sota_comparison} we can observe that a Pareto front that explains the trade-off between interpretability and performance seems to exist. 
However, our best solution achieves a comparable performance w.r.t. the best score of the state-of-the-art, while having a substantially smaller complexity.
Our best solution is shown in Figure \ref{fig:ll_best_tree}

\newpage

\begin{table}
    \centering
    \resizebox{0.9\textwidth}{!}{
    \begin{tabular}{|l|l|l|l|} \hline
    \textbf{Source} & \textbf{Method} & \textbf{Score} & \textbf{$\mathcal{M}$} \\ \hline
    Keavnn\tablefootnote{github.com/StepNeverStop/RLs, accessed: 11 dec 2020.} &   Soft Actor Critic &   217.92 &   210733.2 \\
    liu\tablefootnote{github.com/createamind/DRL, accessed: 11 dec 2020.} &   Soft Q Network &   217.09 &   647691.1 \\
    Ash Bellet\tablefootnote{github.com/nextgrid/deep-learning-labs-openAI, accessed: 11 dec 2020.} &   Deep Q Network &   225.79 &   1295307.1 \\
    Sanket Thakur\tablefootnote{github.com/sanketsans/openAIenv, accessed: 11 dec 2020.} &   Deep Q Network &   200.65 &   259285.8 \\
    Mahmood\tablefootnote{github.com/cpow-89/Extended-Deep-Q-Learning-For-Open-AI-Gym-Environments, accessed: 11 dec 2020.} &   Deep Q Network &   200.3 &   237079.7 \\
    Daniel Barbosa\tablefootnote{github.com/danielnbarbosa/angela, accessed: 11 dec 2020.} &   Proximal Policy Opt. &   201.47 &   1673 \\
    XinlyYu\tablefootnote{github.com/XinliYu/Reinforcement\_Learning-Projects, accessed: 11 dec 2020.} &   Deep Q Network &   \textbf{278.23} &   518153 \\
    Ruslan\tablefootnote{github.com/RMiftakhov/LunarLander-v2-drlnd, accessed: 11 dec 2020.} &   Dueling Deep Q N. &   200.22 &   30878.1 \\
    Ollie Graham\tablefootnote{github.com/Cozmo25/openai-lunar-lander-v2, accessed: 11 dec 2020.} &   Deep Q Network &   201.46 &   30878.1 \\
    Nikhil Barhate\tablefootnote{github.com/nikhilbarhate99/Actor-Critic-PyTorch, accessed: 11 dec 2020.} &   Actor Critic &   254.58 &   4337.3 \\
    Udacity\tablefootnote{github.com/udacity/deep-reinforcement-learning, accessed: 11 dec 2020.} &   Deep Q Network &   201.46 &   30878.1 \\
    Sigve Rokenes\tablefootnote{evgiz.net/article/2019/02/02/, accessed: 11 dec 2020.} &   Deep Q Network &   266 & $1.21\cdot10^8$ \\
    Peng et al.\cite{peng_advantage-weighted_2019} &   Advantage-weighting & $229 \pm 2$ &   518153 \\
    Xu et al.\cite{xu_deep_2020} &   Value-difference &   $248.2 \pm 21$ &   632620.2 \\
    Malagon et al.\cite{malagon_evolving_2019} &   Shallow NN &   258.8 & \textbf{77.6} \\
    Silva et al.\cite{silva_optimization_2020} &   Rule List &   -78.4 &   89 \\
    Dhebar et al.\cite{dhebar_interpretable-ai_2020} & NLDT* – Depth 3 &   132.83 &   136.7 \\ \hline
    Ours – Best Score & Oblique DT &   272.14 &   118.9 \\
    Ours – Best $\mathcal{M}$& Oblique DT  &   262.18 &   86.9 \\ 
    Ours – Mean & Oblique DT & $246.05 \pm 18.72$ & $123.34 \pm 39$ \\ \hline
    \end{tabular}
    }
\caption{Comparison of the proposed solution with respect to the state-of-the-art on the LunarLander-v2 environment.
The results from \cite{peng_advantage-weighted_2019} are averaged on 5 runs.
The results from \cite{xu_deep_2020} and \cite{malagon_evolving_2019} are averaged on 10 runs.
}
\label{tab:ll_sota_comparison}
\end{table}

\begin{center}
    \begin{figure}
        \centering
            \begin{tikzpicture}[scale=0.75, transform shape]
            \node [box, inner sep=-0.1cm] (ycmqwmqn) {\begin{tabular}{c}$0.401 p_x-0.104 p_y+$\\$+0.495 v_x-0.055 v_y+$\\$-0.69 \theta-0.845 \omega+$\\$-0.2 c_l-0.597 c_r < 0$\end{tabular}};
            \node [box, inner sep=-0.1cm, below=0.5cm of ycmqwmqn, xshift=-3cm] (bligrmck) {\begin{tabular}{c}$0.448 p_x-0.366 p_y+$\\$+0.431 v_x-0.462 v_y+$\\$-0.693 \theta-0.821 \omega+$\\$+0.461 c_l-0.132 c_r < 0$\end{tabular}};
            \draw (ycmqwmqn) -| (bligrmck) node [midway, above] (TextNode) {True};
            \node [leaf, below=0.5cm of bligrmck, xshift=-3cm] (oqwaptbv) {left};
            \draw (bligrmck) -| (oqwaptbv) node [midway, above] (TextNode) {True};
            \node [box, inner sep=-0.1cm, below=0.5cm of bligrmck, xshift=+3cm] (gtfgqfsc) {\begin{tabular}{c}$-0.101 p_x+0.133 p_y+$\\$-0.791 v_x+0.653 v_y+$\\$-0.207 \theta+0.731 \omega+$\\$+0.068 c_l+0.525 c_r < 0$\end{tabular}};
            \draw (bligrmck) -| (gtfgqfsc) node [midway, above] (TextNode) {False};
            \node [leaf, below=0.5cm of gtfgqfsc, xshift=-3cm] (ilmcxxzy) {main};
            \draw (gtfgqfsc) -| (ilmcxxzy) node [midway, above] (TextNode) {True};
            \node [box, inner sep=-0.1cm, below=0.5cm of gtfgqfsc, xshift=+3cm] (cxuedzaz) {\begin{tabular}{c}$0.12 p_x-0.044 p_y+$\\$-0.772 v_x-0.136 v_y+$\\$-0.169 \theta+0.821 \omega+$\\$-0.573 c_l-0.251 c_r < 0$\end{tabular}};
            \draw (gtfgqfsc) -| (cxuedzaz) node [midway, above] (TextNode) {False};
            \node [leaf, below=0.5cm of cxuedzaz, xshift=-3cm] (txmhqwia) {nop};
            \draw (cxuedzaz) -| (txmhqwia) node [midway, above] (TextNode) {True};
            \node [leaf, below=0.5cm of cxuedzaz, xshift=+3cm] (lnhlljdi) {main};
            \draw (cxuedzaz) -| (lnhlljdi) node [midway, above] (TextNode) {False};
            \node [leaf, below=0.5cm of ycmqwmqn, xshift=+3cm] (rlqqrnpt) {right};
            \draw (ycmqwmqn) -| (rlqqrnpt) node [midway, above] (TextNode) {False};

            \end{tikzpicture}
        \caption{Best oblique decision tree (w.r.t. score) evolved in the LunarLander-v2 environment.}
        \label{fig:ll_best_tree}
    \end{figure}
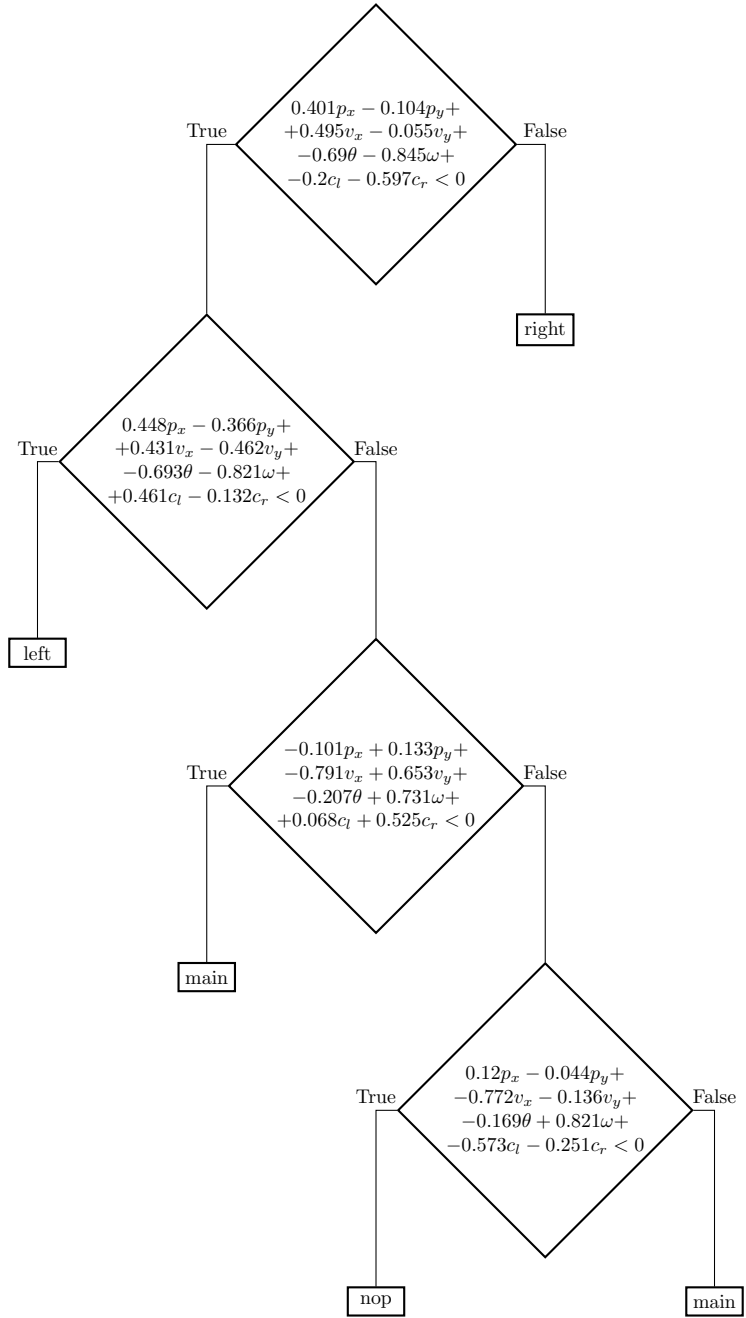
\end{center}

\newpage
\section{Discussion}
\label{sec:discussion}

In this section we briefly describe the interpretable techniques proposed in literature and we discuss our results in comparison to them.
Then, we will perform an ablation study and, finally, we will interpret the produced trees.

\subsection{CartPole}
\subsubsection{Differentiable Decision Trees}
Silva et al. \cite{silva_optimization_2020} propose an approach based on differentiable decision trees, i.e. decision trees that replace hard splits with sigmoids.
This means that they refactor the conditions from $variable > constant$ to $\sigma(variable - constant)$.
By replacing hard splits with sigmoids, the decision of the tree can be seen as the sum of all the leaves weighted by the product of the outputs of the sigmoids for that path (i.e. the product of all the $\sigma(variable - threshold)$ for the true branch and $(1 - \sigma(variable - threshold))$ for the false branch for each split encountered).
They optimize the splits of the tree and the actions taken by using PPO \cite{schulman_proximal_2017} and backpropagation.
The solution proposed for the CartPole-v1 environment is the decision tree shown in Figure \ref{fig:cp_tree_silva}.

\begin{center}
    \begin{figure}[!ht]
        \centering
        \begin{tikzpicture}[scale=0.75, transform shape]
            \node [box] (idlgqrag) {$\omega > 0.44$};
            \node [box, below=0.5cm of idlgqrag, xshift=-3cm] (cdfyyrsr) {$\omega > -0.3$};
            \draw (idlgqrag) -| (cdfyyrsr) node [midway, above] (TextNode) {True};
            \node [box, below=0.5cm of cdfyyrsr, xshift=-2.5cm] (iedmkmgf) {$\theta > -0.41$};
            \draw (cdfyyrsr) -| (iedmkmgf) node [midway, above] (TextNode) {True};
            \node [leaf, below=0.5cm of iedmkmgf, xshift=-1.5cm] (idlztbdt) {move\_right};
            \draw (iedmkmgf) -| (idlztbdt) node [midway, above] (TextNode) {True};
            \node [leaf, below=0.5cm of iedmkmgf, xshift=+1.5cm] (hkhaqmgb) {move\_left};
            \draw (iedmkmgf) -| (hkhaqmgb) node [midway, above] (TextNode) {False};
            \node [box, below=0.9cm of cdfyyrsr, xshift=+2.5cm] (twskyclb) {$\theta > 0$};
            \draw (cdfyyrsr) -| (twskyclb) node [midway, above] (TextNode) {False};
            \node [leaf, below=0.85cm of twskyclb, xshift=-1.5cm] (igvhucae) {move\_left};
            \draw (twskyclb) -| (igvhucae) node [midway, above] (TextNode) {True};
            \node [leaf, below=0.85cm of twskyclb, xshift=+1.5cm] (iqhmnfzs) {move\_right};
            \draw (twskyclb) -| (iqhmnfzs) node [midway, above] (TextNode) {False};
            \node [box, below=0.55cm of idlgqrag, xshift=+3cm] (xckiihtx) {$\theta > 0.01$};
            \draw (idlgqrag) -| (xckiihtx) node [midway, above] (TextNode) {False};
            \node [leaf, below=0.5cm of xckiihtx, xshift=-1.5cm] (pfyvajgh) {move\_right};
            \draw (xckiihtx) -| (pfyvajgh) node [midway, above] (TextNode) {True};
            \node [leaf, below=0.5cm of xckiihtx, xshift=+1.5cm] (qakddjuu) {move\_left};
            \draw (xckiihtx) -| (qakddjuu) node [midway, above] (TextNode) {False};
        \end{tikzpicture}
    \caption{Tree representation of the solution proposed in \cite{silva_optimization_2020}.}
    \label{fig:cp_tree_silva}
    \end{figure}
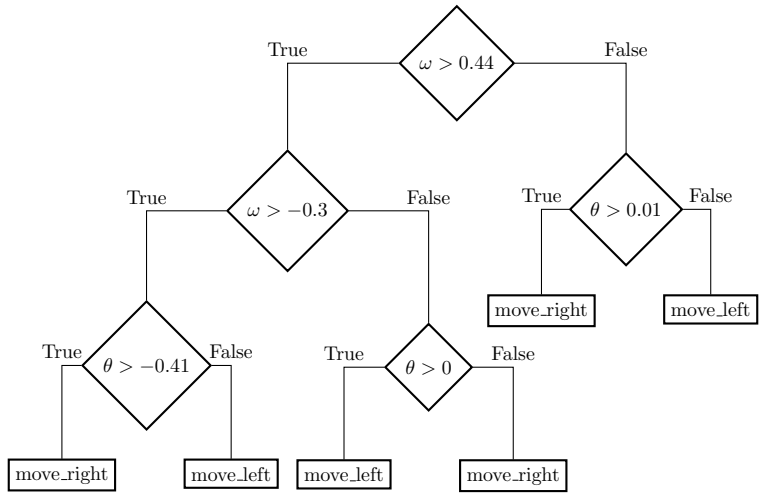
\end{center}

It is interesting to observe that their optimization process ``selects'' the same variables that have been selected in our case by artificial evolution.

Moreover, the tree proposed by them is slightly more complex than our best tree.
In fact, while our best tree has a maximum depth of 2 conditions, their has a maximum depth of 3 conditions. 
This increase in complexity is reflected by the difference in the $\mathcal{M}$ measure.

Moreover, since the performance of this solution are not satisfactory, the authors gave us the solution coming from a follow-up work in a personal communication.
This solution is shown in Figure \ref{fig:cp_tree_silva_followup}.

\begin{center}
    \begin{figure}[!ht]
        \centering
        \begin{tikzpicture}[scale=0.75, transform shape]
            \node [box] (seubmjlf) {$\omega > 0.18$};
            \node [box, below=0.5cm of seubmjlf, xshift=-4cm] (mvsgmznd) {$\omega > -0.30$};
            \draw (seubmjlf) -| (mvsgmznd) node [midway, above] (TextNode) {True};
            \node [box, below=0.cm of mvsgmznd, xshift=-2.25cm] (igsfueld) {$\theta > -0.41$};
            \draw (mvsgmznd) -| (igsfueld) node [midway, above] (TextNode) {True};
            \node [leaf, below=0.5cm of igsfueld, xshift=-1.5cm] (owlyoznb) {move\_right};
            \draw (igsfueld) -| (owlyoznb) node [midway, above] (TextNode) {True};
            \node [leaf, below=0.5cm of igsfueld, xshift=+1.5cm] (nnsjonoz) {move\_left};
            \draw (igsfueld) -| (nnsjonoz) node [midway, above] (TextNode) {False};
            \node [box, below=0.44cm of mvsgmznd, xshift=+2.25cm] (yielwpwi) {$\theta > 0$};
            \draw (mvsgmznd) -| (yielwpwi) node [midway, above] (TextNode) {False};
            \node [leaf, below=0.8cm of yielwpwi, xshift=-1.1cm] (puwnyted) {move\_left};
            \draw (yielwpwi) -| (puwnyted) node [midway, above] (TextNode) {True};
            \node [leaf, below=0.8cm of yielwpwi, xshift=+1.1cm] (vfdamteb) {move\_right};
            \draw (yielwpwi) -| (vfdamteb) node [midway, above] (TextNode) {False};
            \node [box, below=0.5cm of seubmjlf, xshift=+4cm] (wkngnnbb) {$\theta > 0$};
            \draw (seubmjlf) -| (wkngnnbb) node [midway, above] (TextNode) {False};
            \node [box, below=0.74cm of wkngnnbb, xshift=-1.2cm] (nuwsmqvp) {$\omega > -0.30$};
            \draw (wkngnnbb) -| (nuwsmqvp) node [midway, above] (TextNode) {True};
            \node [leaf, below=0.5cm of nuwsmqvp, xshift=-1.5cm] (amexlakc) {move\_right};
            \draw (nuwsmqvp) -| (amexlakc) node [midway, above] (TextNode) {True};
            \node [leaf, below=0.5cm of nuwsmqvp, xshift=+1.5cm] (iqxgxaow) {move\_left};
            \draw (nuwsmqvp) -| (iqxgxaow) node [midway, above] (TextNode) {False};
            \node [leaf, below=0.5cm of wkngnnbb, xshift=+1.2cm] (eiduykbf) {move\_left};
            \draw (wkngnnbb) -| (eiduykbf) node [midway, above] (TextNode) {False};
            
        \end{tikzpicture} 
        \caption{Tree representation of the solution obtained by private communication with the first author of \cite{silva_optimization_2020}}
        \label{fig:cp_tree_silva_followup}
    \end{figure}
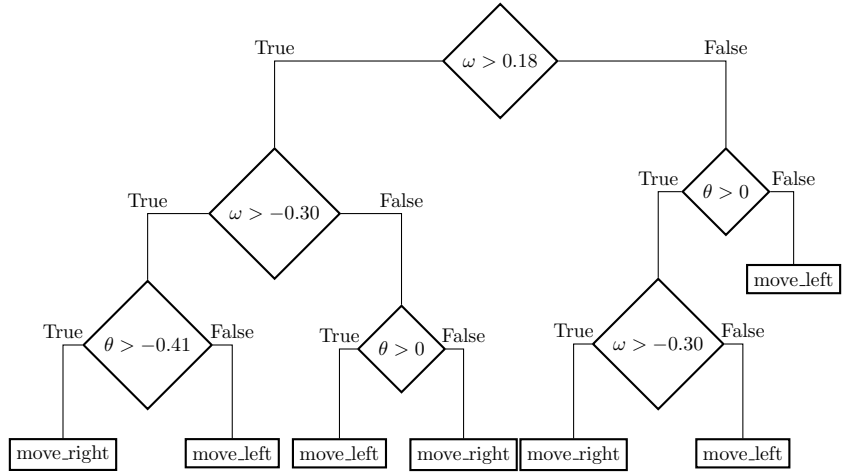
\end{center}

Besides the performance comparison performed in Table \ref{tab:cp_sota_comparison}, we compare here the robustness to noise, similarly to what we did in Figure \ref{fig:cp_noise_comparison}.

\begin{figure}[!ht]
    \centering
    \includegraphics[scale=0.7]{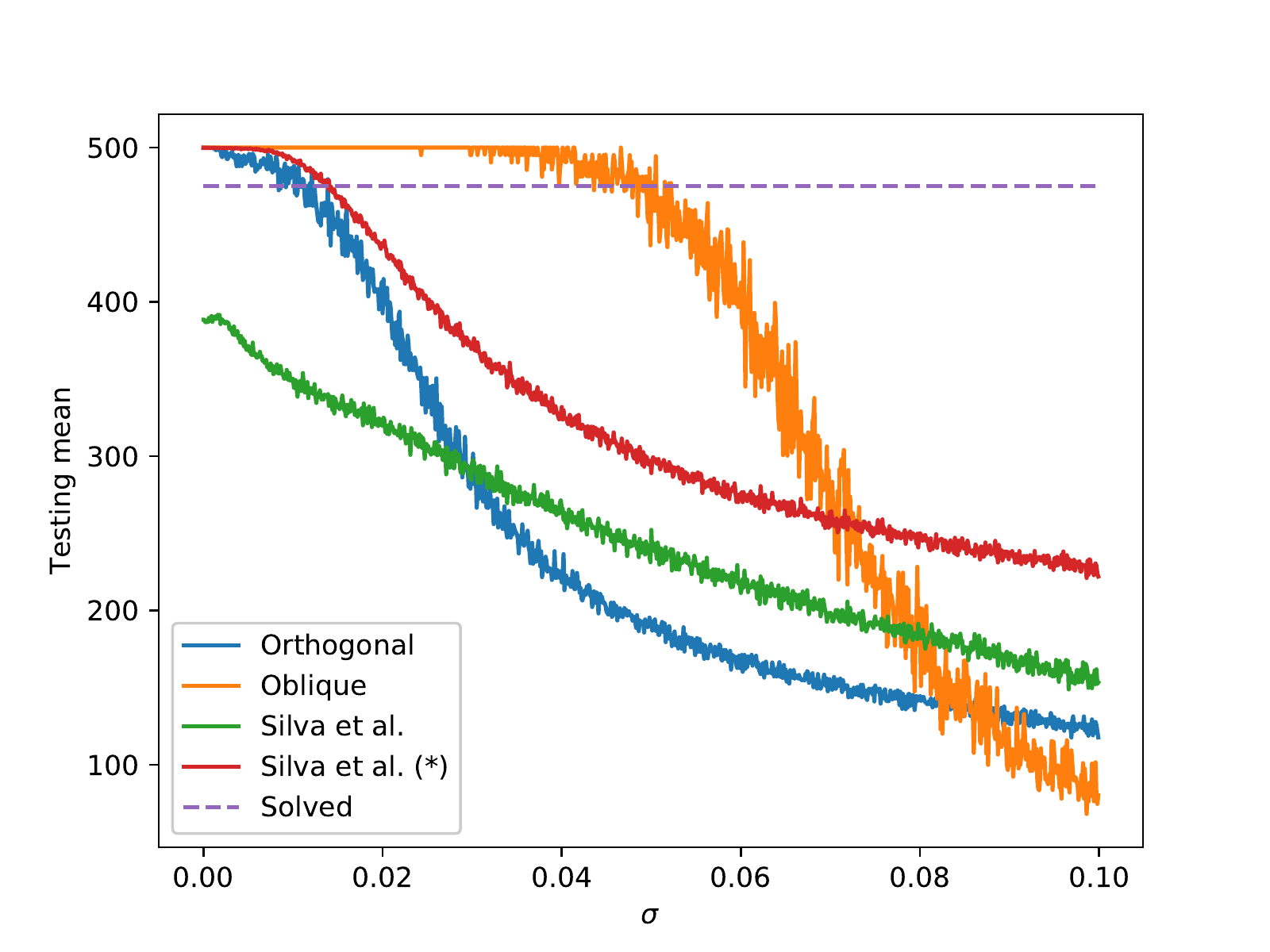}
    \caption{Robustness to input noise on the CartPole-v1 environment.
    (*): Private communication with the first author.}
    \label{fig:cp_noise_sota_comparison}
\end{figure}

As we can see in Figure \ref{fig:cp_noise_sota_comparison}, the orthogonal trees obtained by Silva et al. have a robustness that is comparable to our orthogonal tree.
This suggests us that orthogonal trees may be intrinsically less robust than oblique ones.

\subsection{MountainCar}
Most of the approaches we used for the comparison in the MountainCar-v0 environments come from the OpenAI Gym leaderboard.

\subsubsection{Zhiqing Xiao}
The system proposed by this entry\footnote{github.com/ZhiqingXiao/OpenAIGymSolution/tree/master/MountainCar-v0\_close\_form} consists in a closed-form policy. 
However, it is not clear whether the policy has been derived by a human or learned by a machine.

Anyway, this solution achieves the best performance (let alone our solutions) on this task while also having the best degree of interpretability (according to our modification of the metric proposed in \cite{virgolin_learning_2020}).
The policy is the following:

$$
a = min(-0.09 (x + 0.25)^2 + 0.03, 0.3 (x + 0.9)^4 - 0.008)
$$
$$
b = -0.07 (x + 0.38)^2 + 0.07
$$
\begin{equation*}
\pi(x, v) = \begin{cases}
acc\_right & \text{if $a < v < b$}\\
acc\_left & \text{else}
\end{cases}
\end{equation*}

While $\mathcal{M}$ is lower for this policy than for our best tree, it may be a bit harder to interpret this model.
We think that this is due to the fact that the $\mathcal{M}$ metric has been proposed to evaluate the interpretability of mathematical formulae, while we are interested in interpreting \textit{hyperplanes}.
While hyperplanes are defined by mathematical formulae, the interpretability of an hyperplane may also depend on the number on non-linear operations that are used to determine the hyperplane.

\subsubsection{Amit}
This entry\footnote{github.com/amitkvikram/rl-agent/blob/master/mountainCar-v0-sarsa.ipynb} uses SARSA to solve the task.

While tabular approaches like SARSA and Q-learning are transparent, their interpretability depends heavily on the number of states and actions.
Table \ref{tab:mc_sota_comparison} shows that, even if this approach is transparent and easily interpretable, our solutions are able to achieve a better degree of interpretability. 
In our opinion, this is due to the fact that using decision trees as function approximators leads to the ``grouping'' of some states of the table used in tabular approaches.
This is especially useful when we want to \textit{extract} knowledge. 
In fact, by grouping some states, we take into account only the variables and the thresholds that have a big impact on the policy, discarding irrelevant details.

\subsubsection{Dhebar et al.}
Dhebar et al., in \cite{dhebar_interpretable-ai_2020}, propose an approach to reinforcement learning that uses nonlinear decision trees.
They first approximate an oracle policy and then they fine-tune it by using evolutionary algorithms. The policies obtained in these two phases are called ``open-loop'' and ``closed-loop'' policies.

In this case, we only had access to the open-loop policy for the MountainCar-v0 environment, which is shown in Figure \ref{fig:mc_tree_dhebar}.

\begin{center}
    \begin{figure}[!ht]
        \centering
        \begin{tikzpicture} [scale=0.8, transform shape]
            \node [box] (oxzvdqqz) {\begin{tabular}{c}$ \mid -0.22 \hat{x}\hat{y} +$\\$0.28 \hat{y}^-1 +$\\$- 0.63\hat{x}^-2 +$\\$ 0.96\mid + $\\$ - 0.36 \leq 0$\end{tabular}};
            \node [box, below=0.cm of oxzvdqqz, xshift=-3cm] (nylfboqe)
            {\begin{tabular}{c}$\mid-0.30\hat{y}^2 +$\\$- 0.28\hat{x}^2 +$\\
            $ 1.39\mid + $\\$ - 0.53 \leq 0$\end{tabular}};
            \draw (oxzvdqqz) -| (nylfboqe) node [midway, above] (TextNode) {True};
            \node [leaf, below=0.5cm of nylfboqe, xshift=-3cm] (hppztfbs) {acc\_right};
            \draw (nylfboqe) -| (hppztfbs) node [midway, above] (TextNode) {True};
            \node [leaf, below=0.5cm of nylfboqe, xshift=+3cm] (mdnawkka) {no\_acc};
            \draw (nylfboqe) -| (mdnawkka) node [midway, above] (TextNode) {False};
            \node [leaf, below=0.5cm of oxzvdqqz, xshift=+3cm] (adbwcmni) {acc\_left};
            \draw (oxzvdqqz) -| (adbwcmni) node [midway, above] (TextNode) {False};
    
        \end{tikzpicture}
    \caption{Tree representation of the solution proposed in \cite{dhebar_interpretable-ai_2020} for the MountainCar-v0 environment.
    The variables with a hat are normalized by using this way: $1 + \frac{x - x_{min}}{x_{max}-x_{min}}$.}
    \label{fig:mc_tree_dhebar}
    \end{figure}
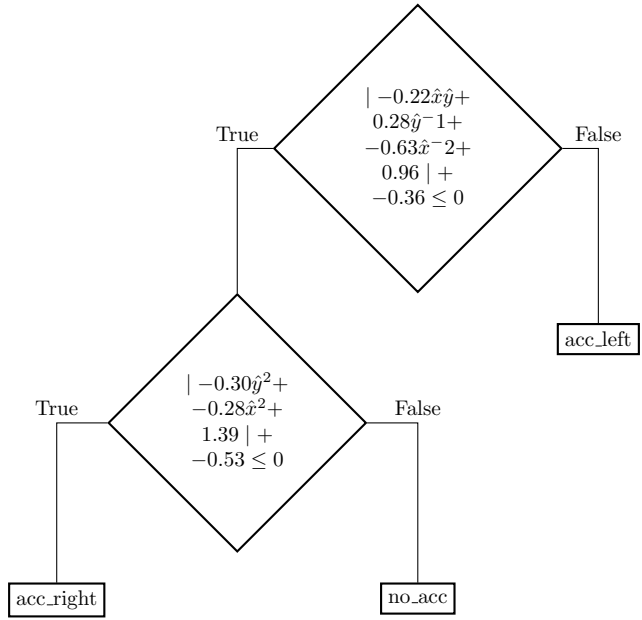
\end{center}

Also in this case, while $\mathcal{M}$ for this solution is better than our best solution (w.r.t. test score), it seems harder to interpret, due to the non-linearity of the hyperplanes.
In fact, in our solution $\mathcal{M}$ is higher due to the higher number of splits in the tree, but that does not take into account the fact that in our case the hyperplanes that divide the feature space are simpler than the ones proposed in \cite{dhebar_interpretable-ai_2020}.

\begin{figure}[!ht]
    \centering
    \includegraphics[scale=0.7]{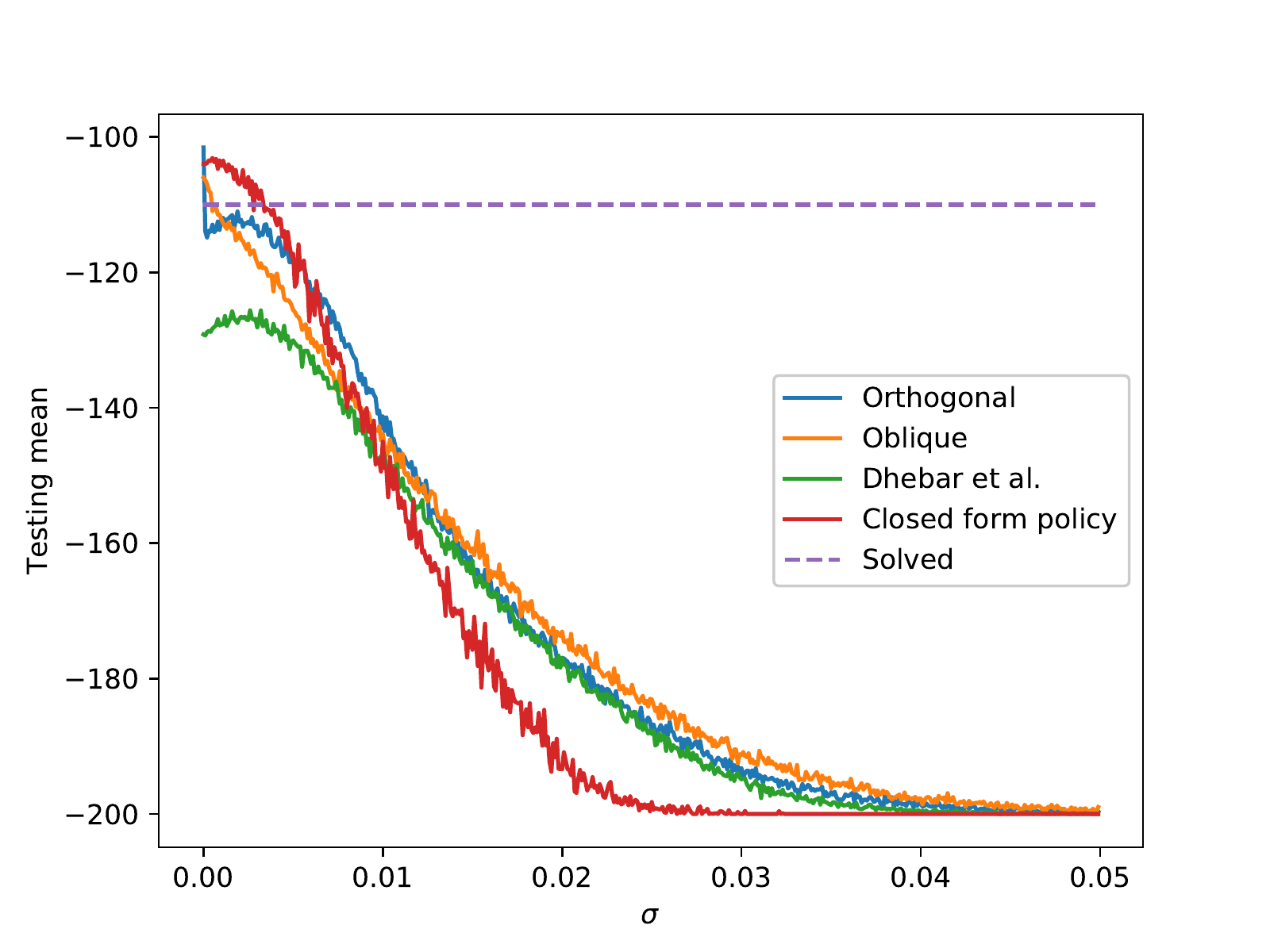}
    \caption{Comparison of the robustness to input noise between our solutions and the interpretable ones on the MountainCar-v0 environment.}
    \label{fig:mc_noise_sota_comparison}
\end{figure}

Finally, we perform a comparison on the robustness to input noise with the solutions provided by ``Zhiqing Xiao'' and the one provided by Dhebar et al.
Figure \ref{fig:mc_noise_sota_comparison} shows how performance vary by varying the standard deviation of the additive Gaussian noise.
We observe that there is no significant difference between the solutions, meaning that all of them have high sensitivity to input noise.

\subsection{LunarLander-v2}
\subsubsection{Silva et al.}
In \cite{silva_optimization_2020} the authors, besides regular trees, use also decision lists.
A decision list is a tree that is extremely unbalanced, i.e. il collapses to a list.

Figure \ref{fig:ll_tree_silva} shows the solution obtained.
However, as shown in Table \ref{tab:ll_sota_comparison}, it does not achieve satisfactory performance.
This is due to the fact that, while the differentiable tree is able to achieve better performance (even though it does not solve the task), its discretization modifies the final distribution of the actions.

\begin{center}
    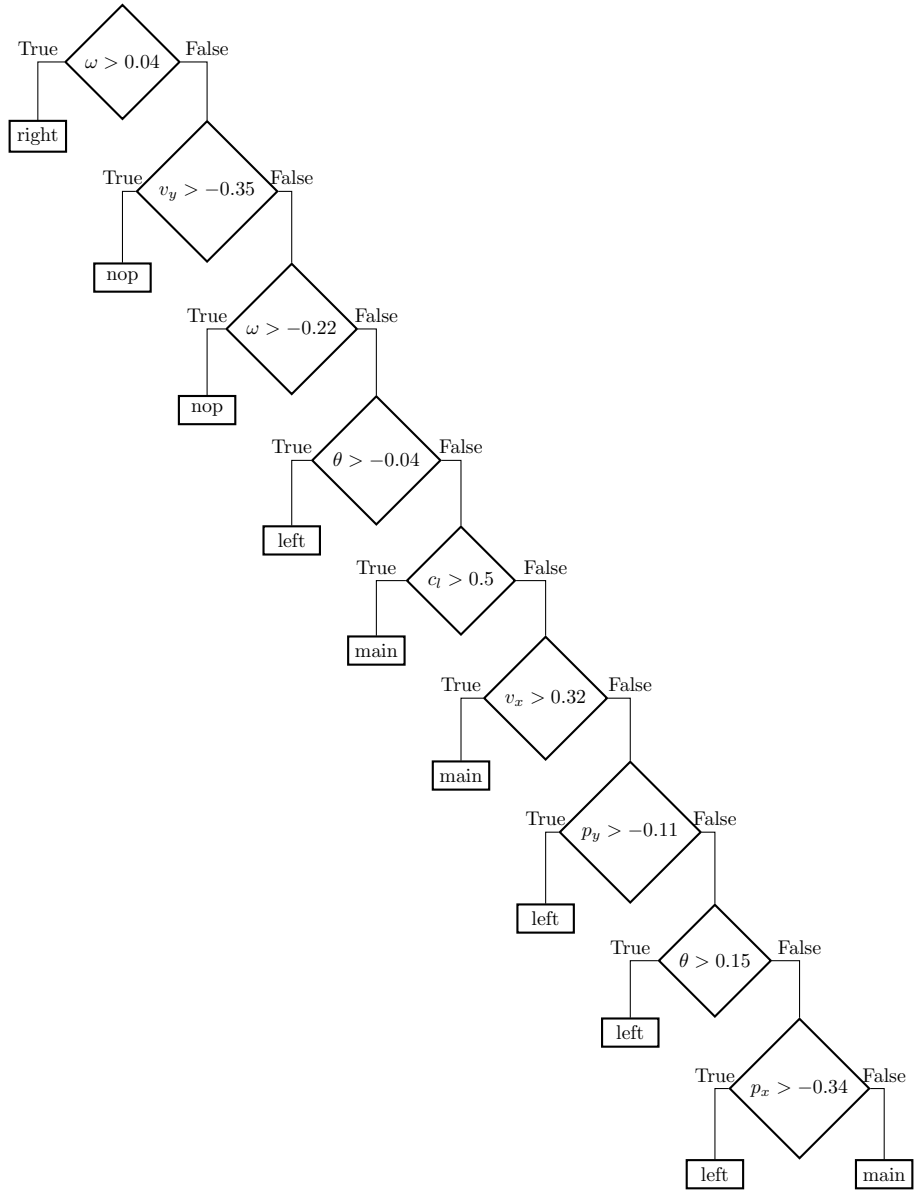
\begin{figure}[!ht]
        \centering
        \begin{tikzpicture} [scale=0.75, transform shape]
        \node [box] (exkkvuvh) {$\omega > 0.04$};
        \node [leaf, below=0cm of exkkvuvh, xshift=-1.5cm] (ntbleaqe) {right};
        \draw (exkkvuvh) -| (ntbleaqe) node [midway, above] (TextNode) {True};
        \node [box, below=0cm of exkkvuvh, xshift=+1.5cm] (dvvqtcxi) {$v_y > -0.35$};
        \draw (exkkvuvh) -| (dvvqtcxi) node [midway, above] (TextNode) {False};
        \node [leaf, below=0cm of dvvqtcxi, xshift=-1.5cm] (xpzquwwx) {nop};
        \draw (dvvqtcxi) -| (xpzquwwx) node [midway, above] (TextNode) {True};
        \node [box, below=0cm of dvvqtcxi, xshift=+1.5cm] (ytxfcrio) {$\omega > -0.22$};
        \draw (dvvqtcxi) -| (ytxfcrio) node [midway, above] (TextNode) {False};
        \node [leaf, below=0cm of ytxfcrio, xshift=-1.5cm] (llijlllv) {nop};
        \draw (ytxfcrio) -| (llijlllv) node [midway, above] (TextNode) {True};
        \node [box, below=0cm of ytxfcrio, xshift=+1.5cm] (ukvciuot) {$\theta > -0.04$};
        \draw (ytxfcrio) -| (ukvciuot) node [midway, above] (TextNode) {False};
        \node [leaf, below=0cm of ukvciuot, xshift=-1.5cm] (qjndaqcu) {left};
        \draw (ukvciuot) -| (qjndaqcu) node [midway, above] (TextNode) {True};
        \node [box, below=0cm of ukvciuot, xshift=+1.5cm] (mnninkud) {$c_l > 0.5$};
        \draw (ukvciuot) -| (mnninkud) node [midway, above] (TextNode) {False};
        \node [leaf, below=0cm of mnninkud, xshift=-1.5cm] (vrlxgjtw) {main};
        \draw (mnninkud) -| (vrlxgjtw) node [midway, above] (TextNode) {True};
        \node [box, below=0cm of mnninkud, xshift=+1.5cm] (axtsrkvj) {$v_x > 0.32$};
        \draw (mnninkud) -| (axtsrkvj) node [midway, above] (TextNode) {False};
        \node [leaf, below=0cm of axtsrkvj, xshift=-1.5cm] (iwanuaok) {main};
        \draw (axtsrkvj) -| (iwanuaok) node [midway, above] (TextNode) {True};
        \node [box, below=0cm of axtsrkvj, xshift=+1.5cm] (kwqtptjl) {$p_y > -0.11$};
        \draw (axtsrkvj) -| (kwqtptjl) node [midway, above] (TextNode) {False};
        \node [leaf, below=0cm of kwqtptjl, xshift=-1.5cm] (lgzywnst) {left};
        \draw (kwqtptjl) -| (lgzywnst) node [midway, above] (TextNode) {True};
        \node [box, below=0cm of kwqtptjl, xshift=+1.5cm] (eaqyzsyf) {$\theta > 0.15$};
        \draw (kwqtptjl) -| (eaqyzsyf) node [midway, above] (TextNode) {False};
        \node [leaf, below=0cm of eaqyzsyf, xshift=-1.5cm] (ygtxgeei) {left};
        \draw (eaqyzsyf) -| (ygtxgeei) node [midway, above] (TextNode) {True};
        \node [box, below=0cm of eaqyzsyf, xshift=+1.5cm] (nrxwzqij) {$p_x > -0.34$};
        \draw (eaqyzsyf) -| (nrxwzqij) node [midway, above] (TextNode) {False};
        \node [leaf, below=0cm of nrxwzqij, xshift=-1.5cm] (pvbdchwg) {left};
        \draw (nrxwzqij) -| (pvbdchwg) node [midway, above] (TextNode) {True};
        \node [leaf, below=0cm of nrxwzqij, xshift=+1.5cm] (unlkjjck) {main};
        \draw (nrxwzqij) -| (unlkjjck) node [midway, above] (TextNode) {False};   
        \end{tikzpicture}
    \caption{Tree representation of the solution proposed in \cite{silva_optimization_2020} for the LunarLander-v2 environment.}
    \label{fig:ll_tree_silva}
    \end{figure}
\end{center}

\subsubsection{Malagon et al.}
In \cite{malagon_evolving_2019} the authors use the Univariate Marginal Distribution Algorithm to evolve a neural network without hidden layers in the LunarLander-v2 domain.
Since the neural network has no hidden layers the whole system reduces to 
\[
a = \underset{i}{argmax}(\sigma(\mathbf{w_i}^T\cdot \mathbf{x} + b_i))
\]
where $i$ refers to the output neurons.

This results in an easy-to-interpret system, according to both \cite{lipton_mythos_2017} and the metric $\mathcal{M}$.

\subsubsection{Dhebar et al.}
In \cite{dhebar_interpretable-ai_2020} the authors propose a nonlinear decision tree that achieves a mean testing score of 234.98 points. 
However, the rules associated with this tree are not shown, so we only had access to the 3-levels-deep NLDT.

The tree is shown in Figure \ref{fig:ll_tree_dhebar}.
It is important to note that even if the solution obtained is a tree, the interpretation is not easy, since the hyperplanes contained in each split are not linear.

\begin{center}
    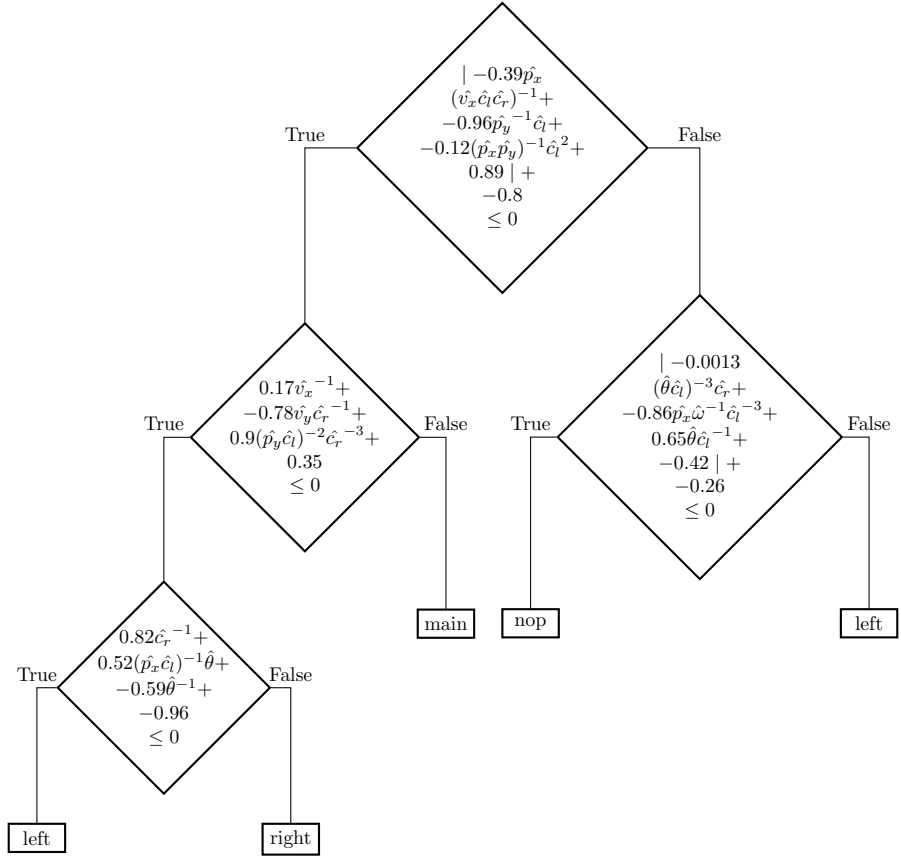
\begin{figure}[!ht]
        \centering
        \begin{tikzpicture} [scale=0.75, transform shape]
        \node [box, inner sep=-0.3cm] (lfeaftoh) {\begin{tabular}{c}$\mid-0.39\hat{p_x}$\\$(\hat{v_x}\hat{c_l}\hat{c_r}) ^{-1}+$\\ $-0.96\hat{p_y} ^{-1}\hat{c_l}+$\\ $-0.12(\hat{p_x}\hat{p_y}) ^{-1}\hat{c_l} ^ 2+$\\ $0.89 \mid+$\\ $-0.8 $\\$\leq 0$\end{tabular}};
        \node [box, inner sep=-0.3cm, below=0.5cm of lfeaftoh, xshift=-3.5cm] (oqwaextu) {\begin{tabular}{c}$0.17\hat{v_x} ^{-1}+$\\ $-0.78\hat{v_y}\hat{c_r} ^{-1}+$\\ $0.9(\hat{p_y}\hat{c_l}) ^{-2}\hat{c_r} ^{-3}+$\\ $0.35 $\\$\leq 0$\end{tabular}};
        \draw (lfeaftoh) -| (oqwaextu) node [midway, above] (TextNode) {True};
        \node [box, inner sep=-0.3cm, below=0.5cm of oqwaextu, xshift=-2.5cm] (huycufuy) {\begin{tabular}{c}$0.82\hat{c_r} ^{-1}+$\\ $0.52(\hat{p_x}\hat{c_l}) ^{-1}\hat{\theta}+$\\ $-0.59\hat{\theta} ^{-1}+$\\ $-0.96 $\\$\leq 0$\end{tabular}};
        \draw (oqwaextu) -| (huycufuy) node [midway, above] (TextNode) {True};
        \node [leaf, below=0.5cm of huycufuy, xshift=-2.25cm] (sokmbgig) {left};
        \draw (huycufuy) -| (sokmbgig) node [midway, above] (TextNode) {True};
        \node [leaf, below=0.5cm of huycufuy, xshift=+2.25cm] (vavktuyk) {right};
        \draw (huycufuy) -| (vavktuyk) node [midway, above] (TextNode) {False};
        \node [leaf, below=1.0cm of oqwaextu, xshift=+2.5cm] (qdvnoaro) {main};
        \draw (oqwaextu) -| (qdvnoaro) node [midway, above] (TextNode) {False};
        \node [box, inner sep=-0.3cm, below=0.cm of lfeaftoh, xshift=+3.5cm] (kkmlkoao) {\begin{tabular}{c}$\mid-0.0013$\\$(\hat{\theta}\hat{c_l}) ^{-3}\hat{c_r}+$\\ $-0.86\hat{p_x}\hat{\omega} ^{-1}\hat{c_l} ^{-3}+$\\ $0.65\hat{\theta}\hat{c_l} ^{-1}+$\\ $-0.42\mid+$\\ $-0.26 $\\$\leq 0$\end{tabular}};
        \draw (lfeaftoh) -| (kkmlkoao) node [midway, above] (TextNode) {False};
        \node [leaf, below=0.5cm of kkmlkoao, xshift=-3cm] (hnbvtpyv) {nop};
        \draw (kkmlkoao) -| (hnbvtpyv) node [midway, above] (TextNode) {True};
        \node [leaf, below=0.5cm of kkmlkoao, xshift=+3cm] (imenwdtw) {left};
        \draw (kkmlkoao) -| (imenwdtw) node [midway, above] (TextNode) {False};
    \end{tikzpicture}
    \caption{Tree representation of the solution proposed in \cite{dhebar_interpretable-ai_2020} for the LunarLander-v2 environment.
    The variables with a hat are normalized by using this way: $1 + \frac{x - x_{min}}{x_{max}-x_{min}}$.}
    \label{fig:ll_tree_dhebar}
    \end{figure}
\end{center}

\begin{figure}[!ht]
    \centering
    \includegraphics[scale=0.7]{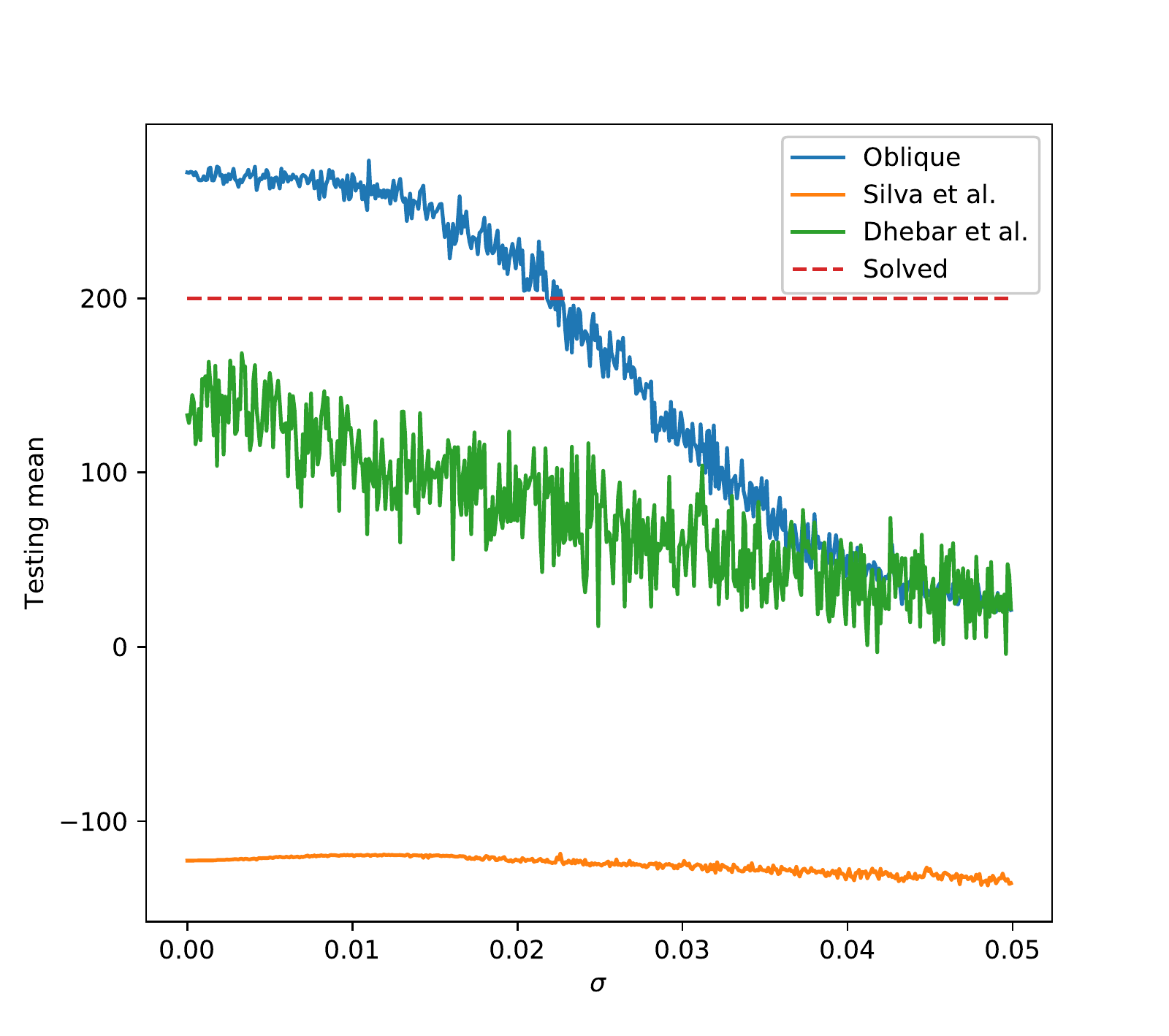}
    \caption{Comparison on the robustness to input noise w.r.t. other interpretable solutions on the LunarLander-v2 environment.}
    \label{fig:ll_noise_sota_comparison}
\end{figure}

Also in this case, we performed a comparison on the robustness to input noise, the result is shown in Figure \ref{fig:ll_noise_sota_comparison}.
However, for this comparison we could not include the results from Malagon et al. since the weights were not publicly accessible.

\subsection{Ablation study}
In order to assess whether our two-level optimization approach is convenient with respect to a single-level optimization approach, we perform an ablation study in which we use Grammatical Evolution alone to evolve decision trees.
Moreover, we perform statistical tests to test whether the difference are statistically significant by fixing a threshold for the p-value of $\alpha = 0.05$.

\subsubsection{CartPole}
\paragraph{Orthogonal trees}
To evolve orthogonal trees, we used the grammar shown in Table \ref{tab:cp_abl_ort_grammar}, which has been evolved by using the same parameters described in Table \ref{tab:cp_ort_params_ge}.
Also in this case, the fitness was computed as the mean score on 10 episodes.

\begin{table}
    \centering
    \begin{tabular}{|c|c|} \hline
        \textbf{Rule} & \textbf{Production} \\ \hline
        dt & $<if>$ \\ 
        if & $if\ <condition>\ then\ <action>\ else\ <action>$ \\ 
        condition & $input\_var\ <comp\_op>\ <const_{input\_var}>$ \\ 
        action & $<output>\ |\ <if>$ \\ 
        output & $move\_left \mid move\_right$ \\ 
        comp\_op & $lt\ |\ gt$ \\ 
        $const_x$ & [-4.8, 4.8) with step 0.5 \\ 
        $const_v$ & [-5, 5) with step 0.5 \\ 
        $const_\theta$ & [-0.418, 0.418) with step 0.01 \\ 
        $const_\omega$ & [-0.836, 0.836) with step 0.01 \\ \hline
    \end{tabular}
    \caption{Grammar used to evolve orthogonal decision trees in the CartPole-v1 environment without Q-learning.}
    \label{tab:cp_abl_ort_grammar}
\end{table}

The results are shown in Table \ref{tab:cp_abl_ort_results}.
As we can observe, while in most cases the evolution is able to evolve agents that achieve a perfect training score, they have poor generalization capabilities.
In our opinion, this is akin to overfitting.
In fact, in this case, the agents did not understand the ``value'' of going in a certain state, but just learned a rule that worked in the tested cases.
Moreover, a two-tailed Mann-Whitney U-Test gives us a p-value of $9\cdot10^{-3}$ that allows us to reject the null hypothesis (i.e. that the mean testing score come from the same distribution) with threshold $\alpha = 0.05$.

\begin{table}[ht]
    \centering
    \begin{tabular}{|c|c|c|c|c|} \hline
        \textbf{Run} & \textbf{Training score} & \textbf{Testing mean} & \textbf{Testing std} & \textbf{$\mathcal{M}$} \\ \hline
        R1 & 500.00 & \textbf{500.00} & 0 & 53.40 \\ 
        R2 & 500.00 & 436.39 & 120.91 & 71.20 \\ 
        R3 & 500.00 & 473.32 & 80.42 & 35.60 \\ 
        R4 & 500.00 & \textbf{498.3} & 11.21 & 53.40 \\ 
        R5 & 500.00 & 418.30 & 136.11 & 35.60 \\ 
        R6 & 500.00 & \textbf{489.6} & 23.46 & 35.60 \\ 
        R7 & 500.00 & \textbf{486.63} & 45.32 & 35.60 \\ 
        R8 & 500.00 & 468.20 & 82.34 & 35.60 \\ 
        R9 & 500.00 & 455.57 & 126.74 & 71.20 \\ 
        R10 & 500.00 & \textbf{483.11} & 41.99 & 71.20 \\ \hline
    \end{tabular}
    \caption{Results obtained by evolving orthogonal decision trees for the CartPole-v1 environment by using Grammatical Evolution alone.}
    \label{tab:cp_abl_ort_results}
\end{table}

\paragraph{Oblique trees}
We perform the same test also in the oblique setting.
We use the grammar shown in Table \ref{tab:cp_abl_obl_grammar} and the parameters used in Table \ref{tab:cp_obl_params_ge}.

\begin{table}
    \centering
    \begin{tabular}{|c|c|} \hline
        \textbf{Rule} & \textbf{Production} \\ \hline
        dt & $<if>$ \\ 
        if & $if\ <condition>\ then\ <action>\ else\ <action>$ \\ 
        condition & $lt((\sum\limits_{i=0}^{n\_variables} <const> * input_i), <const>)$ \\ 
        action & $<output>\ |\ <if>$ \\ 
        output & $move\_left \mid move\_right$ \\ 
        $const$ & $[-1, 1]$ with step $10^{-3}$ \\ \hline
    \end{tabular}
    \caption{Grammar used to evolve oblique decision trees in the CartPole-v1 environment without Q-learning.}
    \label{tab:cp_abl_obl_grammar}
\end{table}

The results are shown in Table \ref{tab:cp_abl_obl_results}.
We can easily observe that in this case the results are similar to the ones shown in Table \ref{tab:cp_obl_results}.
To check whether this similarity has a statistical significance, we perform a Two-tailed Mann-Whitney U-Test.
The null hypothesis states that the results obtained by using the GE with Q-learning and GE alone come from the same statistical distribution.
The p-value obtained with this test is 0.73, so we are not able to reject the null hypothesis with threshold $\alpha = 0.05$.
For this reason, we will assume that they come from the same distribution.

\begin{table}[ht]
    \centering
    \begin{tabular}{|c|c|c|c|c|} \hline
        \textbf{Run} & \textbf{Training score} & \textbf{Testing mean} & \textbf{Testing std} & \textbf{$\mathcal{M}$} \\ \hline
        R1 & 500.00 & \textbf{500.00} & 0.00 & 24.10 \\ 
        R2 & 500.00 & \textbf{500.00} & 0.00 & 24.10 \\ 
        R3 & 500.00 & \textbf{500.00} & 0.00 & 24.10 \\ 
        R4 & 500.00 & \textbf{500.00} & 0.00 & 24.10 \\ 
        R5 & 500.00 & \textbf{477.29} & 71.38 & 48.20 \\ 
        R6 & 500.00 & \textbf{500.00} & 0.00 & 24.10 \\ 
        R7 & 500.00 & \textbf{500.00} & 0.00 & 24.10 \\ 
        R8 & 500.00 & \textbf{500.00} & 0.00 & 24.10 \\ 
        R9 & 500.00 & \textbf{500.00} & 0.00 & 46.80 \\ 
        R10 & 500.00 & \textbf{500.00} & 0.00 & 24.10 \\ \hline
    \end{tabular}
    \caption{Results obtained by evolving oblique decision trees for the CartPole-v1 environment by using Grammatical Evolution alone.}
    \label{tab:cp_abl_obl_results}
\end{table}

This suggests us that, since oblique trees seem to be both more robust to noise and more stable than orthogonal trees, an agent can learn good policies in \textit{simple} environments without the need for $\mathcal{Q}$-learning.

\subsubsection{MountainCar}
\paragraph{Orthogonal trees}
We evolve orthogonal trees for the MountainCar-v0 environment by using the grammar shown in Table \ref{tab:mc_abl_ort_grammar} and the parameters shown in Table \ref{tab:mc_ort_grammar}.
Since in this case the number of episodes is low and the environment is harder to explore than CartPole, we expect GE alone to perform comparably with our approach.

\begin{table}
    \centering
    \begin{tabular}{|c|c|} \hline
        \textbf{Rule} & \textbf{Production} \\ \hline
        dt & $<if>$ \\ 
        if & $if\ <condition>\ then\ <action>\ else\ <action>$ \\ 
        condition & $input\_var\ <comp\_op>\ <const_{input\_var}>$ \\ 
        action & $<output>\ |\ <if>$ \\ 
        output & $acc\_left \mid no\_acc \mid acc\_right$ \\ 
        comp\_op & $lt\ |\ gt$ \\ 
        $const_x$ & [-1.2, 0.6) with step 0.05 \\ 
        $const_v$ & [-0.07, 0.07) with step 0.005 \\ \hline
    \end{tabular}
    \caption{Grammar used to evolve orthogonal decision trees in the CartPole-v1 environment without Q-learning.}
    \label{tab:mc_abl_ort_grammar}
\end{table}

The results are shown in Table \ref{tab:mc_abl_ort_results}.
As we expected, the performance are quite similar.
To ensure that there are no statistical significant differences between the two approaches, we performed a Two-tailed Mann-Whitney U-Test on the testing mean score obtained by using the two approaches, which stated that the null hypothesis (i.e. the scores obtained come from the same distribution) cannot be rejected with threshold $\alpha = 0.05$.

\begin{table}[ht]
    \centering
    \begin{tabular}{|c|c|c|c|c|} \hline
        \textbf{Run} & \textbf{Training score} & \textbf{Testing mean} & \textbf{Testing std} & \textbf{$\mathcal{M}$} \\ \hline
        R1 & -104.10 & \textbf{-107.44} & 16.02 & 106.8 \\ 
        R2 & -115.60 & -115.60 & 1.31 & 35.60 \\ 
        R3 & -119.40 & -119.34 & 3.66 & 17.80 \\ 
        R4 & -118.70 & -125.86 & 28.68 & 71.20 \\ 
        R5 & -119.40 & -119.34 & 3.66 & 35.60 \\ 
        R6 & -103.00 & \textbf{-106.02} & 15.21 & 106.80 \\ 
        R7 & -103.20 & \textbf{-108.31} & 18.53 & 124.60 \\ 
        R8 & -103.00 & \textbf{-105.98} & 15.03 & 89.00 \\ 
        R9 & -105.40 & \textbf{-104.71} & 3.66 & 106.80 \\ 
        R10 & -101.80 & -114.09 & 30.81 & 89.00 \\ \hline 
    \end{tabular}
    \caption{Results obtained by evolving orthogonal decision trees for the MountainCar-v0 environment by using Grammatical Evolution alone.}
    \label{tab:mc_abl_ort_results}
\end{table}

\paragraph{Oblique trees}
We perform the test also by using oblique trees.
We use the grammar described in Table \ref{tab:mc_abl_obl_grammar} with the parameters shown in Table \ref{tab:mc_obl_params_ge}.

\begin{table}
    \centering
    \begin{tabular}{|c|c|} \hline
        \textbf{Rule} & \textbf{Production} \\ \hline
        dt & $<if>$ \\ 
        if & $if\ <condition>\ then\ <action>\ else\ <action>$ \\ 
        condition & $lt((\sum\limits_{i=0}^{n\_variables} <const> * input_i), <const>)$ \\ 
        action & $<output>\ |\ <if>$ \\ 
        output & $acc\_left \mid no\_acc \mid move\_right$ \\ 
        $const$ & $[-1, 1]$ with step $10^{-3}$ \\ \hline
    \end{tabular}
    \caption{Grammar used to evolve oblique decision trees in the CartPole-v1 environment without Q-learning.}
    \label{tab:mc_abl_obl_grammar}
\end{table}

The results are shown in Table \ref{tab:mc_abl_obl_results}.
While these results seem to be better than the ones shown in \ref{tab:mc_obl_results}, they do not seem to be statistically significant according to a two-tailed Mann-Whitney U-Test (p-value 0.38).
Thus, this seems to confirm our hypothesis that states that solving MountainCar-v0 with oblique trees seems to be harder than the case with orthogonal trees (with the proposed grammar).

\begin{table}[ht]
    \centering
    \begin{tabular}{|c|c|c|c|c|} \hline
        \textbf{Run} & \textbf{Training score} & \textbf{Testing mean} & \textbf{Testing std} & \textbf{$\mathcal{M}$} \\ \hline
        R1 & -102.00 & \textbf{-105.83} & 16.49 & 139.80 \\ 
        R2 & -97.10 & \textbf{-106.8} & 23.61 & 93.40 \\ 
        R3 & -102.00 & \textbf{-107.74} & 20.33 & 70.20 \\ 
        R4 & -101.40 & -111.36 & 23.98 & 70.00 \\ 
        R5 & -101.80 & \textbf{-109.71} & 23.12 & 93.40 \\ 
        R6 & -101.90 & \textbf{-108.79} & 21.58 & 116.80 \\ 
        R7 & -101.30 & -110.56 & 25.63 & 93.40 \\ 
        R8 & -97.20 & \textbf{-108.09} & 30.34 & 116.60 \\ 
        R9 & -101.80 & \textbf{-107.48} & 20.00 & 93.20 \\ 
        R10 & -105.80 & \textbf{-107.14} & 14.66 & 93.20 \\ \hline
    \end{tabular}
    \caption{Results obtained by evolving oblique decision trees for the MountainCar-v0 environment by using Grammatical Evolution alone.}
    \label{tab:mc_abl_obl_results}
\end{table}

\subsubsection{LunarLander}
Finally, we perform the same test on LunarLander-v2, using only oblique trees.
We use the grammar described in Table \ref{tab:ll_abl_obl_grammar} and the parameters present in Table \ref{tab:ll_obl_params_ge}.
We expect that in this task, since it is harder than the previous two, GE performs worse than our approach.

\begin{table}
    \centering
    \begin{tabular}{|c|c|} \hline
        \textbf{Rule} & \textbf{Production} \\ \hline
        dt & $<if>$ \\ 
        if & $if\ <condition>\ then\ <action>\ else\ <action>$ \\ 
        condition & $lt((\sum\limits_{i=0}^{n\_variables} <const> * input_i), 0)$ \\ 
        action & $<output>\ |\ <if>$ \\ 
        output & $nop \mid left\_engine \mid main\_engine \mid right\_engine$ \\ 
        $const$ & $[-1, 1]$ with step $10^{-3}$ \\ \hline
    \end{tabular}
    \caption{Grammar used to evolve oblique decision trees in the LunarLander-v2 environment without Q-learning.}
    \label{tab:ll_abl_obl_grammar}
\end{table}

The results of this experiment are shown in Table \ref{tab:ll_abl_obl_results}.
According to our expectations, we are able to solve the task only in the 60\% of the cases.
Moreover, we perform a two-tailed Mann-Whitney U-Test to test the statistical significance of the differences between the two approaches (on the mean testing score).
We obtain a p-value of $0.017$ that allows us to reject the null hypothesis.
We can thus hypothesize that the use of the two-level optimization technique gives us a boost in performance in complex environments such as LunarLander-v2.

\begin{table}[ht]
    \centering
    \begin{tabular}{|c|c|c|c|c|} \hline
    \textbf{Run} & \textbf{Training score} & \textbf{Testing mean} & \textbf{Testing std} & \textbf{$\mathcal{M}$} \\ \hline
    R1 & -88.83 & -90.25 & 34.46 & 147.40 \\ 
    R2 & 216.45 & 172.83 & 76.83 & 60.50 \\ 
    R3 & 241.37 & \textbf{228.12} & 47.7 & 59.10 \\ 
    R4 & 272.25 & \textbf{252.88} & 54.27 & 115.40 \\ 
    R5 & 231.83 & \textbf{216.65} & 60.59 & 117.50 \\ 
    R6 & 103.36 & 49.15 & 127.21 & 89.00 \\ 
    R7 & 266.90 & \textbf{251.28} & 42.95 & 120.30 \\ 
    R8 & 247.40 & \textbf{205.25} & 73.41 & 58.40 \\ 
    R9 & 254.00 & \textbf{243.95} & 34.58 & 88.30 \\ 
    R10 & 5.47 & -57.84 & 120.95 & 122.40 \\ \hline
    \end{tabular}
    \caption{Results obtained by evolving oblique decision trees for the LunarLander-v2 environment by using Grammatical Evolution alone.}
    \label{tab:ll_abl_obl_results}
\end{table}

\subsection{Interpretation of the solutions}
In this subsection, we will look at the agents produced and try to interpret the policies.

\subsubsection{CartPole}
\paragraph{Orthogonal tree}
The tree shown in Figure \ref{fig:cp_ort_best} is extremely easy to interpret.
In fact, this agent moves the cart to the left if 
\begin{equation}
    \omega < 0.074 \land \theta < 0.022
    \label{eq:cp_ort_condition}
\end{equation}
otherwise, it moves the cart to the right.
Note that there is a case in which the pole is falling to the right but the agent moves the cart to the left: $\theta \in [0, 0.022) rad \land \omega \in [0, 0.074) rad/s$.
This is not a problem because when the agent moves the cart to the right, it increases the velocity of the pole, resulting in a ``move\_right'' action in the subsequent steps.

\paragraph{Oblique tree}
In this case, the interpretation of the policy is a bit harder.
The condition used by the agent to discriminate between the two states is:
\begin{equation}
    - 0.274x_k - 0.543v_k - 0.904\theta_k - 0.559\omega_k < -0.169
    \label{eq:cp_obl_condition}
\end{equation}
where $k$ refers to the current timestep.
To simplify the process, we write Equation \ref{eq:cp_ort_condition} as the following:
\begin{equation}
    - ax_k - bv_k - c\theta_k -d\omega_k < t
    \label{eq:cp_obl_condition_wletters}
\end{equation}

First of all, we want to analyze the role of the constant $t$ in the policy.
By testing it with different values (i.e. $t = -0.169$, $t = 0.169$, $t=-0.1$, $t=0.1$, $t=0$) we observed that it holds that the final point in which the pole is balanced can be obtained as follows:
\begin{equation}
    x_n \approx -\frac{t}{a} 
\end{equation}
where $n$ is the index of the last timestep.
For simplicity, let's assume that $x_n = -\frac{t}{a}$.
This means that we can rewrite Equation \ref{eq:cp_obl_condition_wletters} as follows:
\begin{equation}
    - x_k - \frac{b}{a} v_k - \frac{c}{a} \theta_k - \frac{d}{a} \omega_k < \frac{t}{a} = -x_n
\end{equation}
We can then perform other steps and obtain:
\begin{equation}
- x_k - b' v_k - c' \theta_k - d' \omega_k < -x_n \Rightarrow
\end{equation}
\begin{equation}
- b' v_k - c' \theta_k - d' \omega_k < -x_n + x_k \Rightarrow
\end{equation}
\begin{equation}
- b' v_k - c' \theta_k - d' \omega_k < -x_n + x_{n-1} - x_{n-1} + ... + x_k = \sum\limits_{j=n}^{k+1} -x_j + x_{j-1}
\label{eq:cp_obl_condition_wdifferences}
\end{equation}

Then, by noting that
\begin{equation}
\frac{x_k - x_{k-1}}{\tau} = v_k
\end{equation}
we can rewrite Equation \ref{eq:cp_obl_condition_wdifferences} as:

\begin{equation}
- b' v_k - c' \theta_k - d' \omega_k < - \sum\limits_{j=k+1}^{n} v_j \tau
\end{equation}
\begin{equation}
- c' \theta_k - d' \omega_k < - \sum\limits_{j=k}^n g_j v_j \tau
\end{equation}
where 
\begin{equation*}
g_j = \begin{cases}
\frac{-b'}{\tau} & \text{if $j == k$}\\
1 & \text{otherwise}
\end{cases}
\end{equation*}

Now, by observing that
\begin{equation}
\frac{\theta_k - \theta_{k-1}}{\tau} = \omega
\end{equation}
we obtain
\begin{equation}
    -c' \theta_k - d' \frac{\theta_k - \theta_{k-1}}{\tau} < - \sum\limits_{j=k+1}^{n} g_j v_j \tau
\end{equation}
\begin{equation}
    - (d' + \tau c') \theta_k + d' \theta_{k - 1} < - \tau^2 \sum\limits_{j=k+1}^{n} g_j v_j
\end{equation}

Finally, noting that after that usually, in the first 50 timesteps of the simulations the velocities are high ($\underset{k}{max} \mid v_k \mid < 1.5$) and then the velocity become small ($\underset{k}{max} \mid v_k \mid < 0.55$) because the pole is balanced, we can write that:
\begin{equation}
    \mid\sum\limits_{j=k+1}^{n} g_j v_j\mid \lessapprox  \frac{b'}{\tau} \cdot 1.5 + 49 \cdot 1.5 + 450 \cdot 0.55 = 420
\end{equation}
where the approximate equality holds in the worst case (i.e. $k=0$ and all the velocities have the same sign).
However, considering that in our observations the magnitude of the velocities was usually significantly smaller than the maximum and that the summation is multiplied by $\tau^2$ ($\tau = 0.02$ in this environment), we can safely consider only the term with the highest magnitude, i.e. $\frac{b'}{\tau}v_k$.
Moreover, using only $v_k$ sets $x_n\approx 0$, which makes the system easier to understand intuitively.

Then, we obtain
\begin{equation}
    - (d' + \tau c') \theta_k + d' \theta_{k - 1} < \tau b' v_k
\end{equation}
\begin{equation}
    c \theta_k > - (b v_k + d \omega_k)
\end{equation}

Approximating the constants, we set $b=0.543\approx0.5$, $c=0.904\approx1$, $d=0.559\approx0.5$, so the final policy is\footnote{Implementing this policy by using the $\omega$ given by the environment may give slightly lower than perfect scores, in our opinion this is due to the error carried by the integration method used. On the other hand, using $\omega_k=(\theta_k-\theta_{k-1})/\tau$ gives the desired results.}:
\begin{equation*}
\pi(x, v, \theta, \omega) = \begin{cases}
move\_right & \text{if $\theta_k > -\frac{1}{2}(v_k + \omega_k)$}\\
move\_left & \text{otherwise}
\end{cases}
\end{equation*}

A dimensionally consistent policy is $\theta_k + \frac{1}{2} (v_k / l + \omega) \frac{n_{ts} \tau}{\tau} > 0$, where $l=1$ is the pole length and $n_{ts}$ is the number of steps that we are taking into consideration to balance the pole (in our case $n_{ts} = 1$.
This policy can be interpreted as follows.
If the sum of the current angle and the mean angle given by the two contributions (i.e. linear velocity of the cart and angular velocity of the pole) are positive (it is a kind of ``prediction'' of the future angle), then move the cart to the right, because it is going to fall to the right. 
Otherwise, move the cart to the left.

\subsubsection{MountainCar}
\paragraph{Orthogonal tree}
Also in this case, the orthogonal tree (Figure \ref{fig:mc_ort_best}) is easy to interpret.
In fact, if we look at the leaves, we see that the agent accelerates to the left only in two cases: $(v < 0 \land x > -0.9) \lor (v \in [0, 0.035) \land x \in [-0.4, -0.3])$.
This means that the agents accelerates to the left when: 
\begin{itemize}
    \item it is going towards the hill on the left to build momentum and it is far from the border ($x > -0.9$), so it tries to maximize the potential energy of the car
    \item velocity is positive but not enough ($v < 0.035$) and it is near the valley
\end{itemize}
In all the other cases, the agent accelerates to the right.

\paragraph{Oblique trees}
In this case, the agent accelerates to the left when both conditions are false.
This means that we have to solve the following system of two inequalities:
\begin{equation*}
\begin{cases}
0.717 \widehat{x} - 0.697 \widehat{v} \geq -0.229 \\
0.138 \widehat{x} - 0.883 \widehat{v} \geq -0.389 \\
\end{cases}
\end{equation*}

\begin{figure}[!ht]
    \centering
    \includegraphics[scale=0.7]{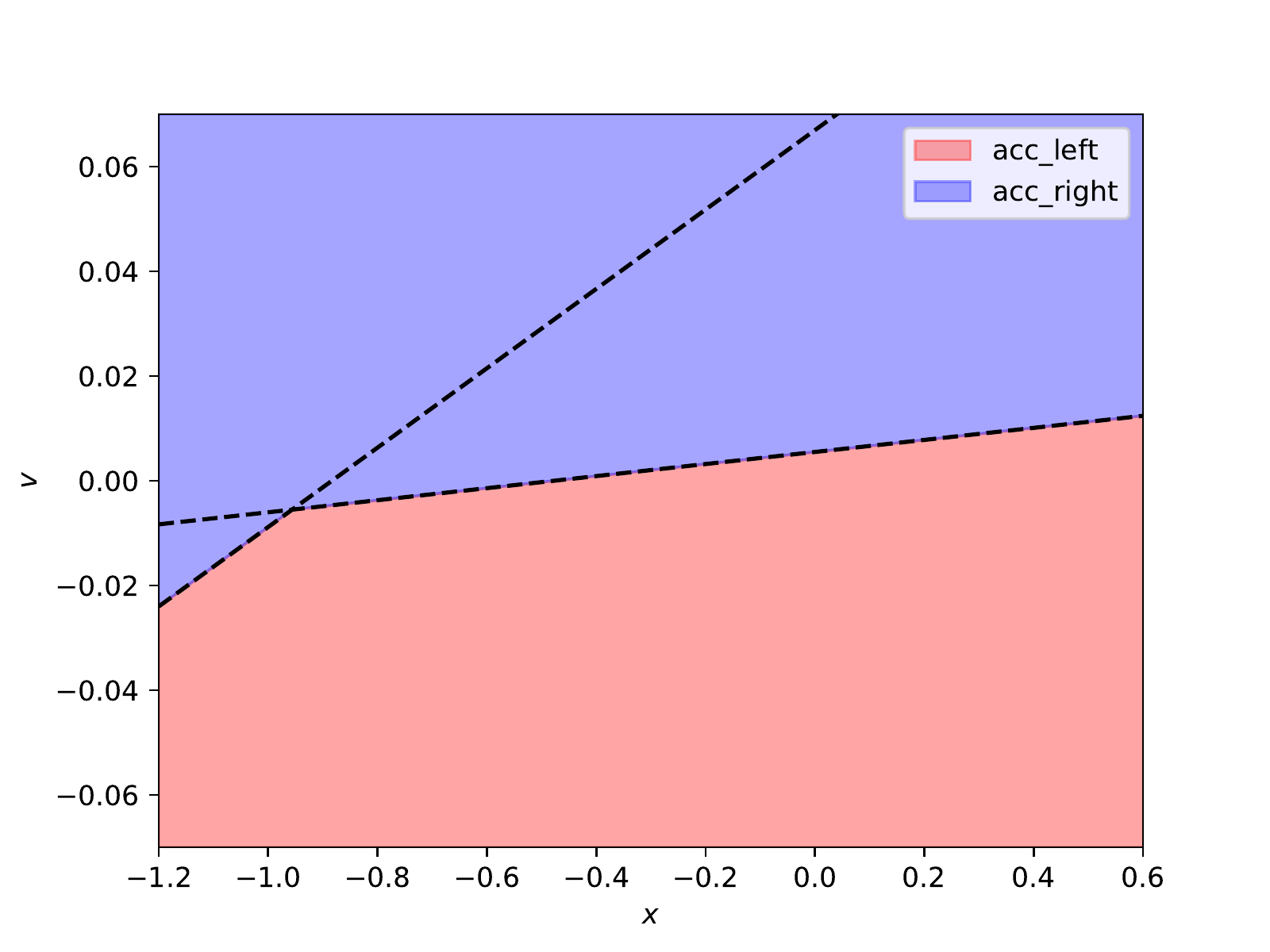}
    \caption{Decision regions for the best oblique tree evolved in the MountainCar-v0 environment.}
    \label{fig:mc_obl_decisionregions}
\end{figure}

This means that the agent accelerates to the left when $v \leq 7.5799 \cdot 10^{-2} \cdot x+ 6.6955 \land v \leq 1.1516 \cdot 10^{-2} \cdot x + 5.495 \cdot 10^{-3}$. 
This corresponds to the decision regions shown in Figure \ref{fig:mc_obl_decisionregions}.

It is important to note that the lack of robustness for this solution does not allow us to further approximate the constants of the two hyperplanes.

\subsubsection{LunarLander}
In this case, since the oblique tree (Figure \ref{fig:ll_best_tree}) has 4 conditions and 8 unknowns, it is a bit harder to interpret.
\paragraph{First condition}
This condition, when it evaluates to False, turns on the right engine for a timestep.
So, we turn on the right engine when
\begin{equation}
    a p_x - b p_y + c v_x - d v_y - e \theta - f \omega - g c_l - h c_r \geq 0
    \label{eq:ll_condition1}
\end{equation}
where $a, b, ..., h$ replace the constants shown in Figure \ref{fig:ll_best_tree}.

To simplify the analysis, let's assume $c_l = c_r = 0$, since they can assume only two values: $0, 1$.
This simplification does not affect the generality of our analysis, since we are only assuming that there is no contact with the ground.
We can simply say that, when contact with the ground happens, then the threshold is not $0$ anymore, but it can take the following values: $0.2$ (only right leg touches the ground), $0.597$ (only right leg touches the ground), $0.797$ (both legs touch the ground).

So, we can rewrite condition \ref{eq:ll_condition1} as follows:
\begin{equation}
    a p_x + c v_x - b p_y - d v_y - e \theta - f \omega \geq 0
\end{equation}
By merging some terms we obtain:
\begin{equation}
    a (p_x + v_x c') - b (p_y - v_y d') -e (\theta - \omega f') \geq 0
\end{equation}
We analyzed the terms in parenthesis and we discovered that they approximate the position (or the angle) in the following timestep.
The constants $c' \approx f' \approx 1.23$ lead to an overestimation of the magnitude of the future position (or angle), while the constant $d'\approx0.53$ increases the precision of the approximation.
By denoting the predictions of the next position on  x, y and $\theta$ with $p_x^{k+1}$, $p_y^{k+1}$, $\theta^{k+1}$ respectively, we can write:
\begin{equation}
    a p_x^{k+1} - b p_y ^ {k+1} \geq e \theta^{k+1}
\end{equation}
To understand how this condition works, let's suppose that $p_x^{k+1}\approx0$ (i.e. the lander is in the center of the environment). 
Then, if $p_y^{k+1}\approx 1$ (i.e. near the starting point), we will fire the right engine if $\theta^{k+1} \leq -b/e\approx-0.15 rad$, i.e. the angle of the lander is going to fall to the right.
When $p_y^{k+1}\approx 0$ (i.e. near the landing pad), the agent will fire the right engine if $\theta^{k+1} \leq 0$, so we can say that the farther the lander is from the landing pad (vertically), the more margin we have on the threshold of the angle.
Let's now suppose that $p_y^{k+1}=0$ to study the effect of $p_x^{k+1}$ on the policy.
Then, we can say that the agent turns on the right engine when $\theta \leq \frac{a}{e} p_x^{k+1}$ so, when the agent is on the right part of the environment, the agent uses a linear threshold to activate the engine in order to avoid both high angles and high displacements from the landing pad location.
Similarly, when $p_x^{k+1}$ is negative, the threshold is negative so the agent tries both to compensate negative angles (that would move it farther on the left) and distance from the landing point.

\paragraph{Second condition}
The second condition, when evaluates to True, leads to the firing of the left engine.
Also in this case, let's neglect the terms $c_l$ and $c_r$.
We can write the condition as:
\begin{equation}
    a p_x - b p_y + c v_x - d v_y - e \theta - f \omega < 0
\end{equation}
Of course, in this case the coefficients $a$, ..., $f$ are different from the previous ones.
By grouping the terms as before we obtain
\begin{equation}
    a (p_x + v_x c') - b (p_y + v_y d') -e (\theta + \omega f') < 0
\end{equation}
Also in this case, the constants seem to have the same role (i.e. some lead to overestimation of the next position and some to a better estimate) so we can write:
\begin{equation}
    a p_x^{k+1} - b p_y^{k+1} < e \theta^{k+1}
\end{equation}

This means that this condition is easy to understand given the previous one: it is the opposite.
This means that we can use the same reasoning used above to understand this condition.

\paragraph{Third condition}
This condition handles the firing of the main engine.
For this reason, we expect it to work differently from the previous two.
In fact, we can easily observe that the signs of the terms in $x$ and $y$ are inverted. 
Moreover, the two angular terms do not have the same sign.
Also in this case, let's use $a$, ..., $f$ to rename the constants and ignore $c_l$ and $c_r$.
This leads to:
\begin{equation}
    -a p_x + b p_y - c v_x + d v_y - e \theta + f \omega < 0
\end{equation}

By performing a grouping of the variables similarly to the previous to conditions we obtain:
\begin{equation}
    -a (p_x + v_x) + b (p_y + v_y) - (c - a) v_x + (d - b) v_y - e \theta + f \omega < 0
\end{equation}
Then, by denoting with $v^{k+1}$ and $v^{k-1}$ the value of the variable $v$ in the next and the previous timestep respectively, we can write:
\begin{equation}
    -a p_x^{k+1} + b p_y^{k+1} - c' v_x + d' v_y - e \theta + f \frac{\theta - \theta^{k-1}}{\tau} < 0
\end{equation}
An experimental measurement of the $\tau$ variable led us to set $\tau=0.05$.
By multiplying all the members by $\tau$ we obtain:
\begin{equation}
    -\tau a p_x^{k+1} +\tau  b p_y^{k+1} -\tau  c' v_x +\tau  d' v_y - \tau e \theta + f (\theta - \theta^{k-1}) < 0
\end{equation}
Then, by noting that $\tau a\approx5\cdot 10^{-3}$, $\tau b\approx6.7\cdot 10^{-3}$, $\tau c'\approx3.5\cdot 10^{-2}$, $\tau d'\approx2.6\cdot 10^{-2}$ and $\tau e\approx 10^{-2}$, we can decide to neglect the effects of the first two terms.
So we have:
\begin{equation}
    - \tau c' v_x + \tau d' v_y + (f - \tau e) \theta - f \theta^{k-1} < 0 
\end{equation}
By merging the terms in $\theta$ and $\theta^{k-1}$ we obtain: 
\begin{equation}
    - c' v_x + d' v_y + (f - \tau e) \omega + \tau e \theta^{k-1} < 0
\end{equation}
By moving all the terms except the one in $\omega$ to the second member we get:
\begin{equation}
    \omega < \frac{1}{f - \tau e} (c' v_x - d' v_y - \tau e \theta^{k-1})
\end{equation}
Then, by noting that all the states that are tested in this condition have $c' \overline{\mid v_x\mid} \approx 5 d' \overline{\mid v_y\mid}$ and $c' \overline{\mid v_x\mid}\approx120 e \overline{\mid \theta^{k-1}\mid }$ (where $\overline{v}$ is the mean value of the variable v), we can neglect (as shown by experimental results) the effects of $v_y$ and $\theta^{k-1}$.
Finally, the rule used to fire the main engine is:
\begin{equation}
    \omega < c'' v_x
\end{equation}

While we expected the main engine to depend on $p_y$ or $v_y$, by analyzing the activation of the condition in several episodes we found that this rule represents the landing phase. 
In fact, the goal of this rule is to balance angular velocity and linear velocity to make the agent gently stop on the landing pad.

\paragraph{Fourth condition}
This condition, when evaluates to True, does not fire any engine.
On the other hand, when it evaluates to False, it fires the main engine.

The condition is the following (also in this case we replace the constants with letters):
\begin{equation}
    a p_x - b p_y - c v_x - d v_y - e \theta + f \omega < 0
\end{equation}

By analyzing the mean values of the variables and their coefficients we obtain: $a \overline{\mid p_x\mid}\approx 8.5 \cdot 10^{3}$, $b \overline{\mid p_y\mid}\approx 8.7 \cdot 10^{3}$, $c \overline{\mid v_x\mid}\approx 7 \cdot 10^{2}$, $d \overline{\mid v_y\mid}\approx 2.5 \cdot 10^{2}$, $e \overline{\mid \theta \mid}\approx 1.3 \cdot 10^{2}$, $f \overline{\mid \omega \mid}\approx 4.7 \cdot 10^{2}$.
This suggests that we can neglect the values of $p_x$, $p_y$, $\theta$ because their mean value is low w.r.t. the maximum. 
The experiments confirmed that these variables have a low impact on the performance of the agent.

So, the agent does not fire any engine when:
\begin{equation}
    \omega < \frac{c}{f} v_x + \frac{d}{f} v_y
\end{equation}

This seems an extension of what we obtained in the previous condition, where we also have a dependency from $v_y$.
Moreover, it is important to note that this check is performed only when the third condition is not true.
Finally, from experiments we observed that this condition is true usually when the agent has successfully landed. 
In this case, the terms in $c_l$ and $c_r$ can be seen as a further margin to the agent, so that when a leg touches the ground the agent is more likely to not fire any engine.

In the opposite case, i.e. when $\omega \geq c' v_x + d' v_y$, we the agent turns on the main engine to balance the high angular velocity of the agent. Note, again, that if the angular velocity is too low it is balanced by the previous condition.

\subsubsection{Considerations}
In this subsection, we interpreted the policy produced in various settings. 
We showed that the decision trees produced are interpretable and give an understanding about how the agent works.
It is important to note that in several cases we performed approximations to ease the understanding process. 
However, this is not a limitation of the method, because more exact interpretations can be obtained by not neglecting details.
This is especially important in high-stakes or safety-critical settings, where humans need to have a thorough understanding to validate and trust the systems produced.

Finally, while some solutions may seem hard to interpret (i.e. oblique decision trees), it is important to see the them in a bigger context: while they may not be easy to interpret at a first sight, their analysis is pretty straightforward (as shown earlier).
On the other hand, black-box models (such as deep neural networks) are way harder to inspect, due to the significantly bigger number of operations performed in the decision making process.

\section{Conclusions}
\label{sec:conclusions}
While in recent years AI made a huge progress, the need of being able to understand \textit{how} a model works is becoming more and more important.
To overcome this issue, significant effort was put to advance the XAI field.
However, XAI is not always a suitable solution.
In fact, they suffer from some problems that make their use unsafe in safety-critical or high-stakes processes.

Interpretable AI, instead, consists in using transparent approaches in order to have a complete understanding of what happens in the model. 
However, these models are not widely used in practice because of their widely-thought lower performance.

In this paper, we propose a two-level optimization method that allows to induce decision trees that can perform reinforcement learning.

Our results show that the proposed approach is able to generate decision trees that are comparable or even better than the non-interpretable state-of-the-art (from the performance point of view) while having significantly better interpretability.
Furthermore, the results obtained in this work suggest that the widely though performance-interpretability trade-off does not always hold (as suggested by \cite{rudin_stop_2019}) and that interpretable models can be competitive with state-of-the-art techniques. For this reason, research in this field must be encouraged.

Moreover, we compared the solutions obtained to the state-of-the-art from the point of view of the interpretability. 
While the metric of interpretability does not perfectly suit our purpose, we can easily observe the difference in complexity with respect to black-box models.
While we expect that changing the metric of interpretability does not significantly affect the difference w.r.t. black-box models, we think that future work should focus on the study of more tailored interpretability metrics (i.e. tailored on machine learning models).

Since it is important for practical applications, we also compared our solutions to the interpretable (and publicly available) state-of-the-art w.r.t. robustness to input noise.
The results show that our approach is comparably or more robust than the other solutions.

Finally, we demonstrated that the produced agents can be interpreted, practically showing the advantage of interpretable models w.r.t. black boxes.


Other future developments include: experimental tests on more complex reinforcement learning domains; the extension of the proposed method to the imitation learning domain; the development of a method that can automatically tune the constants, reducing the prior knowledge that must be included in the grammar; a flexible grammar that easily allows oblique trees to become orthogonal, to automatically choose the appropriate type of splits depending on the problem.

\bibliographystyle{elsarticle-num}
\bibliography{main}

\end{document}